\documentclass[10pt,twocolumn,letterpaper]{article}

\usepackage[pagenumbers]{wacv}

\usepackage{array}
\newcommand{\red}[1]{{\color{red}#1}}
\newcommand{\todo}[1]{{\color{red}#1}}
\newcommand{\TODO}[1]{\textbf{\color{red}[TODO: #1]}}

\definecolor{wacvblue}{rgb}{0.21,0.49,0.74}
\usepackage[pagebackref,breaklinks,colorlinks,allcolors=wacvblue]{hyperref}

\def\confName{WACV}
\def\confYear{2027}

\title{Volley3D: Unified 3D Full-Sequence Reconstruction from Monocular Broadcasts}
\vspace{-1.8em}

\author{
  Noâm Boussouf$^{1,2}$\thanks{Corresponding author (e-mail: \href{mailto:noam.boussouf@keio.jp}{noam.boussouf@keio.jp}).}\quad
  Adrien Maglo$^{3}$\quad
  Mohamed Chaouch$^{3}$\quad
  Hideo Saito$^{1}$\\[2mm]
  $^{1}$Keio University, Japan \quad $^{2}$École Centrale de Nantes, France \\ \quad $^{3}$Université Paris-Saclay, CEA-LIST, F-91120, Palaiseau, France
\\
  {\tt\small noam.boussouf@keio.jp}\\[1mm]
}

\begin{document}

\twocolumn[{
  \renewcommand\twocolumn[1][]{#1}
  \vspace{-1.8em}
  \maketitle
  \begin{center}
    \centering
    \captionsetup{type=figure}
    \begin{subfigure}[b]{0.33\textwidth}
        \centering
        \includegraphics[width=\linewidth]{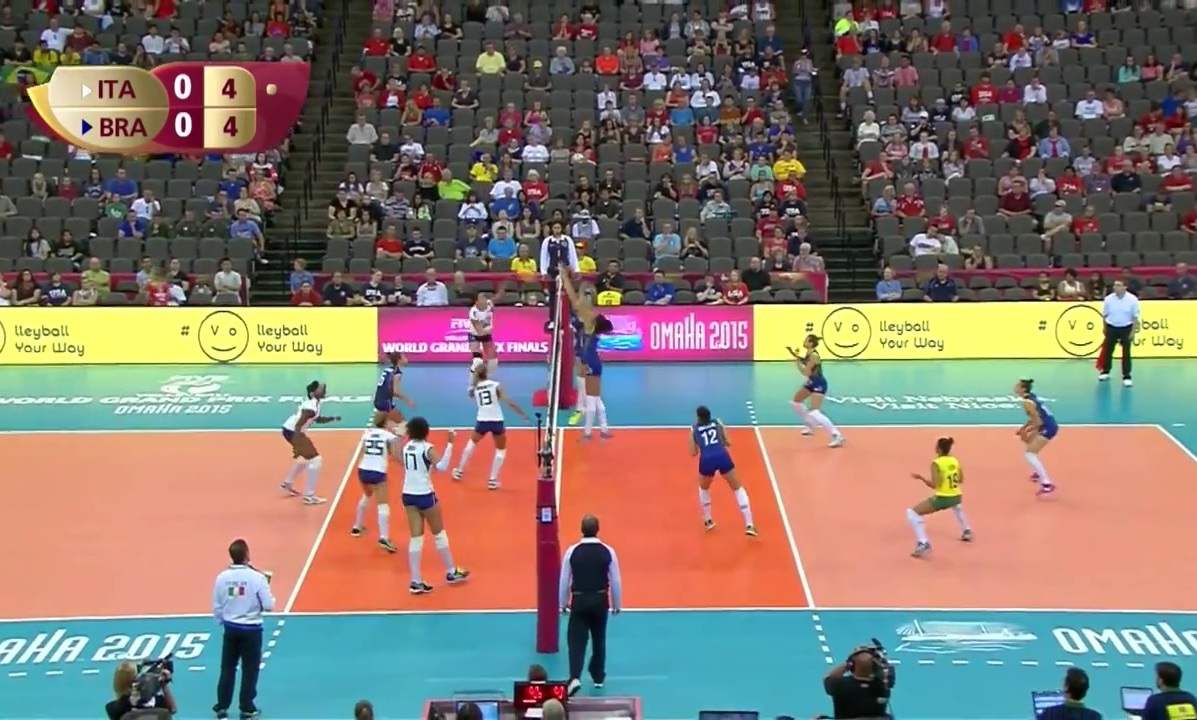}
        \caption{Input 2D Broadcast View}
        \label{fig:teaser_broadcast}
      \end{subfigure}
      \hfill
      \begin{subfigure}[b]{0.33\textwidth}
        \centering
        \includegraphics[width=\linewidth]{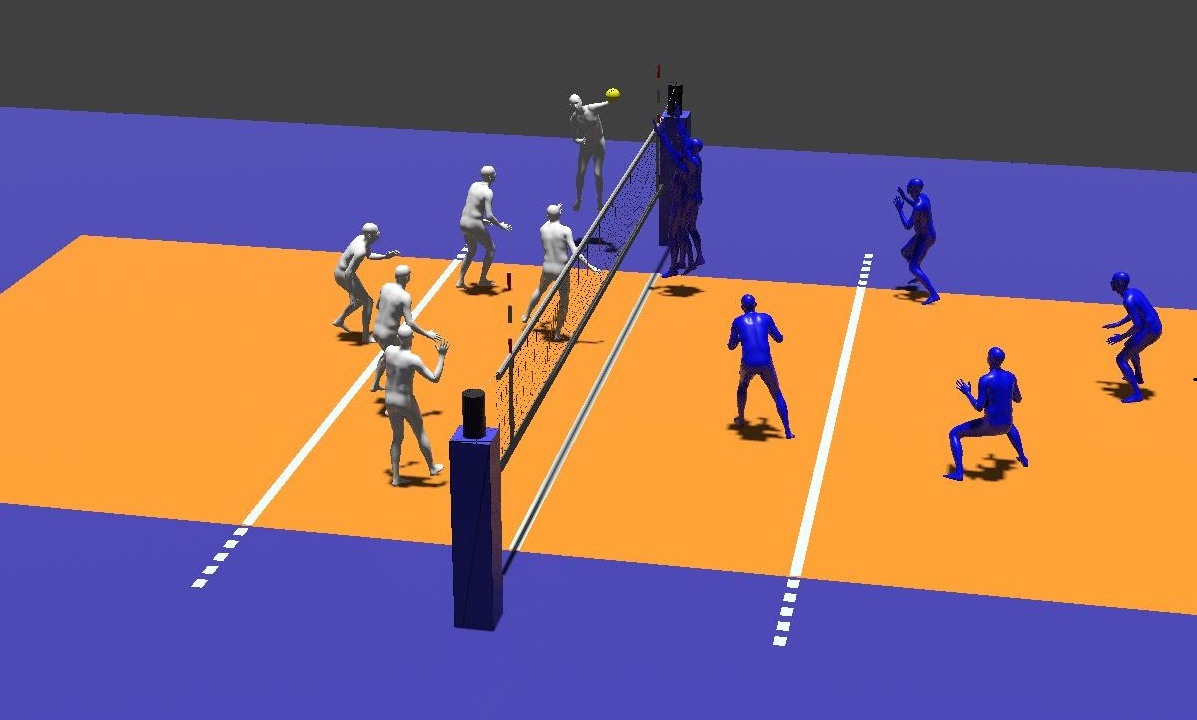}
        \caption{Reconstructed 3D Scene (Broadcast POV)}
        \label{fig:teaser_pov1}
      \end{subfigure}
      \hfill
      \begin{subfigure}[b]{0.33\textwidth}
        \centering
        \includegraphics[width=\linewidth]{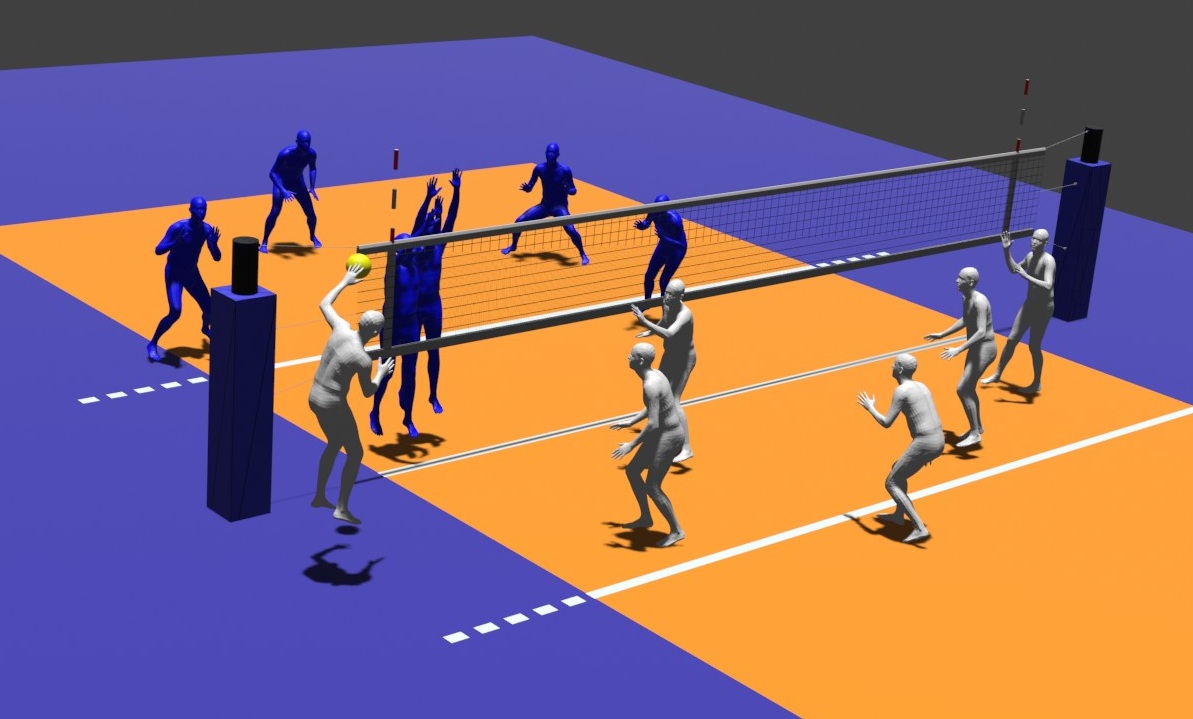}
        \caption{Reconstructed 3D Scene (Other POV)}
        \label{fig:teaser_pov2}
      \end{subfigure}
      \caption{\textbf{Monocular 3D collective reconstruction from broadcast volleyball video.} From a single uncalibrated broadcast sequence (a), our framework recovers 3D player poses, jump dynamics, and physical ball trajectories in global court coordinates, enabling realistic novel-view synthesis from multiple vantage points (b, c).}
      \label{fig:teaser}
  \end{center}
}]

\begin{abstract}
    3D reconstruction from single-view sports broadcast is a pivotal capability for automated analytics and immersive AR/VR replays, yet notoriously ill-posed due to difficult-to-calibrate moving cameras and depth ambiguity. Unlike widely studied racket sports, team sports such as volleyball feature unpredictable multi-contact sequences and high-altitude jumps. These dynamics categorically invalidate the standard planar projection assumptions relied upon by existing methods. To resolve these limitations without 3D training supervision, we propose a complete, physics-grounded pipeline tailored to monocular broadcast volleyball. First, we include a correction module to the perspective drift induced by ground-plane projection by coupling temporal jump spotting with Newtonian interpolation for 3D player placement. Second, for 3D ball trajectory estimation, we design a dedicated sub-trajectory segmentation architecture alongside a collision-matching algorithm. By anchoring physical equations to reconstructed 3D hand vertices, we lift noisy ball 2D detections into a spatiotemporally coherent 3D scene. Evaluations on multi-camera benchmarks demonstrate our approach significantly reduces 3D spatial drift compared to 2D-reprojection baselines. Project Webpage : \href{https://noambsf.github.io/Volley3D}{noambsf.github.io/Volley3D}
\end{abstract}    
\section{Introduction}
\label{sec:intro}

In recent years, computer vision has revolutionized sports analysis, enabling automated data extraction for tactical insights and performance optimization. Within this domain, 3D reconstruction has emerged as a pivotal task. While multi-camera setups currently power elite-level tools like semi-automated offside technology, their reliance on expensive, synchronized infrastructure renders them inapplicable for universally accessible broadcast footage. Shifting to a monocular paradigm is therefore essential. Not only to democratize sports analytics but also to unlock next-generation immersive applications. Specifically, lifting a standard 2D broadcast feed into 3D space enables viewers to experience historical matches through Virtual Reality (VR) and synthesizes novel camera angles for augmented reality (AR) broadcasting. However, recovering depth from a single uncalibrated viewpoint transforms 3D reconstruction into a highly ill-posed inverse problem. While recent monocular frameworks show promising results in sports with constrained, ground-bound movements (e.g., table tennis~\cite{gossard2025tt3d} or badminton~\cite{lin2023vird}), collective sports featuring unpredictable aerial gameplay remain under-explored.

Volleyball stands out as a highly challenging yet under-researched domain. Due to its unique physical and spatial dynamics, it serves as a rigorous testing ground for monocular 3D vision; a framework capable of accurately reconstructing a volleyball match possesses the versatility to adapt to various other sports. Unlike racket sports characterized by strictly alternating turns, volleyball features unpredictable, intra-team multi-contact sequences. Chief among its challenges is the omnipresence of high-amplitude vertical jumps, which completely invalidates the static ground-contact assumptions relied upon by current state-of-the-art pipelines. Furthermore, reconstructing this 3D space from broadcast feeds is severely hindered by the lack of reliable 3D calibration cues. The court provides almost no known 3D reference points above the ground plane. While the net features standardized dimensions, its reference antennas are visually ambiguous from distant broadcast angles. Consequently, while pseudo-3D calibration might be acceptable near the net, it degrades toward the backcourt, making traditional reprojection heuristics highly unstable. Finally, this depth ambiguity is exacerbated by moving cameras, low broadcast angles, and severe mutual players occlusions.

Compounding these physical and visual hurdles is severe data scarcity. Unlike major sports such as football or basketball, volleyball lacks large-scale datasets providing 3D ground-truth for both player and ball trajectories. This deficiency renders the training of massive, end-to-end deep learning architectures for direct 2D-to-3D mapping unfeasible. Consequently, we propose a modular pipeline that decomposes the overarching problem into specialized computer vision sub-tasks. Importantly, instead of treating ball trajectory estimation independently of human motion, our framework explicitly addresses the lack of 3D ball data by mathematically coupling them. By anchoring the ball's physical trajectory to detected player interactions, we enforce strict spatiotemporal consistency, establishing a coherent 3D reconstruction from a single video stream without requiring 3D training supervision.

\begin{figure*}[ht]
  \centering
  \includegraphics[width=1.0\textwidth]{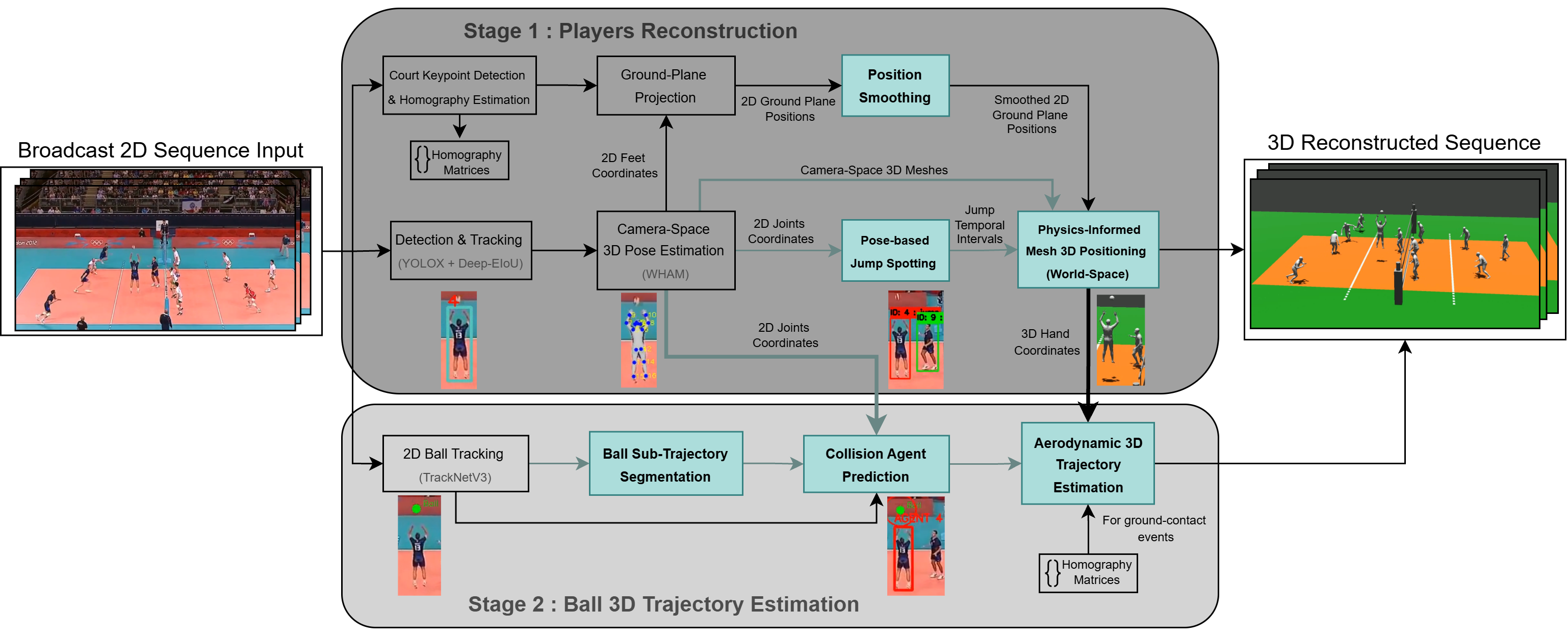}
    \caption{Overview of our proposed two-stage monocular 3D reconstruction pipeline. In the first stage, we extract and position the athletes in 3D world-space, combining smoothed homographic projections with a pose-based jump spotting module to handle aerial phases. In the second stage, we segment 2D ball tracks and explicitly anchor them to the reconstructed athletes, using the predicted 3D collision points to physically constrain the final 3D ballistic trajectory. The blue modules highlight the contributions introduced in this paper.}
  \label{fig:pipeline}
  \vspace{-1.1em}
\end{figure*}

Specifically, we propose a two-stage framework for monocular 3D volleyball reconstruction (\cref{fig:pipeline}). The first stage focuses on player reconstruction, leveraging court keypoint detection, 2D tracking, and camera-space 3D pose extraction, followed by player position temporal smoothing. Building upon this foundation, the core innovation of this stage lies in handling jumping players. Because monocular depth recovery is inherently ill-posed, standard pipelines rely on a strict ground-contact assumption to anchor players onto the court plane~\cite{rematas2018soccer, lin2023vird, gossard2025tt3d, 2004soccer}. However, this simplifying constraint categorically fails during vertical extensions. When a player jumps, their vertical elevation in the 2D image plane is misinterpreted by the homography matrix as a retreat in depth, causing severe spatial drift. To overcome this, we integrate a pose-based event spotting module tailored for precise takeoff and landing localization. We then apply Newtonian equations of motion to interpolate the player's true 3D trajectory in world space, yielding visually and physically accurate airborne localizations.

The second stage reconstructs the 3D ball trajectory by exploiting these player-anchored mechanics. First, we introduce a temporal sub-trajectory segmentation architecture that outperforms existing state-of-the-art models in isolating ball deflection events. Next, a training-free, heuristic-based collision matching algorithm identifies the specific agent and the exact hand responsible for the contact, utilizing the spatial features extracted in the first stage. These identified 3D hand coordinates serve as precise sub-trajectory boundary conditions. Finally, we employ an optimization algorithm governed by aerodynamic physics to lift the 2D ball detections into a coherent 3D trajectory. To rigorously validate our physics-grounded lifting against state-of-the-art methods, we conduct extensive ablation studies on a 4 corner cameras, multi-view, semi-professional dataset for which we extract ground-truth 3D ball trajectories. Our evaluations demonstrate that anchoring physical equations to player kinematics outperforms existing monocular methods~\cite{liu2022monotrack}, which rely on 2D reprojection for initial conditions estimation and consequently suffer from important 3D spatial drift.

\vspace{0.3em}
In summary, our main contributions are as follows:
\begin{itemize}
    \item \textbf{A comprehensive monocular 3D reconstruction framework} tailored to broadcast volleyball, which unifies player and 3D ball tracking into a spatiotemporally coherent scene without relying on 3D training datasets.
    \item \textbf{A physics-informed jump handling method} that resolves the depth drift caused by ground-plane projection models. By combining event spotting with Newtonian interpolation, our approach yields visually and physically accurate 3D airborne player localizations.
    \item \textbf{A context-aware 3D ball trajectory lifting pipeline}, introducing a sub-trajectory segmentation architecture, a training-free collision matching algorithm, and a player-anchored aerodynamic solver. Evaluations across a calibrated multi-view benchmark validate that grounding physical equations in player kinematics reduces 3D spatial drift compared to 2D-reprojection baselines.
\end{itemize}
\vspace{-0.9em}

\section{Related Work}
\label{sec:sota}

\textbf{3D Multi-View Reconstruction in Sports.} The 3D reconstruction of sports matches has been a prominent computer vision research vector for over two decades. Early exploratory works in the 2000s primarily focused on virtualizing soccer broadcast scenes \cite{2000soccer, 2003soccer, 2004soccer}. Since then, multi-camera architectures have achieved high-fidelity results leveraging advanced stereo segmentation~\cite{stereo} and multi-person triangulation~\cite{Bridgeman_2019_CVPR_Workshops, triang}. While highly accurate, these systems require meticulously calibrated and synchronized camera networks, limiting their applicability to standard, single-feed volleyball broadcast footage. Consequently, recent efforts have shifted toward monocular reconstruction, which fundamentally relies on strong spatial priors to resolve the ill-posed nature of single-view depth estimation.

\textbf{Monocular 3D Reconstruction in Sports.} To mitigate single-view ambiguities, recent monocular frameworks targeted constrained environments like badminton~\cite{lin2023vird} and table tennis~\cite{gossard2025tt3d}. These sports offer strict spatial separation between players, static cameras, and alternating turns. Such structured settings allow methods to leverage fixed viewpoints and strict planar ground assumptions, effectively bypassing mutual occlusions and simplifying 3D projectile tracking. However, these systems rely heavily on domain-specific heuristics (e.g., the \textit{MonoTrack}~\cite{liu2022monotrack} method for shuttlecocks) and fail to handle jump phases, as they strictly anchor players to the ground plane.

\textbf{Monocular 3D Reconstruction in Collective Sports.} Transitioning to collective sports exposes the vulnerabilities of classical monocular pipelines due to continuous mutual occlusions and unpredictably moving broadcast cameras. Pioneering works like \textit{Soccer on Your Tabletop}~\cite{rematas2018soccer} bridged this gap using 2D poses and massive synthetic datasets to generate individual depth maps, though the resulting volumetric approximations often suffer from topological artifacts such as disappearing player body parts. Similar frameworks for American football~\cite{Sawafuji_2026_CVPR} face operational bottlenecks, requiring manual tracking initialization and relying on frame-by-frame NLF~\cite{sarandi2024neural} reconstructions that lack temporal consistency. Crucially, these collective sport pipelines do not support two major requirements for accurate volleyball 3D reconstruction: they entirely ignore the ball's 3D trajectory, losing essential tactical context, and they remain rigidly bound to the planar ground assumption, failing to capture high-amplitude jumps.

To resolve these ambiguities, recent state-of-the-art methods have reverted to multi-view setups. For instance, \textit{LiveSplats}~\cite{livesplats} applies 3D Gaussian Splatting across synchronized streams for basketball reconstruction. However, such dense camera networks preclude applicability to standard broadcast footage, underscoring the need for monocular frameworks capable of holistically capturing high-amplitude jumps and ball-player dynamics.

\textbf{Monocular 3D Ball Trajectory Estimation.} To reconstruct the 3D trajectory of a fast-moving projectile from a single camera feed, classical frameworks heavily leverage physics-based optimization by fitting aerodynamic equations to 2D image observations. These methods typically segment the continuous video into discrete sub-trajectories, subsequently optimizing the initial 3D position and velocity for each interval~\cite{chen2012, shen2016, dong2021, liu2022monotrack, Grad_2026_CVPR}. Recent frameworks like \textit{MonoTrack}~\cite{liu2022monotrack} and Grad \textit{et al.}'s~\cite{Grad_2026_CVPR} attempt to bound this non-linear optimization by minimizing a 2D reprojection loss to estimate these initial conditions. However, relying on 2D reprojection is fundamentally flawed for depth estimation, as it inherently yields severe spatial drift on real-world datasets. Furthermore, this reprojection mechanism strictly requires an accurate 3D camera calibration matrix. \textit{MonoTrack}~\cite{liu2022monotrack} achieves this by relying on the explicit detection of the four court corners and two specific reference points on the net (from a corner view). In profile broadcast volleyball, net reference points are frequently visually ambiguous, occluded, or completely out of frame, making stable 3D calibration exceedingly difficult and inevitably leading to sub-optimal trajectory lifting. Additionally, \textit{MonoTrack}~\cite{liu2022monotrack} relies heavily on the inherent simplifications of badminton dynamics, where the hitting player alternates deterministically across the net, entirely bypassing the challenge of identifying which specific athlete is responsible for the ball deflection. It also presumes a static camera and assumes the projectile never contacts the ground. Similarly, while Grad \textit{et al.}~\cite{Grad_2026_CVPR} introduce robust aerodynamic fitting pipelines over 2D projections, their methodology requires manual temporal segmentation, enforces rigid ground-verticality heuristics that fail during aerial rallies, and remains highly sensitive to monocular calibration noise.

Parallel to explicit physical modeling, alternative frameworks explore geometric and deep learning methods to directly lift 2D ball detections into 3D space~\cite{van20223d, Ponglertnapakorn_2025_CVPR, kienzle2026uplifting}. While Zandycke \textit{et al.}~\cite{van20223d} propose an ingenious scale-based ray projection framework, its reconstruction accuracy remains sub-optimal as it processes frames independently without leveraging temporal video dynamics while still requiring a fully calibrated camera. Alternatively, learning-based methods directly map pixel tracks to 3D coordinates~\cite{Ponglertnapakorn_2025_CVPR, kienzle2026uplifting}. However, due to the severe scarcity of real-world 3D ball annotations, their heavy reliance on synthetic training data introduces a pronounced domain gap that hinders generalization on broadcast footage.

To overcome these limitations, we propose a comprehensive, modular framework for the holistic 3D reconstruction of volleyball broadcast sequences. Unlike existing collective sports pipelines that strictly rely on planar ground assumptions and ignore ball dynamics, our approach introduces a physics-informed jump handling module to correct jump-induced perspective drift. By coupling temporal event spotting with Newtonian interpolation, we natively resolve the breakdown of projective models during high-amplitude aerial phases, yielding physically accurate 3D player localizations. Furthermore, to circumvent the lack of 3D ball supervision and the spatial drift caused by 2D-reprojection optimizations, we propose a context-aware 3D ball trajectory lifting pipeline. After introducing a temporal sub-trajectory segmentation architecture and a heuristic collision matching algorithm, we anchor the ball's aerodynamic optimization solver to the reconstructed 3D hand vertices. By using player kinematics as precise boundary conditions, our approach completely bypasses the reliance on noise-sensitive 3D camera calibration, unifying athlete and ball dynamics into a spatiotemporally coherent 3D scene.

\section{Global Pipeline Overview}

\label{sec:pipeline}

The proposed end-to-end framework, illustrated in Figure~\ref{fig:pipeline}, is designed to map raw monocular broadcast videos onto a pre-modeled 3D volleyball court. To circumvent the severe lack of multi-agent 3D ground truth, our framework is decoupled into two interconnected stages. Stage 1 establishes the structural foundation by extracting and positioning the athletes in 3D space, explicitly handling the transition between ground-contact and aerial phases. Building upon this, Stage 2 leverages the reconstructed players kinematics to constrain and estimate the 3D trajectory of the ball.

\subsection{Player Reconstruction}

To establish the global spatial context, athletes must first be localized and tracked across the temporal sequence. We employ YOLOX~\cite{ge2021yolox} coupled with Deep-EIoU~\cite{huang2024iterative} for multi-object tracking. 

To register these tracked 2D bounding boxes within the global 3D court space, we continuously compute the homography matrix between the 2D image plane and the court ground plane. This is achieved utilizing the court keypoint detection architecture proposed by Maglo et al.~\cite{maglo2023}. Assuming a strict ground-contact constraint, players' 2D foot coordinates are projected through this homography to obtain absolute world-space positions. However, extracting coordinates via 2D homographic projections from a moving camera introduces high-frequency spatial noise. To enforce temporal stability prior to 3D lifting, we introduce a filtering module applying a Butterworth low-pass filter to the projected 2D ground-plane trajectories, mitigating projection jitter (as shown on~\cref{fig:perspective_drift_ablation}). Simultaneously, we extract the full-body 3D kinematics of the athletes using WHAM~\cite{shin2024wham} to generate camera-space 3D poses. 

Fundamentally, the ground-plane projection of the player feet structurally fails during high-amplitude jumps. To resolve this, we introduce a physics-informed jump handling module. Because raw pixel-based event spotting struggles to isolate individual actions within a densely populated scene, we opt for PoseConvSpot~\cite{poseconv} a pose-based architecture. This discrete temporal isolation allows the pipeline to explicitly override the planar projection. By applying Newtonian equations of motion between the homography-derived takeoff and landing positions, we ensure physically accurate 3D mesh positioning during these aerial phases.

\subsection{Context-Aware Ball 3D Trajectory Estimation}

The second stage reconstructs the projectile dynamics, abandoning standalone 3D trajectory optimizations in favor of a context-aware anchoring approach. Initially, 2D ball tracking is performed using TrackNetV3~\cite{tracknetv3}, which proved to be highly accurate for localizing small, high-velocity, and motion-blurred projectiles in sports feeds. Given the discontinuous nature of volleyball trajectories, these global 2D tracks are subsequently segmented into discrete sub-trajectories with our introduced architecture.

Instead of relying on unstable calibration-based 2D-reprojection optimizations to lift these tracklets, we explicitly anchor the ball to the reconstructed environment. A training-free, heuristic-based collision matching algorithm is deployed to identify the agent, either a player or the court surface (assuming no net contacts in this study), and the exact hand responsible for initiating each sub-trajectory. By retrieving the corrected 3D kinematics generated in the first stage, these identified 3D hand vertices establish precise boundary conditions. Finally, we employ an aerodynamic optimization solver to lift the 2D detections into a continuous 3D trajectory. By natively bounding these physical equations between the verified player interaction points, our approach completely bypasses the need for explicit 3D camera calibration, seamlessly unifying the athletes and the ball into a single, spatiotemporally coherent representation.
\section{Jump-Aware 3D Player Reconstruction}
\label{sec:jump_reconstruction}

As highlighted in Section~\ref{sec:pipeline}, standard monocular 3D positioning heavily relies on a strict ground-contact assumption. However, volleyball is defined by frequent, high-amplitude jumps. During these airborne phases, the 2D homographic projection fundamentally breaks down. When a player jumps, their elevation in the 2D image plane is misinterpreted by the homography matrix as a retreat in depth, causing severe perspective drift along the court's plane. To resolve this, we propose to decouple the problem into two tasks: precisely isolating the temporal interval of each jump, and physically correcting the 3D trajectory during this interval.

\textbf{Precise Jump Spotting.} Extracting frame-accurate micro-events in unconstrained broadcast footage is highly challenging. Traditional RGB-based pipelines struggle with dense multi-player scenarios, where spatial overlaps and mutual occlusions severely degrade pixel-level feature extraction. To circumvent this, we utilize PoseConvSpot~\cite{poseconv}, a purely pose-based temporal event spotting architecture. Relying exclusively on 2D joint coordinates, already extracted during our initial tracking phase, makes this approach computationally efficient and robust to visual overlaps that plague crowded broadcast sports while being more robust to domain shifts~\cite{poseconv}. This module allows us to reliably identify the exact discrete takeoff ($t=0$) and landing ($t=t_f$) frames for each jumping athlete.

\begin{figure}[t]
  \centering
  \includegraphics[width=\columnwidth]{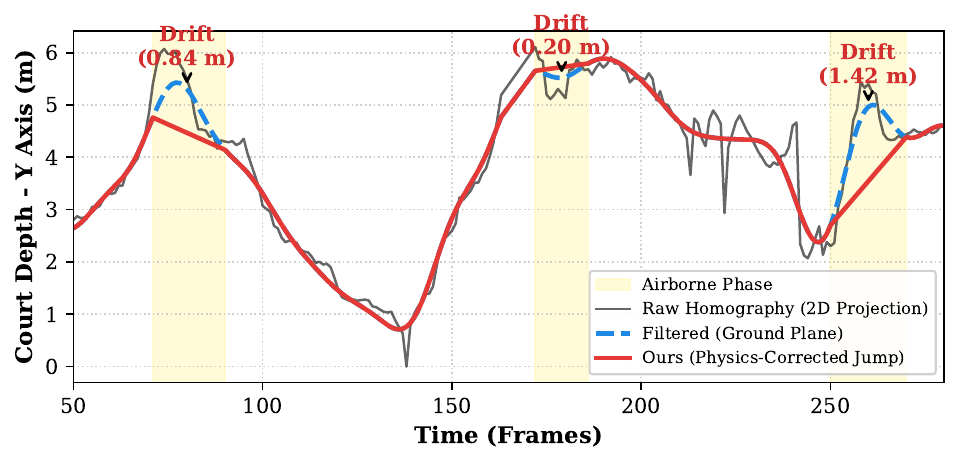}
  \caption{\textbf{ButterWorth Filtering and Perspective Drift Correction .} The raw 2D homographic projection of a player's position during a sequence (grey) exhibits high-frequency jitter. While temporal filtering (blue dashed line) stabilizes ground locomotion, it fails during airborne phases (yellow zones). Our physics-informed interpolation (solid red line) overrides the ground-plane.}
  \label{fig:perspective_drift_ablation}
  \vspace{-1.5em}
\end{figure}

\textbf{Airborne Trajectory Correction.} Once the exact temporal boundaries of a jump are isolated, we reconstruct the player's true 3D trajectory by overriding the flawed homographic projection. The corresponding world-space boundary coordinates at takeoff, $(x_0, y_0, 0)$, and landing, $(x_f, y_f, 0)$, are obtained from the ground-plane homography . Integrating the acceleration twice over the interval $t \in [0, t_f]$ with our ground-contact boundary conditions yields the corrected spatiotemporal trajectory:
\begin{equation}
    \mathbf{p}_{xy}(t) = \mathbf{p}_0 + \frac{t}{t_f}(\mathbf{p}_f - \mathbf{p}_0), \quad z(t) = \frac{g}{2}t(t_f - t)
\end{equation}
where $g$ is the gravitational acceleration ($9.81 \text{ m/s}^2$).
\Cref{fig:perspective_drift_ablation} shows the effect of Butterworth filtering and jumping player position correction during jumps.
\vspace{-0.6em}

\section{Scene-Constrained Ball Trajectory Lifting}
\label{sec:ball_lifting}

Unlike racquet sports where projectile dynamics are strictly alternating and highly predictable, a volleyball rally is characterized by various possible trajectories such as intra-team passes, spikes, and bumps, which are often discontinuous. To reconstruct this complex 3D ball motion without relying on 3D training data or noise-sensitive 2D reprojection, we introduce a scene-constrained lifting approach that anchors the ball's physical equations to the reconstructed athletes.

\subsection{2D Ball Sub-Trajectory Segmentation}
\label{subsec:ball_segmentation}


The first step requires isolating the global 2D ball track, extracted continuously via TrackNetV3~\cite{tracknetv3}, into discrete ballistic sub-trajectories separated by deflection points (e.g., hits, passes, or bounces). To extract these frame-accurate micro-events, we propose a lightweight Temporal Convolutional Network (TCN).

For a given temporal window of length $T$, the network takes a 4-channel input sequence consisting of the ball's 2D image coordinates and their first-order temporal derivatives: $X_t = [x_t, y_t, dx_t, dy_t]^T$, and outputs a probability vector of length $T$, where each element represents the likelihood of the corresponding frame containing a deflection point. The architecture processes this sequence through an initial 1D convolution, followed by a series of TCN blocks leveraging dilated convolutions. These dilated residual blocks efficiently expand the network's receptive field to capture broad spatiotemporal context. More details of the architecture in Supplementary.

We formulate sub-trajectory segmentation as continuous per-frame regression rather than frame-wise classification, adopting the formulation and \textit{Soft-Gamma Loss} of PoseConvSpot~\cite{poseconv} for human event spotting, we frame deflection detection as a regression task over Gaussian-smoothed temporal labels. 


\subsection{Heuristics-Based Collision Agent Matching}
\label{subsec:agent_prediction}

Once the 2D track is segmented into discrete sub-trajectories, each deflection point must be associated with its physical source. Bypassing data-hungry networks that lack annotated training data, we propose a training-free heuristic matching algorithm.

At every identified hit frame $t_{hit}$, the algorithm evaluates the spatial and kinematic relationship between the 2D ball detection and the skeletal joints of tracked athletes. Specifically, we compute a collision score $S_{j, k}$ for joint $j$ of player $k$, combining spatial proximity and localized joint velocity:
\begin{equation}
    S_{j, k} = \beta \exp\left(-\frac{d_{j, k}^2}{2\sigma^2}\right) + \lambda V_{j, k}
\end{equation}
where $d_{j, k}$ is the Euclidean pixel distance between the ball and wrist node $j$ of player $k$, $\sigma$ controls the Gaussian proximity decay, and $V_{j, k}$ is the wrist velocity computed via finite differences over a temporal window ($t_{hit} \pm \tau$). 

The athlete with the highest wrist score is identified as the collision agent, with the left/right score margin determining the contacting hand(s). If an ankle-to-wrist proximity ratio indicates the ball is closer to the ground than player wrists, the event is classified as a court bounce.

\subsection{Physics-Informed 3D Trajectory Estimation}
\label{subsec:balltraj_estimation}

Existing monocular lifting methods attempt to resolve depth by incorporating 2D reprojection losses into non-linear trajectory optimizations. However, this assumes pristine 3D camera calibration matrices and 2D Ball Tracking. Thus, it suffers from severe spatial drift in difficult-to-calibrate broadcast scenarios. Our framework bypasses this depth ambiguity by framing 3D trajectory lifting as a strictly bounded Boundary Value Problem (BVP).

For any identified sub-trajectory spanning between consecutive deflection frames $t_i$ and $t_{i+1}$, the collision agents provide the 3D interaction points. By retrieving the precise world-space coordinates of the hitting hand at $t_i$ (denoted $\mathbf{P}_i = (x_i, y_i, z_i)$) and the receiving hand or court ground at $t_{i+1}$ (denoted $\mathbf{P}_{i+1}$), we establish rigid initial and final boundary conditions. 

The flight of a volleyball is governed by gravity and aerodynamic drag. Neglecting the Magnus effect, the velocity vector $\mathbf{v}(t) = [v_x(t), v_y(t), v_z(t)]^T$ is governed by the following first-order differential equation:
\begin{equation}
    \frac{\mathrm{d}\mathbf{v}(t)}{\mathrm{d}t} = \mathbf{g} - C_d \|\mathbf{v}(t)\| \mathbf{v}(t)
\end{equation}
where $\mathbf{g} = [0, 0, -9.81]^T \text{ m/s}^2$ and $C_d = \frac{\rho A C_{\text{drag}}}{2m}$ is the drag-to-mass coefficient (computed to $0.035 \text{ m}^{-1}$ using standard volley ball measurements).

Since the trajectory must connect $\mathbf{P}_i$ and $\mathbf{P}_{i+1}$ over the time interval $\Delta t = t_{i+1} - t_i$, the only unknown variable is the initial velocity vector $\mathbf{v}_i$ for this specific segment. Rather than relying on 2D reprojection heuristics, we directly solve for the exact $\mathbf{v}_i$ that satisfies these physical constraints. We define an optimization problem to minimize the Euclidean endpoint error:
\begin{equation}
    \mathbf{v}_i^* = \arg\min_{\mathbf{v}_i} \left\| \hat{\mathbf{P}}_{i+1}(\mathbf{v}_i) - \mathbf{P}_{i+1} \right\|^2
\end{equation}
where $\hat{\mathbf{P}}_{i+1}(\mathbf{v}_i)$ is the simulated final position obtained by numerically integrating the differential equation over $\Delta t$. The optimization is efficiently solved using the Nelder-Mead algorithm. The solver is initialized with the closed-form pure ballistic velocity (assuming zero drag, $C_d = 0$).

Once $\mathbf{v}_i^*$ is optimized, the 3D position of the ball at any intermediate frame $t \in [t_i, t_{i+1}]$ is computed using Semi-Implicit Euler Method. We rigorously validate the superiority of this physics-grounded lifting approach over standard 2D-reprojection optimizations in~\cref{subsec:balltraj_estimation}.
\section{Experiments and Results}
\label{sec:experiments}

While qualitative assessment of our pipeline can be performed through visual inspection of the reconstructions (see~\cref{sec:qualitative}), obtaining a quantitative evaluation is more challenging due to the lack of broadcast video recordings with calibrated
cameras, synchronized human motion capture and 3D ball
tracking. Therefore, we evaluate the individual components
of our pipeline to the best of our ability.

\subsection{Experimental Setup}
\label{subsec:exp_setup}

To train and evaluate our modules, we leverage established sports benchmarks and extensive novel manual annotations.

\vspace{0.3em}
\noindent \textbf{Broadcast Volleyball Dataset.} Because our pipeline is specifically designed for unconstrained broadcast environments, we substantially extend the Volleyball dataset by Saad \textit{et al.}~\cite{saad}. We manually annotated multiple subsets while maintaining the original cross-match splits:
\begin{itemize}
    \item \textbf{Court Calibration:} We manually annotated 4 to 9 visible court keypoints in 1,500 diverse frames to robustly evaluate homography estimation.
    \item \textbf{Perspective Drift:} 50 challenging aerial phases were annotated with takeoff/landing foot coordinates and human estimation of their orthogonal ground projections to evaluate spatial drift mitigation.
    \item \textbf{Trajectory Segmentation and Agent Collision Matching:} We labeled over 1,600 trajectory-change events across various arenas to train, evaluate and compare segmentation modules. We also included SportsMOT~\cite{cui2023sportsmot} videos to have a wider range of courts.
\end{itemize}
For 2D ball tracking, we utilized the existing annotations provided by Perez et al.~\cite{PEREZ2022108360} to train TrackNetV3~\cite{tracknetv3}. To foster future research in volleyball analytics, all our annotations will be made publicly available.

\vspace{0.3em}
\noindent \textbf{VolliQ 3D Validation Dataset.} While our pipeline is designed for uncalibrated moving broadcast cameras, rigorously evaluating the core 3D physics-informed lifting mechanism against existing 2D-reprojection baselines requires 3D ground truth. As none exists as far as we know, to avoid the domain gap introduced by synthetic datasets, we utilized VolliQ~\cite{volliq}, a semi-professional dataset featuring four synchronized, static cameras positioned at the court corners. We semi-manually annotated the exact 2D ball coordinates across 10 continuous, fully active rallies (yielding over 12,000 bounding boxes across 3,000 multi-view frames). These views were triangulated, after cameras 3D calibration, to establish a real-world 3D ballistic ground truth for our ablation studies.


\vspace{0.3em}
\noindent \textbf{Implementation Details.} 
Extensive training protocols, hyperparameter configurations for all tasks and dataset details are deferred to the Supplementary Material.

\begin{table}[t]
  \centering
  \footnotesize 
  \renewcommand{\arraystretch}{0.95}

  \begin{subtable}{\columnwidth}
    \centering
    \setlength{\tabcolsep}{5.5pt}
    \begin{tabular}{l ccc}
      \toprule
      \textbf{Backbone} & \textbf{Mean IoU (\%)} & \textbf{Proj. Err (m)} & \textbf{Reproj. Err (px)} \\
      \midrule
      ViT-Tiny~\cite{wu2022tinyvitfastpretrainingdistillation} & 96.03 & 0.181 & 0.009 \\
      PVTv2~\cite{Wang_2022}    & \textbf{96.67} & \textbf{0.144} & \textbf{0.007} \\
      \bottomrule
    \end{tabular}
    \caption{Performance of court keypoint detection backbones.}
    \label{tab:calib}
  \end{subtable}

  \vspace{0.4em}

  \begin{subtable}{\columnwidth}
    \centering
    \setlength{\tabcolsep}{15pt} 
    \begin{tabular}{l cc}
      \toprule
      \textbf{Method} & \textbf{MAE (m)} & \textbf{RMSE (m)} \\
      \midrule
      Standard Homography     & 1.050 & 1.431 \\
      Ours (Physics-Informed) & \textbf{0.230} & \textbf{0.412} \\
      \bottomrule
    \end{tabular}
    \caption{Perspective drift mitigation along court depth ($Y$-axis).}
    \label{tab:drift}
    \vspace{-0.6em}
  \end{subtable}

  \caption{\textbf{Quantitative Evaluation of 3D Player Reconstruction Components.}}
  \vspace{-1.6em}
  \label{tab:player_reconstruction}
\end{table}





\begin{table}[t]
  \centering
  \small 
  \renewcommand{\arraystretch}{0.95} 
  \setlength{\tabcolsep}{5.5pt}      
  \begin{tabular}{l ccc}
    \toprule
    \textbf{Method} & \textbf{$\delta = 1$} & \textbf{$\delta = 2$} & \textbf{$\delta = 3$} \\
    \midrule
    \multicolumn{4}{c}{\textit{Using Oracle 2D Ball Trajectory}} \\
    \midrule
    Baseline 1 ($Y(t)$)                            & 0.801 & 0.840 & 0.843 \\
    Baseline 2 ($X(t) + Y(t)$)                     & 0.794 & 0.872 & 0.884 \\
    TT3D (Poly. only)~\cite{gossard2025tt3d}       & 0.713 & 0.751 & 0.764 \\
    TT3D (Poly. + PCA Blur)~\cite{gossard2025tt3d} & 0.762 & 0.807 & 0.810 \\
    HitNet (MonoTrack)~\cite{liu2022monotrack}      & 0.836 & 0.874 & 0.877 \\
    \textbf{Ours (TCN + Soft-Focal)}               & \textbf{0.929} & \textbf{0.945} & \textbf{0.950} \\
    \midrule
    \multicolumn{4}{c}{\textit{Using Predicted 2D Ball Trajectory}} \\
    \midrule
    Heuristic $Y(t)$                               & 0.759 & 0.802 & 0.812 \\
    Heuristic $X(t) + Y(t)$                        & 0.714 & 0.794 & 0.813 \\
    TT3D (Poly. only)~\cite{gossard2025tt3d}       & 0.689 & 0.725 & 0.736 \\
    TT3D (Poly. + PCA Blur)~\cite{gossard2025tt3d} & 0.702 & 0.751 & 0.770 \\
    HitNet (MonoTrack)~\cite{liu2022monotrack}      & 0.815 & 0.858 & 0.866 \\
    \textbf{Ours (TCN + Soft-Focal)}               & \textbf{0.898} & \textbf{0.914} & \textbf{0.925} \\
    \bottomrule
  \end{tabular}
  \vspace{-0.4em}
  \caption{\textbf{2D Ball Trajectory Segmentation (F1-Score).} Performance across temporal tolerances $\delta \in \{1, 2, 3\}$ frames.}
  \vspace{-2.3em} 
  \label{tab:segmentation}
\end{table}

\subsection{3D Player Reconstruction}
\label{subsec:exp_player_reconstruction}

\noindent \textbf{Court Calibration.} To compute the continuous image-to-court homography, we adapt the keypoint detection pipeline of Maglo \textit{et al.}~\cite{s23187938} to volleyball. Following the metrics in Chu \textit{et al.}~\cite{chu2022}, we compare the lightweight ViT-Tiny~\cite{wu2022tinyvitfastpretrainingdistillation} (original backbone) against a Pyramid Vision Transformer (PVTv2)~\cite{Wang_2022}. As shown in \cref{tab:calib}, PVTv2 achieves higher spatial precision at the expense of increased inference time.

\vspace{0.3em}
\noindent \textbf{Jump-Aware Trajectory Correction.} We evaluate aerial perspective drift mitigation along court depth ($Y$-axis) across 50 annotated jumps ($>800$ airborne frames). As shown in \cref{tab:drift}, our physics-informed correction significantly reduces drift compared to standard homography. Qualitative results show that remaining residual errors stem primarily from human annotation ambiguity in estimating orthogonal ground projections from monocular profile views, rather than algorithmic failure.

\begin{table}[t]
  \centering
  \small
  \setlength{\tabcolsep}{1.6pt} 
  \begin{tabular}{l ccc}
    \toprule
    {\footnotesize \textbf{Method}} & {\footnotesize \textbf{3D RMSE\,(m)}} & {\footnotesize \textbf{X/Y/Z RMSE\,(m)}} & {\footnotesize \textbf{2D Reproj.\,(px)}} \\
    \midrule
    MonoTrack~\cite{liu2022monotrack} & 1.56 & 1.42 / 0.62 / \textbf{0.12} & \textbf{3.89} \\
    \midrule
    \multicolumn{4}{c}{\textit{\footnotesize Theoretical Bounds (Using 3D GT Start/End Points)}} \\
    \midrule
    Pure Ballistics    & 0.38 & 0.28 / 0.10 / 0.23 & 19.2 \\
    With Drag          & 0.29 & \textbf{0.14} / \textbf{0.09} / 0.24 & 17.9 \\
    \midrule
    \multicolumn{4}{c}{\textit{\footnotesize Our Pipeline (Using 3D Hand Bounds)}} \\
    \midrule
    Pure Ballistics    & 0.38 & 0.31 / 0.14 / 0.18 & 19.4 \\
    \textbf{With Drag} & \textbf{0.27} & 0.15 / 0.13 / 0.18 & 17.2 \\
    \bottomrule
  \end{tabular}
  \vspace{-0.2em}
  \caption{\textbf{3D Ball Trajectory Estimation on VolliQ.} Evaluation of 3D spatial accuracy and 2D reprojection error. Our holistic physics-informed approach drastically reduces 3D spatial drift compared to reprojection-based optimization.}
  \label{tab:lab_validation}
  \vspace{-1.0em}
\end{table}

\begin{figure}[t]
  \centering
  \includegraphics[width=0.8\columnwidth]{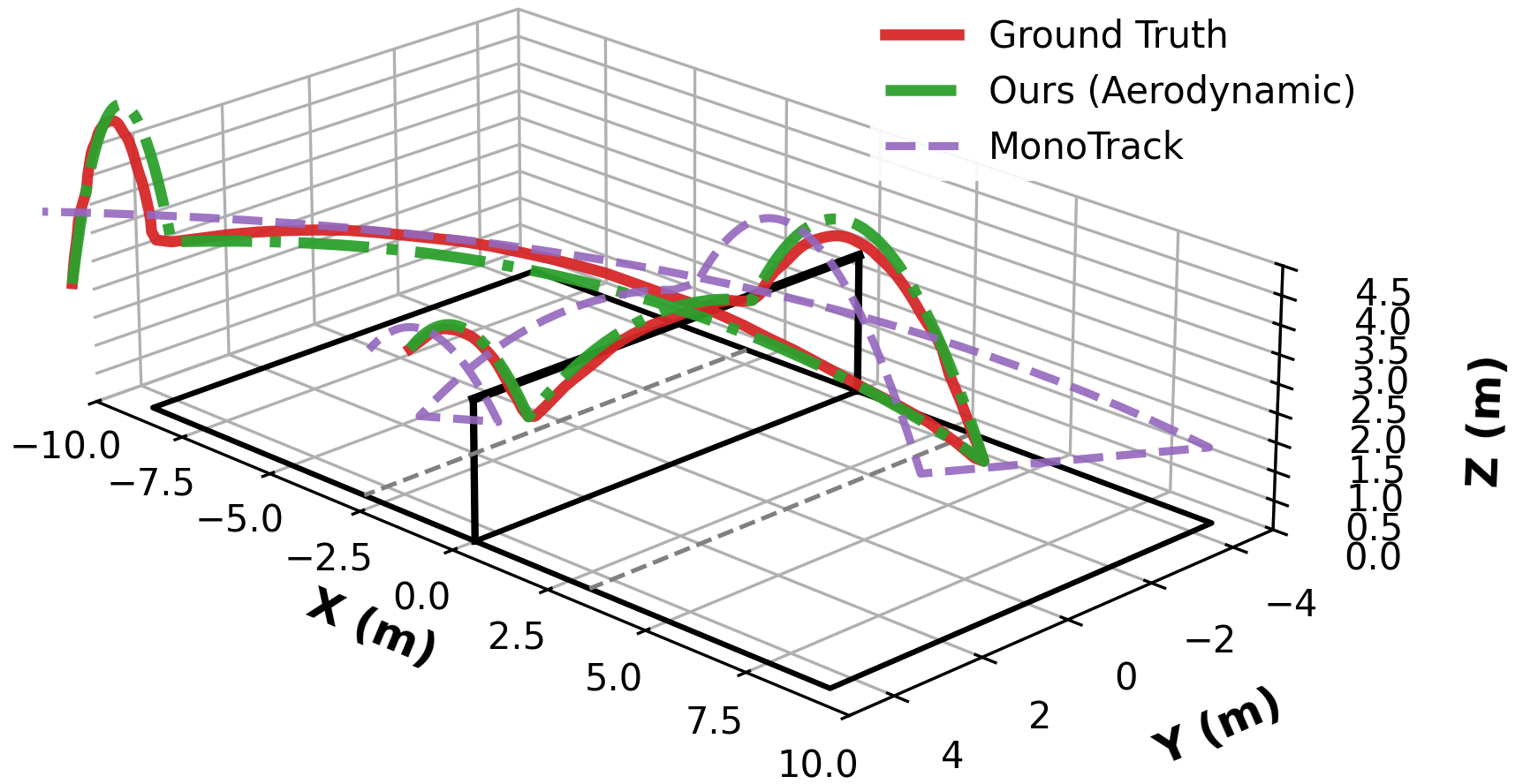}
  \caption{\textbf{Qualitative comparison of 3D trajectory reconstruction on a short rally from Volliq~\cite{volliq}.}  Our drag model (green, err: \textbf{0.19\,m, 12.7 px}) accurately follows the ground truth (red) compared to MonoTrack~\cite{liu2022monotrack} (purple, err: 1.69\,m, 4.32 px).}
  \label{fig:trajectory_qualitative}
  \vspace{-1.6em}
\end{figure}

\subsection{3D Ball Trajectory Estimation}
\label{subsec:exp_ball_trajectory}

\begin{figure*}[t]
  \centering
  \includegraphics[width=0.243\textwidth]{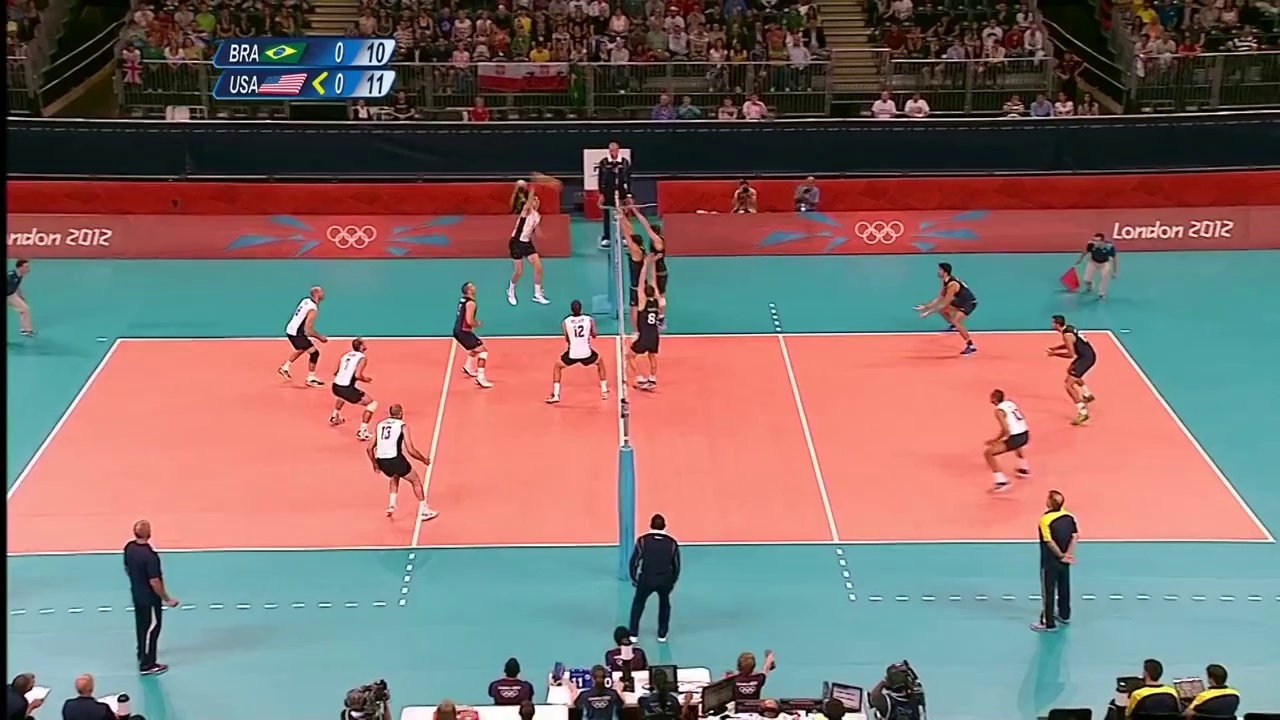}\hfill
  \includegraphics[width=0.243\textwidth]{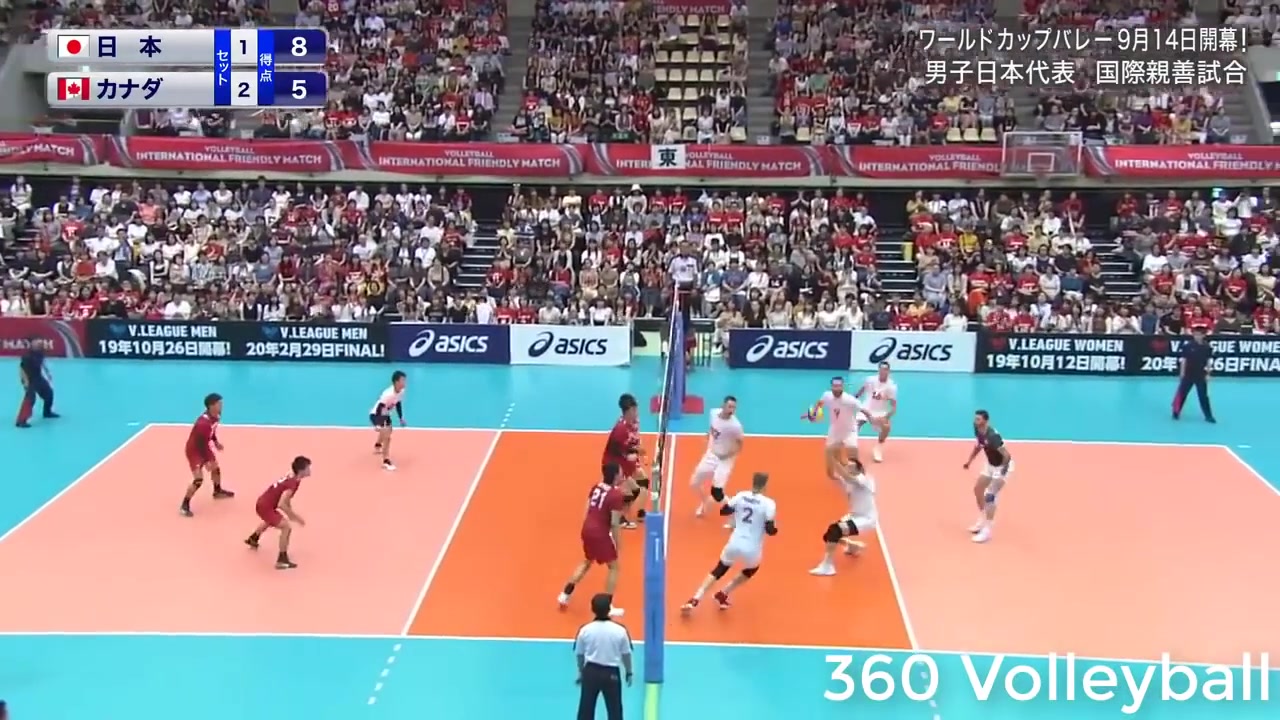}\hfill
  \includegraphics[width=0.243\textwidth]{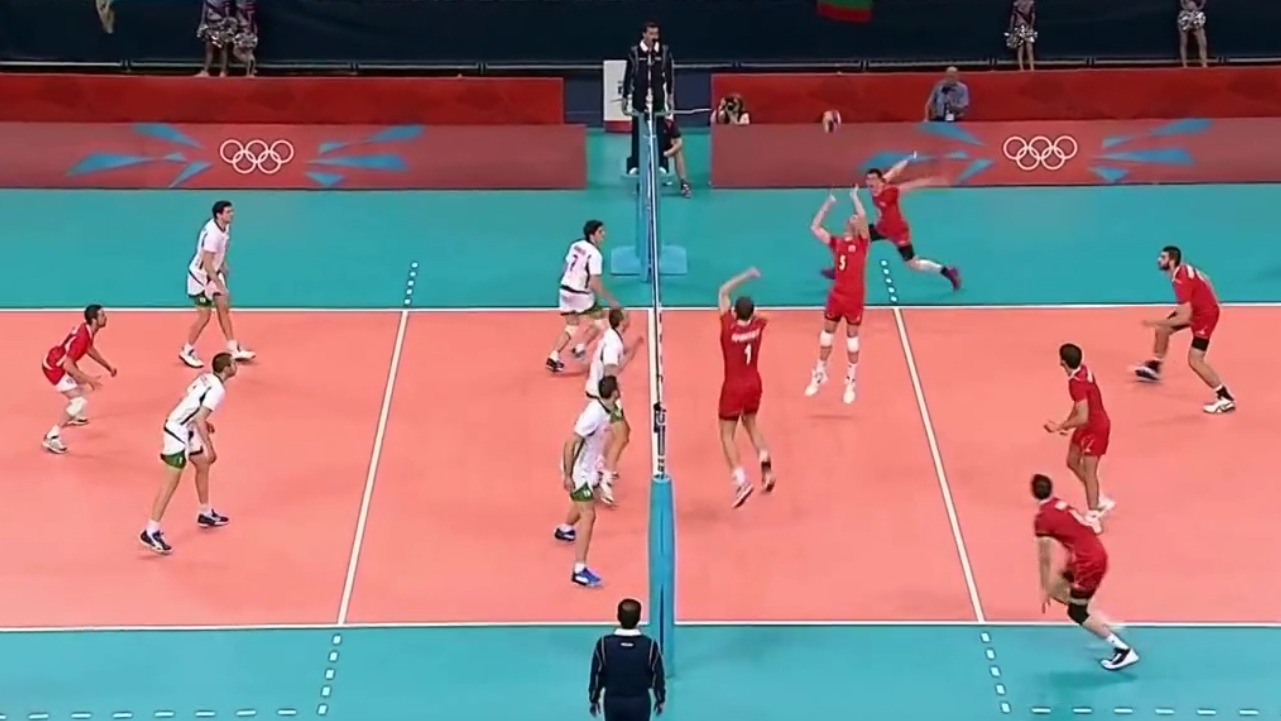}\hfill
  \includegraphics[width=0.243\textwidth]{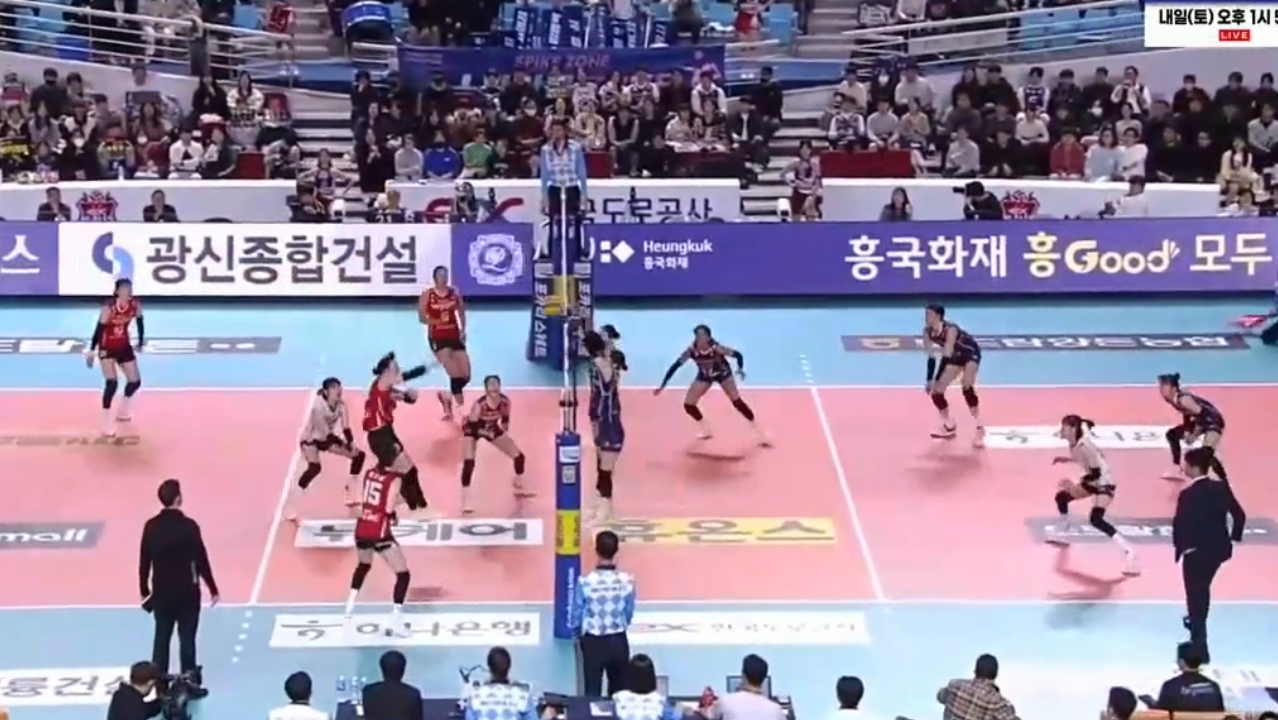}


  \includegraphics[width=0.243\textwidth]{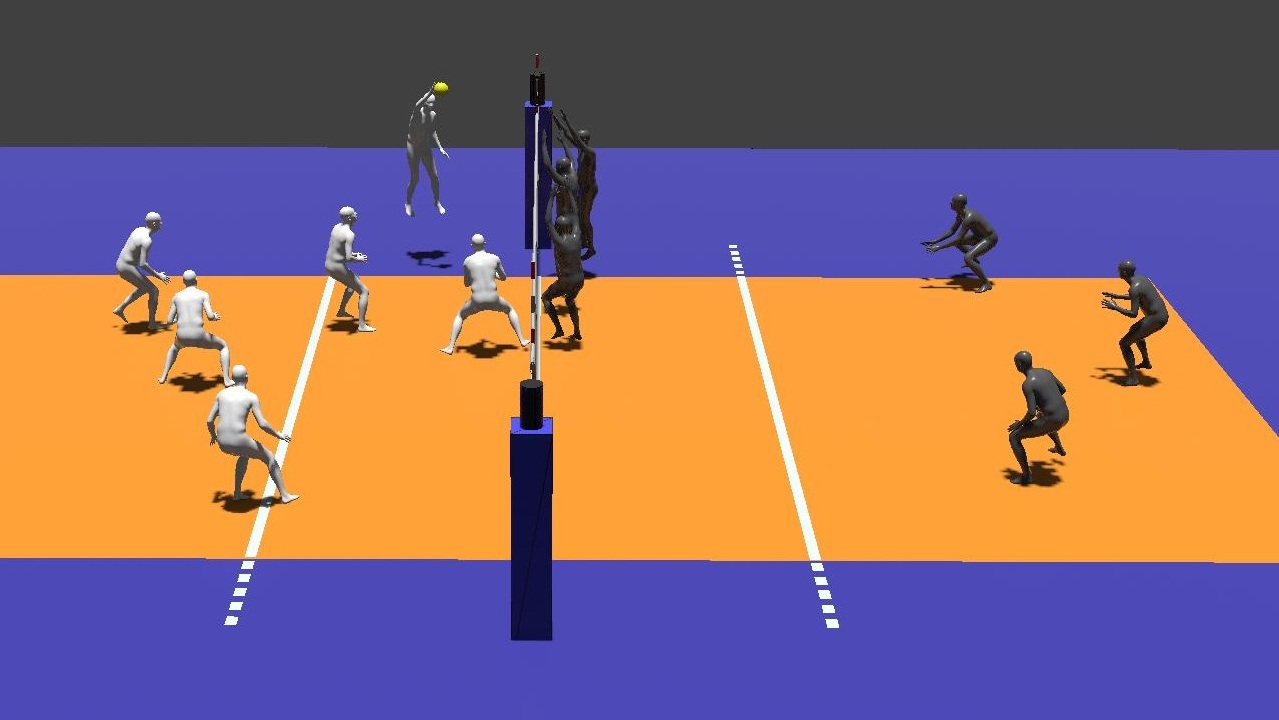}\hfill
  \includegraphics[width=0.243\textwidth]{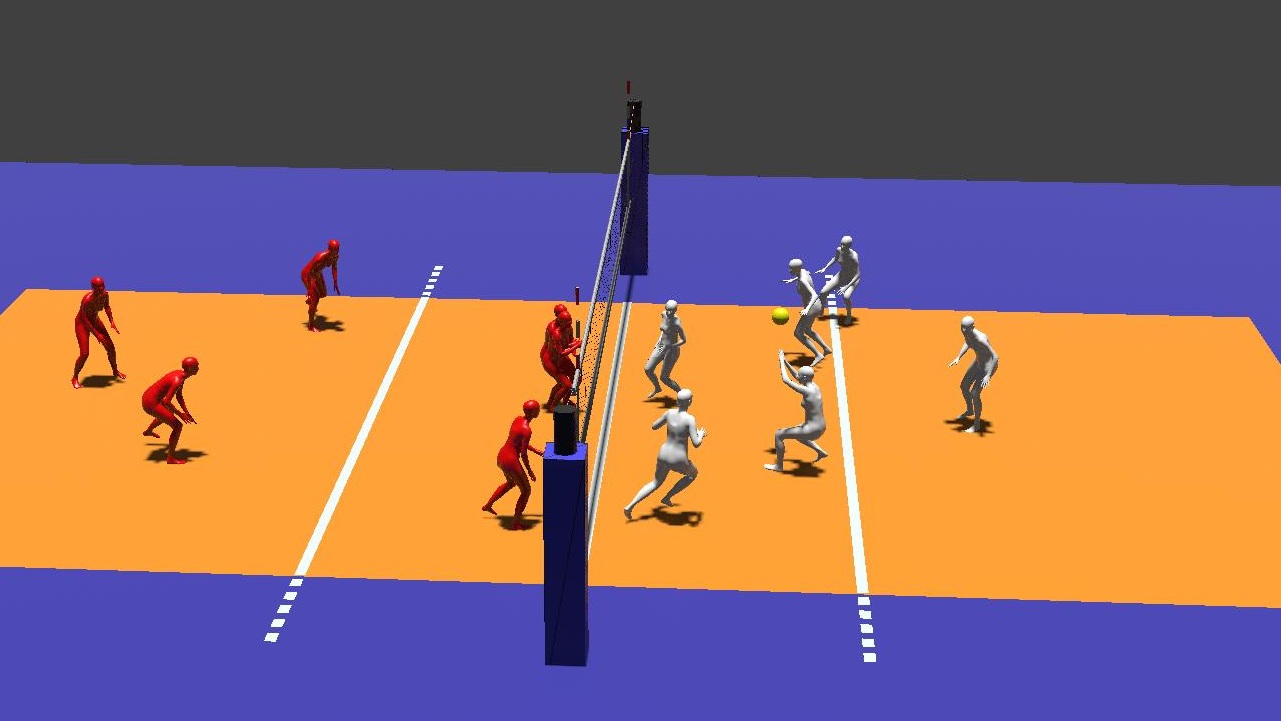}\hfill
  \includegraphics[width=0.243\textwidth]{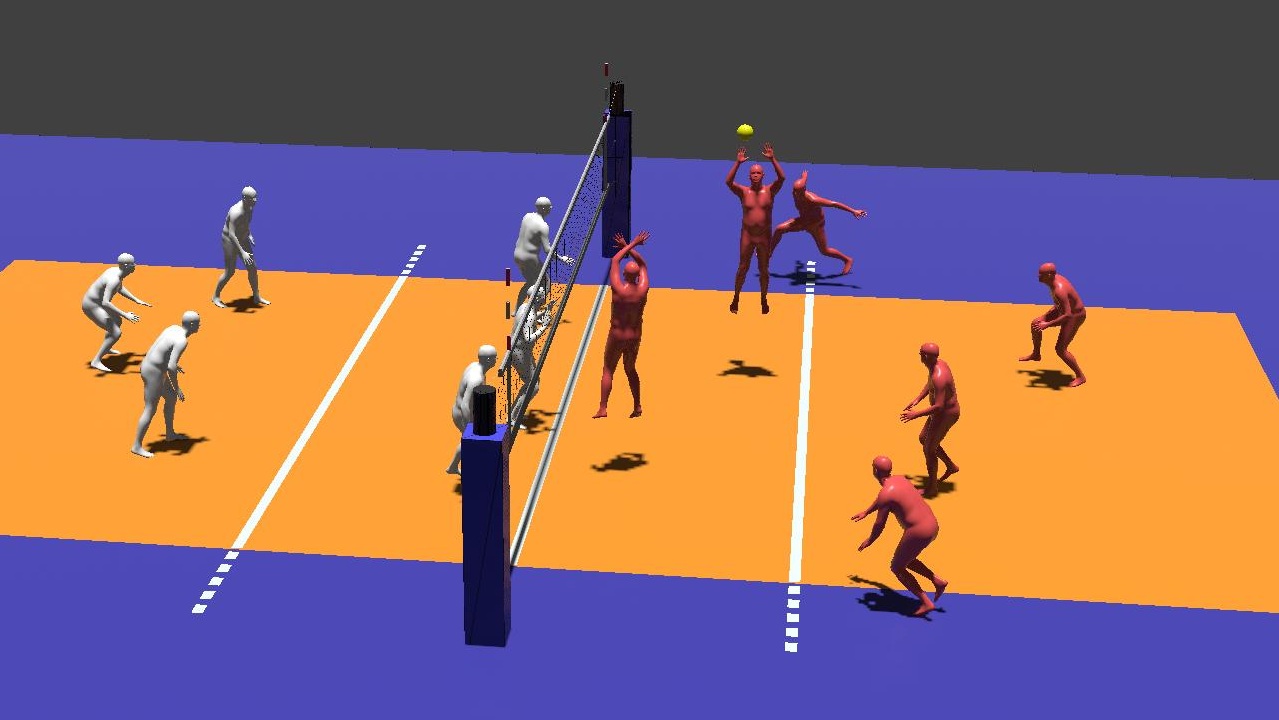}\hfill
  \includegraphics[width=0.243\textwidth]{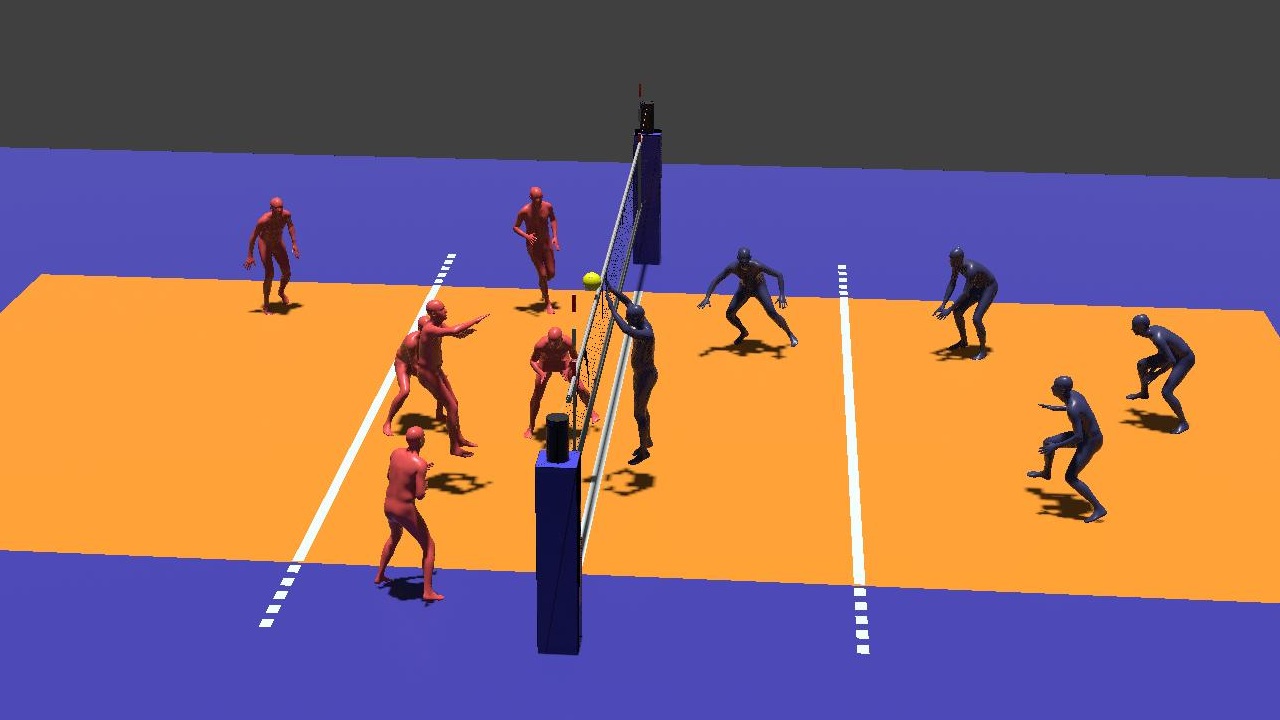}\

  \caption{\textbf{Qualitative 3D reconstruction results across diverse game scenarios.} Top: 2D input broadcast frames from different rallies. Bottom: corresponding unified 3D scene reconstruction.}
  \label{fig:qualitative_results}
  \vspace{-1.5em}
\end{figure*}

\noindent \textbf{2D Ball Tracking and Trajectory Segmentation.} We benchmark our segmentation module against heuristic and learned baselines using the F1-Score across three temporal tolerances ($\delta \in \{1, 2, 3\}$ frames). To decouple segmentation from upstream 2D detection errors, we evaluate on both Oracle and TrackNetV3 predicted trajectories.

We compare our module to four baselines. \textbf{Heuristics $\mathbf{Y(t)}$} and \textbf{$\mathbf{X(t) + Y(t)}$} geometrically detect deflections via gradient sign changes in the ball's 2D pixel coordinates. We also adapt \textbf{TT3D Piecewise Polynomials~\cite{gossard2025tt3d}} in two variants: a standard solver and one augmented with PCA-based motion blur, bypassing its native BlurBall~\cite{gossard2026blurball} module due to the lack of a volleyball-specific blur dataset. Finally, we evaluate HitNet (MonoTrack)~\cite{liu2022monotrack}, a data-driven racquet sports model. Because HitNet requires the hitter's pose, an ambiguous input in volleyball before collision matching, we restrict its inputs to 2D ball tracks and court corners.

As demonstrated in \cref{tab:segmentation}, our TCN architecture significantly outperforms all baselines. 
Moreover, ~\cref{tab:segmentation} reveals that standard heuristics are highly sensitive to 2D tracking noise, whereas our network maintains a robust 0.898 F1-Score ($\delta=1$) even on raw TrackNetV3 predictions.

\vspace{0.3em}

\noindent \textbf{Heuristic Collision Agent Matching.} We evaluate our spatial proximity matching (\cref{subsec:agent_prediction}) using three metrics: (1) \textit{Agent Identification} (player ID or ground); (2) \textit{Strict Hand Assignment} (left, right, or both hands); and (3) a relaxed \textit{Functional Hand Assignment}. In volleyball, two-handed contacts (e.g., setting, bump, block) bring the hands together. Because our 3D solver anchors two-handed hits to a single reference hand vertex (right) without physical bias, the functional metric only penalizes strictly lateral hand assignment errors. On Oracle 2D trajectories, our algorithm achieves \textbf{85.3\% agent identification accuracy}, with 98.8\% functional and 75.8\% strict hand assignment. On noisy ball predictions, performance remains remarkably robust, with agent identification dropping only slightly to 83.3\%.

\vspace{0.3em}
\noindent \textbf{In-The-Lab Validation: Physics-Based Lifting.} We validate our holistic lifting \textbf{methodology} using our triangulated ball 3D ground truth of the multi-camera VolliQ dataset~\cite{volliq}. While VolliQ's static cameras do not assess our broadcast-specific tracking algorithms, they isolate and validate the physical coherence of our 3D formulation against state-of-the-art 2D-reprojection methods.

As shown in~\cref{tab:lab_validation} and~\cref{fig:trajectory_qualitative}, MonoTrack~\cite{liu2022monotrack} overfits to the image plane: achieving minimal 2D reprojection error ($3.9\text{px}$) but severe depth ambiguity ($1.56\text{m}$ 3D RMSE). Conversely, our aerodynamic pipeline prioritizes spatial coherence, slashing the 3D RMSE to $0.272\text{m}$ and ensuring physically sound trajectories consistent with ball dynamics. We ablate our predicted ``GT Hand Bounds'' (triangulated hands) against theoretical ``GT Bounds'' (GT start/end 3D ball coordinates). Surprisingly, anchoring to 3D hands outperforms GT ball bounds ($0.272\text{m}$ vs. $0.293\text{m}$). We attribute this to multi-frame contacts: because we assume instantaneous hits, exact GT ball positions cause simulated trajectories to depart prematurely. Anchoring to hand vertices, which sit lower than the ball during contact, geometrically compensates for this temporal mismatch, as confirmed by the vertical $Z$-axis error drop ($0.182\text{m}$ vs. $0.242\text{m}$).

\vspace{-0.3em}
\section{Qualitative experiments}
\label{sec:qualitative}
\vspace{-0.2em}

\Cref{fig:teaser,fig:qualitative_results} showcase our pipeline across dynamic plays (spikes, sets, blocks). Despite the depth ambiguity inherent to monocular broadcast feeds, our framework reliably recovers the global 3D layout of most players, with physics-informed jump spotting lifting jumping actors to plausible heights. While extreme 2D overlap from rare camera alignments can occasionally conceal players (e.g., the missed blocker in the rightmost column), the reconstructed ball remains consistent with game dynamics, demonstrating robust spatiotemporal coherence. More examples are in Supplementary Materials.

\vspace{-0.3em}
\section{Conclusion and Future Work}
\label{sec:conclusion}
\vspace{-0.2em}



We presented Volley3D, a unified framework for lifting uncalibrated monocular broadcast volleyball footage into a spatiotemporally coherent 3D scene. Our pipeline first corrects ground-plane projection flaws by coupling pose-based jump spotting with Newtonian interpolation, yielding physically accurate 3D localizations for airborne players. Then, by introducing a TCN-based trajectory segmentation architecture with a collision matching algorithm, our method anchors the ball’s aerodynamic equations directly to the athletes’ reconstructed 3D hands, resulting in physically plausible 3D reconstructions of complex rallies.

Future work includes an end-to-end evaluation of the complete pipeline using a new dataset pairing broadcast camera views with synchronized 3D ground truth, as well as adapting the pipeline for real-time processing.

{
    \small
    \bibliographystyle{ieeenat_fullname}
    \bibliography{main}
}

\clearpage

\maketitlesupplementary 
\appendix

\renewcommand{\thesection}{\Alph{section}}
\renewcommand{\thefigure}{S\arabic{figure}}
\renewcommand{\thetable}{S\arabic{table}}
\renewcommand{\theequation}{S\arabic{equation}}
\setcounter{section}{0}
\setcounter{figure}{0}
\setcounter{table}{0}
\setcounter{equation}{0}

\section{Datasets Details}
\label{sec:supp_datasets}

This section provides comprehensive details regarding the datasets used for training and evaluating each module of our pipeline, as well as our custom annotation procedures.

\subsection{Players Reconstruction Modules}

\paragraph{Detection and Tracking} For 2D multi-object detection and tracking, we directly employ the YOLOX~\cite{ge2021yolox} detection model and the Deep-EIoU~\cite{huang2024iterative} re-identification tracker pre-trained on the SportsMOT dataset~\cite{cui2023sportsmot}. No further fine-tuning was performed for these modules.

\paragraph{Court Calibration} Despite prior works addressing court calibration, there is, to the best of our knowledge, no publicly available annotated dataset dedicated to volleyball. To train our keypoints detection network (from which we obtain homography estimation between the 2D image space and court plane), we manually annotated broadcast volleyball frames from the Volleyball Dataset~\cite{saad}. For each frame, we annotated 4 to 9 keypoints of the court to obtain a robust GT homography. As fish-eye limitations can cause the homography to fail, we restrained from annotating points at the extreme limit of the image. The annotated frames details are summarized in \cref{tab:supp_calib_dataset}. For splitting matches, we strictly follow the original dataset splitting~\cite{saad}.

\begin{table}[h]
  \centering
  \small
  \begin{tabular}{l c c}
    \toprule
    \textbf{Split} & \textbf{\# Frames} & \textbf{Avg. Points / Frame} \\
    \midrule
    Train & 923 & 5.64 \\
    Validation & 350 & 5.32 \\
    Test & 164 & 4.53 \\
    \bottomrule
  \end{tabular}
  \vspace{-0.5em}
  \caption{\textbf{Court Calibration Dataset.} We report the number of annotated frames per split and the average number of visible court keypoints annotated per frame.}
  \label{tab:supp_calib_dataset}
\end{table}

\paragraph{Human Mesh Recovery (WHAM)}
For 3D human pose and shape estimation, we utilize the WHAM codebase, 2D body joints are extracted using the ViTPose-Base model~\cite{xu2022vitpose} and are subsequently processed by the official WHAM~\cite{shin2024wham} model, using the official checkpoint trained on the BEDLAM dataset provided by the authors.

\paragraph{Jump Detection}
Our jump spotting module relies on the PoseConvSpot architecture~\cite{poseconv}. We utilize the official model checkpoint trained on the SportsJumpMotion dataset~\cite{Oo_2025_CVPR}, ensuring accurate temporal localization of aerial phases without any further adaptation.

\subsection{Ball Lifting}
\label{subsec:lifting}

\paragraph{2D Ball Tracking.}
We train TrackNetV3's full pipeline~\cite{tracknetv3} (TrackNet and InPaintNet) using the annotations provided by Perez \etal~\cite{PEREZ2022108360} on the Volleyball dataset originally introduced by Saad \etal~\cite{saad}. We adapted the training configuration as detailed in \cref{sec:supp_implementation}. A comparison of our 2D tracking performance against the state-of-the-art results reported directly by Tarashima \etal ~\cite{tarashima2023widelyapplicablestrongbaseline} on this same dataset (marked with $\dagger$) is provided in \cref{sec:supp_2d_tracking_results}.

\paragraph{Trajectory Segmentation and Agent Collision.}
The original dataset from Saad \etal~\cite{saad} consists of short, overlapping, 41-frame video clips. Because our temporal convolutional network (TCN) requires a receptive field of 17 frames, applying it to such short clips introduces severe boundary effects. To mitigate this, we merged consecutive clips belonging to the same rally into continuous sequences. To further increase visual diversity, we also incorporated several volleyball rallies from SportsMOT~\cite{cui2023sportsmot}, for which we added semi-automatically the 2D ball positions. We then manually annotated \textbf{some} of these combined sequences for both sub-trajectory temporal boundaries and collision agent matching. More precisely, we annotate deflection frames, Player ID (or 'ground'/'net') and used hands. \Cref{tab:supp_annotations_stats} details the dataset statistics before and after the merging process, alongside the total number of annotations.

\begin{table}[h]
  \centering
  \small
  \setlength{\tabcolsep}{6.0pt}
  \begin{tabular}{l ccc ccc}
    \toprule
    & \multicolumn{3}{c}{\textbf{Original~\cite{saad}}} & \multicolumn{3}{c}{\textbf{Merged (Ours)}} \\
    \cmidrule(lr){2-4} \cmidrule(lr){5-7}
    \textbf{Split} & \textbf{Clips} & \textbf{Frames} & \textbf{Avg.} & \textbf{Clips} & \textbf{Frames} & \textbf{Avg.} \\
    \midrule
    Train & 2,152 & 88,232 & 41.0 & 805 & 69,320 & 86.1 \\
    Val   & 1,341 & 54,981 & 41.0 & 541 & 43,331 & 80.1 \\
    Test  & 1,337 & 54,817 & 41.0 & 548 & 43,628 & 79.6 \\
    \midrule
    Total & 4,830 & 198,030 & 41.0 & 1,894 & 156,279 & 82.5 \\
    \bottomrule
  \end{tabular}
  \vspace{-0.5em}
  \caption{\textbf{Trajectory Dataset Merging Statistics.} Comparison between the original 41-frame video clips from Saad \etal~\cite{saad} and our continuous merged rally sequences.}
  \label{tab:supp_merging_stats}
\end{table}

\begin{table}[h]
  \centering
  \small
  \setlength{\tabcolsep}{6pt}
  \begin{tabular}{l ccc}
    \toprule
    \textbf{Split} & \textbf{Annotated Clips} & \textbf{Sub-trajectories} & \textbf{Collisions} \\
    \midrule
    Train & 135 & 613 & 0 \\
    Val   & 87 & 353 & 0 \\
    Test  & 155 & 655 & 654 \\
    \midrule
    Total & 377 & 1,621 & 654 \\
    \bottomrule
  \end{tabular}
  \vspace{-0.5em}
  \caption{\textbf{Manual Trajectory and Collision Annotations.} Statistics of our custom annotations across the subset of merged rally sequences.}
  \label{tab:supp_annotations_stats}
\end{table}

\begin{figure}[h]
    \centering
    \includegraphics[width=\linewidth]{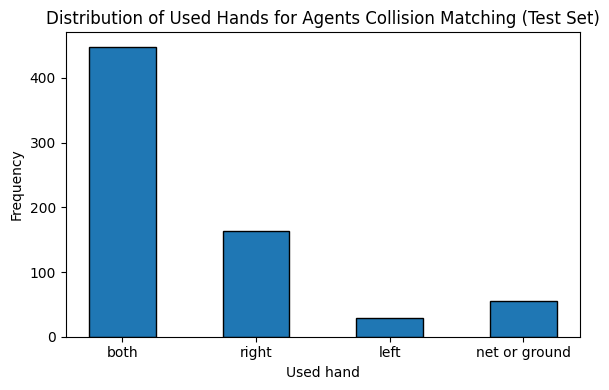}
    \caption{\textbf{Agent Collision Distribution.} Frequency of each physical source across our annotated dataset.}
    \label{fig:supp_collision_dist}
\end{figure}
The collision agent annotations are distributed among four categories: \textit{Left Hand}, \textit{Right Hand} and \textit{Both Hands} for Player assignment and \textit{Ground} or \textit{Net}. In volleyball, actions such as setting naturally involve a symmetrical, two-handed contact (\textit{Both Hands}). To simplify our heuristic 3D solver without sacrificing physical realism, we explicitly anchor these two-handed contacts to the right hand vertex of the player's 3D mesh. This design choice is functionally sound because both hands often converge for a two-hands movement (e.g. they form a tight diamond shape during a set), meaning the spatial offset between the left and right hand is negligible relative to the ball's radius and overall trajectory scale.

\subsection{VolliQ Evaluation Dataset}
\label{subsec:supp_volliq}
To benchmark our full 3D physics-informed pipeline against multi-view ground truth, we leverage \textbf{VolliQ}~\cite{volliq}, a synchronized four-camera semi-professional volleyball dataset. Because the raw data presents temporal offsets and lacks explicit 3D annotations, we established a rigorous annotation protocol:

\begin{itemize}
    \item \textbf{Temporal Synchronization:} We manually re-synchronized the multi-view streams, correcting minor frame-level discrepancies present in the original dataset.
    \item \textbf{Multi-View Camera Calibration:} Static court and net keypoints were annotated across all four views to get intrinsic and extrinsic camera matrices. This calibration serves both as the foundation for multi-view triangulation and as the reference projection for the MonoTrack~\cite{liu2022monotrack} baseline.
    \item \textbf{2D Ball Tracking:} We employed a semi-automatic pipeline to detect 2D ball positions across all four camera feeds, followed by exhaustive manual inspection and per-frame refinement to ensure zero tracking artifacts.
    \item \textbf{3D Ground-Truth Trajectories:} Using our refined multi-view 2D annotations and static calibrations, we reconstructed metric 3D ball coordinates via robust multi-view triangulation providing a rigorous benchmark for monocular trajectory lifting.
    \item \textbf{Trajectory Segmentation:} For each rally, we localized the exact deflection frames marking physical contacts.
    \item \textbf{3D Hand Ground Bounds:} At each deflection frame, we annotated the contacting 2D hand position (or ball bounce on the ground) across all four views, tracking the approximate vertex location defined in our hand mesh model. Triangulating these points yields the metric 3D ``GT Hand Bounds'' anchoring each sub-trajectory.
    \item \textbf{MonoTrack Adaptation:} Following MonoTrack~\cite{liu2022monotrack}, we annotated the hitting players' 2D foot contacts at deflection instants, projected them onto the court surface via homography, and assigned an empirical 3D contact height of $2.0\text{ m}$. These positions supervise MonoTrack's court-bounded optimization alongside its reprojection loss.
\end{itemize}

\begin{table}[h]
  \centering
  \small
  \setlength{\tabcolsep}{6pt}
  \begin{tabular}{l cccc}
    \toprule
    \textbf{Dataset} & \textbf{Rallies} & \textbf{Frames per View} & \textbf{Sub-trajectories} \\
    \midrule
    VolliQ & 10 & 3066 & 101  \\
    \bottomrule
  \end{tabular}
  \vspace{-0.5em}
  \caption{\textbf{VolliQ 3D Benchmark Statistics.} Overview of the synchronized four-camera evaluation dataset, reporting the number of rallies, duration in frames per viewpoint, total multi-view frame count, and evaluated sub-trajectories.}
  \label{tab:supp_volliq_stats}
\end{table}

\clearpage
\section{Implementation Details}
\label{sec:supp_implementation}

In this section, we provide the architectural and training details for each module of our pipeline to ensure full reproducibility.

\subsection{Tracking and Human Mesh Recovery}
For 2D multi-object detection and tracking, we strictly follow the default configurations of YOLOX~\cite{ge2021yolox} and Deep-EIoU~\cite{huang2024iterative}, utilizing their official pre-trained weights without any further fine-tuning. 

For 3D human pose and shape recovery, we adopt the official codebase of WHAM~\cite{shin2024wham}. Following their standard pipeline, we extract 2D body joints using a ViTPose-Base model~\cite{xu2022vitpose} and apply SMPLify for mesh smoothing. The WHAM model processes video inputs using the official checkpoint pre-trained on the BEDLAM dataset.

\subsection{Jump Spotting}
The jump detection module uses the PoseConvSpot~\cite{poseconv} architecture with its default hyper-parameters. During inference, we apply a probability threshold of $0.35$ to consider an event as successfully detected. Furthermore, we apply a bi-causal temporal consistency filtering strategy: a takeoff event is only retained if a corresponding landing event is detected within a physically plausible temporal window, and vice-versa.

\subsection{Continuous Court Calibration}
Because the original implementation of Maglo \etal~\cite{s23187938} is not publicly available, we fully re-implemented their approach and adapted it to volleyball court dimensions.

\begin{figure}[h]
    \centering
    \includegraphics[width=\linewidth]{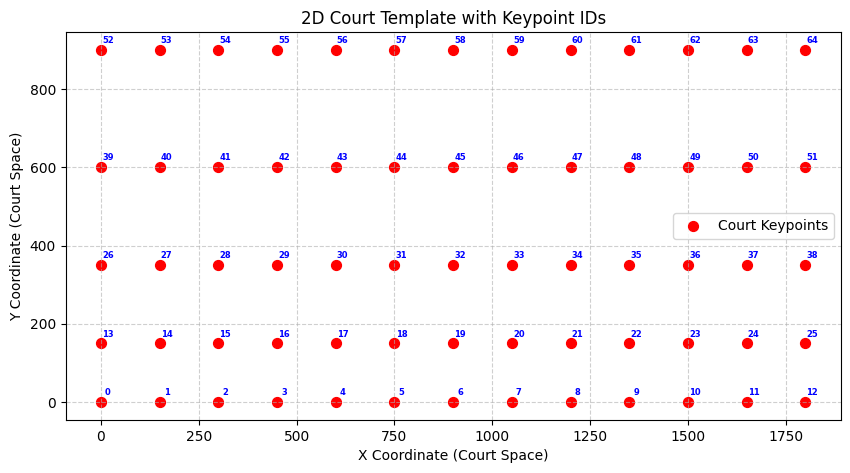}
    \caption{\textbf{2D Court Template with Keypoint IDs.} Distribution of the 65 annotated semantic keypoints defined across the metric court space ($1800\text{ cm} \times 900\text{ cm}$). The non-linear distribution along the $Y$-axis increases keypoint density near the foreground to counteract broadcast perspective foreshortening.}
    \label{fig:supp_court_template}
    \vspace{-1.1em}
\end{figure}

\paragraph{Grid Definition.}
Unlike soccer, which uses 91 keypoints in the original paper, we define a grid of 65 semantic keypoints across the volleyball court (yielding 66 classes, including the background), as illustrated in \cref{fig:supp_court_template}. Following their geometric progression logic, the grid divides the court length ($18\text{ m}$) into 13 equal segments. Along the court width ($9\text{ m}$), we distribute 5 rows of points using a geometric progression (initial step $u_0 = 150\text{ cm}$, increment $r = \text{50 cm}$). This allocates a higher density of keypoints near the baseline/lower edge of the camera view, where perspective distortion from the broadcast camera is the most pronounced.

\paragraph{Training Details.}
The backbone network is PVTv2~\cite{Wang_2022}, taking standard ImageNet-normalized inputs resized to $1280 \times 720$ pixels. Ground-truth heatmaps are generated by drawing constant-value circles with a $3$-pixel radius around each visible keypoint. We train the model for $1,000$ epochs with a batch size of 8. We optimize the network using AdamW with an initial learning rate of $10^{-4}$, decayed by a factor of 10 at two-thirds of the total training epochs via a StepLR scheduler. To handle the class imbalance between the sparse keypoints and the vast background, we use a weighted Cross-Entropy loss where keypoint pixels are assigned a weight $1,000$ times greater than the background pixels.

\paragraph{Inference.}
During inference, after extracting the 2D keypoint coordinates from the predicted heatmaps, we estimate the homography matrix using the RANSAC algorithm. We set the maximum number of RANSAC iterations to $2,000$ and the reprojection error threshold to $8$ pixels.

\paragraph{Trajectory Smoothing.}
Due to independent per-frame keypoint detections and homography estimates, raw 2D player foot locations projected onto the metric court surface can exhibit high-frequency jitter. To ensure smooth and physically coherent trajectories across time, we apply a continuous temporal filtering step. First, missing tracklet frames are linearly interpolated. Second, to eliminate edge transient artifacts, trajectories are padded by repeating boundary values by $20$ frames. Finally, we apply a $5^{\text{th}}$-order zero-phase low-pass Butterworth filter ($\text{fs} = 25\text{ Hz}$) independently to the projected $X$ and $Y$ spatial coordinates. After filtering, the artificial padding is discarded to restore the original sequence length.

\subsection{2D Ball Tracking}
\label{subsec:supp_ball_tracking}
We adopt the TrackNetV3~\cite{tracknetv3} framework composed of TrackNet (detection) and InpaintNet (temporal trajectory inpainting). Unlike the badminton baseline which concatenates a static background frame to the input, we omit background subtraction because of broadcast volleyball cameras continuous movement. Both sub-networks are trained on the annotations from Perez \etal~\cite{PEREZ2022108360} derived from the dataset by Saad \etal~\cite{saad}, with a random seed set to $26$.

\paragraph{TrackNet Details.}
TrackNet is trained for $30$ epochs with a batch size of $16$ and an input temporal window of $8$ consecutive frames. We optimize the network using the Adam optimizer with a learning rate of $10^{-3}$ and no weight decay. We employ Mixup augmentation ($\alpha = 0.5$) applied batch-wise across both input frames and target heatmaps (as formulated in the original codebase):
\begin{equation}
    \tilde{\mathbf{x}} = \lambda \mathbf{x} + (1 - \lambda) \mathbf{x}_{\pi}, \quad \tilde{\mathbf{y}} = \lambda \mathbf{y} + (1 - \lambda) \mathbf{y}_{\pi},
\end{equation}
where $\lambda = \max(\lambda', 1 - \lambda')$ with $\lambda' \sim \text{Beta}(\alpha, \alpha)$, and $\pi$ is a random batch permutation. During validation and evaluation, a predicted ball center is considered a true positive if its Euclidean distance to the ground truth is strictly within $4$ pixels.

\paragraph{InpaintNet Details.}
InpaintNet is trained for $400$ epochs with a batch size of $128$. It processes input sequences of $16$ consecutive 2D coordinates. During training, we apply a random masking ratio of $30\%$ ($p = 0.3$) over the coordinate sequence to simulate tracking failures and occlusions. 

All remaining architectural parameters follow the original TrackNetV3 configuration.

\subsection{Sub-trajectory Segmentation}
\label{subsec:supp_trajectory_segmentation}

The trajectory segmentation module relies on a 1D Temporal Convolutional Network (TCN) to temporally localize ball deflection events (e.g., hits, bounces) from the raw 2D tracking sequence. 

\paragraph{Input Format and Network Architecture}
Extracted 2D ball coordinates are converted into sequences of 4D feature vectors containing normalized spatial positions and frame-to-frame velocities: $(X, Y, dX, dY)$. Normalization is applied according to the input broadcast resolution. 

The TCN architecture begins with an initial 1D convolution (kernel size $5$) projecting the 4-channel input to $128$ channels, followed by 1D Batch Normalization, ReLU activation, and Dropout ($p=0.5$). The core temporal modeling is handled by two residual TCN blocks~\cite{lea2016temporalconvolutionalnetworksunified} (kernel size $5$). The first block maintains $128$ channels with a dilation factor of $2$, while the second expands the feature dimension to $256$ channels with a dilation factor of $4$. Each TCN block contains two dilated convolutional layers with Batch Normalization and ReLU, supplemented by a residual connection. Finally, a $1\times 1$ convolution collapses the features into a single channel, passed through a Sigmoid function to output a frame-wise probability.

\paragraph{Training Details.}
Ground truth labels are smoothed into 1D Gaussian distributions ($\sigma = 1.0$, bounded by a radius of $3$ frames) centered on the exact contact frames. We train the network for $60$ epochs using an input window size of $17$ frames and a batch size of $64$. Optimization is performed using Adam with a learning rate of $10^{-3}$ and a cosine scheduler. The Soft-Focal Loss is defined with a focusing parameter $\gamma = 2.0$ and weighting factor $\alpha = 0.25$. 

To improve model robustness, we apply four spatial and temporal augmentations during training:
\begin{itemize}
    \item \textbf{Horizontal Flipping ($p=0.5$):} The court is mirrored ($X \leftarrow 1.0 - X$ and $dX \leftarrow -dX$).
    \item \textbf{Spatial Shifting ($p=0.5$):} A random global shift $\Delta \sim \mathcal{U}(-0.08, 0.08)$ is applied to both $X$ and $Y$ coordinates.
    \item \textbf{Physical Noise ($p=0.5$):} Additive Gaussian noise $\mathcal{N}(0, 0.006)$ is injected independently into the coordinates.
    \item \textbf{Temporal Cutout ($p=0.3$):} 1 or 2 consecutive frames are randomly zero-masked within the input window to force the network to rely on global trajectory dynamics rather than overfitting to exact local peaks.
\end{itemize}

\paragraph{Inference.}
During inference, the model processes the full rally sequence using a sliding window approach with a receptive field of $17$ frames and a strict stride of $3$ frames. The frame-wise predictions are aggregated, and peaks exceeding a probability threshold of $0.35$ are successfully extracted as sub-trajectory boundaries.

\subsection{Collision Agent Matching}
\label{subsec:supp_agent_matching}
To identify the striking player, we evaluate the spatial proximity and kinematics of wrist nodes, parameterized with a Gaussian proximity decay $\sigma = 55\text{ px}$, proximity weight $\beta = 0.9$, and velocity weight $\lambda = 0.1$. The instantaneous wrist velocity $V_{j, k}$ is computed via finite differences as the maximum frame-to-frame Euclidean pixel displacement over a temporal window of $\tau = 1$ frame (i.e., $[t_{\text{hit}} - 1, t_{\text{hit}} + 1]$). Both proximity and velocity values are independently normalized by their respective maximums across all detected athletes in the frame prior to computing the composite collision score $S_{j, k}$. The athlete maximizing this score is selected as the active agent. For contact-type attribution, we evaluate the score margin between the designated athlete's wrists: if $|S_{\text{left}} - S_{\text{right}}| < 0.05$, the impact is classified as \textit{Both Hands} (associated with setting or overhead passes); otherwise, the higher-scoring wrist is designated as the sole contact source. Finally, ground bounces are detected by computing the proximity of player ankle keypoints with the same Gaussian decay parameter $\sigma$; if the ratio between the maximum ankle proximity score and the maximum wrist proximity score across all candidates satisfies $\ge 1.0$, the event is redirected to a court bounce rather than a player interaction.

\subsection{Physics-Based Optimization for 3D Trajectory Estimation}
\label{subsec:optim_supp}

To solve the bounded trajectory optimization problem formulated in Eq.~(4) of the main paper, we employ the Nelder-Mead simplex algorithm with an absolute coordinate parameter tolerance set to $10^{-5}$.

\paragraph{Initial Velocity Guess.}
We initialize the initial velocity vector $\mathbf{v}_0 = (v_{0,x}, v_{0,y}, v_{0,z})^\top$ using the closed-form ballistic solution (neglecting aerodynamic drag). Given the initial 3D contact boundary $\mathbf{p}_0 = (x_0, y_0, z_0)^\top$, the terminal boundary $\mathbf{p}_T = (x_T, y_T, z_T)^\top$, and the total trajectory duration $\Delta t = T / \text{fps}$, the closed-form initial velocity components are:
\begin{align}
    v_{0,x} &= \frac{x_T - x_0}{\Delta t}, \\
    v_{0,y} &= \frac{y_T - y_0}{\Delta t}, \\
    v_{0,z} &= \frac{z_T - z_0 + \frac{1}{2} g \Delta t^2}{\Delta t}
\end{align}
where $g = 9.81\text{ m/s}^2$ denotes gravitational acceleration.

\paragraph{Forward Numerical Integration.}
Once the optimal initial launch velocity $\hat{\mathbf{v}}_0$ is recovered by minimizing the objective function, we integrate the full 3D trajectory under quadratic aerodynamic drag ($\mathbf{a}_{\text{drag}} = -C_d \|\mathbf{v}\| \mathbf{v}$) using the semi-implicit Euler scheme:
\begin{align}
    \mathbf{v}_{t+\delta t} &= \mathbf{v}_t + \left( \mathbf{g} - C_d \|\mathbf{v}_t\| \mathbf{v}_t \right) \delta t, \\
    \mathbf{p}_{t+\delta t} &= \mathbf{p}_t + \mathbf{v}_{t+\delta t} \delta t,
\end{align}
where $\delta t = 1 / \text{fps}$, $\mathbf{g} = [0, 0, -9.81]^\top \text{ m/s}^2$, and $C_d = \frac{\rho A C_{\text{drag}}}{2m} \approx 0.035\text{ m}^{-1}$ denotes the effective drag-to-mass coefficient computed using standard volleyball specifications ($m = 0.27\text{ kg}$, $r = 0.105\text{ m}$, $A = \pi r^2$, $\rho = 1.225\text{ kg/m}^3$, and $C_{\text{drag}} \approx 0.45$).
We confirmed empirically that this $C_d$ value achieves top-results. 

\subsection{Mesh Export (Players and Ball)}
\label{subsec:supp_mesh_export}

To construct the final 3D scene, we translate and export the estimated SMPL vertices and the ball object into per-frame Wavefront \texttt{.obj} files. 

\paragraph{Player Grounding and Jump Handling.}
We first project the 2D pixel coordinates of the players' feet onto the metric court surface via the estimated inverse homography, followed by a low-pass Butterworth filter ($f_c = 1.3\text{ Hz}$) to eliminate high-frequency projection jitter. To robustly anchor the 3D SMPL meshes to the court, we introduce a \textit{Soft-Pitch} grounding heuristic. We isolate the left and right foot vertices (SMPL vertex indices 3447 and 6770) and compute a continuous blending weight using a Softmax function ($k=20$) over their local $Z$-coordinates. This smoothly assigns a higher spatial weight to the lowest foot, aiming to mitigate step-transition jitter. The entire mesh is then translated such that this hybrid reference point matches the filtered metric court coordinates at $Z=0$. For temporal windows classified as aerial jump phases, the trajectory is corrected as explained in main paper's Section 4.

\paragraph{Ball Export.}
From the continuous 3D ball coordinates generated by the Ball Lifting module, a reference volleyball mesh is automatically translated to these coordinates and exported synchronously with the player meshes.

\subsection{Blender Animation}
\label{subsec:supp_blender_animation}

\paragraph{Rendering Setup.}
To generate multi-view visualizations of the reconstructed rallies, we orchestrate a fully automated rendering pipeline using Blender's EEVEE engine. We enable Ground Truth Ambient Occlusion (GTAO) for volumetric depth, soft shadows, and Screen Space Reflections (SSR) to simulate the glossy finish of indoor volleyball courts. 

\paragraph{Scene Customization and Camera Control.}
Within the 3D environment, players are automatically clustered into their respective teams based on their spatial coordinates relative to the net, allowing for distinct, customizable material assignments for each team and the ball. To better analyze the temporal stability of the 3D reconstruction from novel vantage points, a virtual camera is programmatically animated to revolve around the court, with an adjustable angular velocity per time step. Furthermore, the pipeline adapts to the specific match category: when processing female rallies (if specified by the user), the net height is automatically lowered from the standard men's regulation height ($2.43\text{ m}$) to the women's regulation height ($2.24\text{ m}$).

\section{2D Ball Tracking Evaluation}
\label{sec:supp_2d_tracking_results}

We evaluate our 2D ball tracking module against existing small ball tracking methods on the public Volleyball dataset. Following the standard evaluation protocol, a predicted ball center is counted as a True Positive if its Euclidean distance to the ground-truth center is within a tolerance threshold of $\tau = 4\text{ pixels}$. We report the results of multiple methods on the exact same dataset from WASB~\cite{tarashima2023wasb}.

We compare the performance of TrackNetV3 (composed of TrackNet alongside InPaintNet) against baseline methods. Our implementation achieves a substantial performance improvement, reaching an Accuracy of $94.19\%$ and an F1-Score of $96.78\%$, surpassing WASB ($80.0\%$ Acc., $88.0\%$ F1) and TrackNetV2 ($73.8\%$ Acc., $83.6\%$ F1). This high precision and recall ($98.28\%$) ensure dependable 2D candidate trajectories for the downstream 3D physics lifting. We found out that the significant performance improvement over existing methods is largely due to InPaintNet.

\begin{table}[t]
  \centering
  \small
  \setlength{\tabcolsep}{6pt}
  \begin{tabular}{l cc}
    \toprule
    \textbf{Method} & \textbf{Accuracy (\%)} $\uparrow$ & \textbf{F1-Score (\%)} $\uparrow$ \\
    \midrule
    DeepBall$^\dagger$            & 50.7 & 64.4 \\
    DeepBall-Large$^\dagger$      & 57.5 & 70.4 \\
    BallSeg$^\dagger$             & 17.5 & 19.5 \\
    TrackNetV2$^\dagger$          & 73.8 & 83.6 \\
    ResTrackNetV2$^\dagger$       & 74.7 & 84.2 \\
    MonoTrack$^\dagger$           & 75.9 & 85.1 \\
    \midrule
    WASB$^\dagger$ ($\text{Step}=3$) & 77.9 & 86.5 \\
    WASB$^\dagger$ ($\text{Step}=1$) & 80.0 & 88.0 \\
    \midrule
    \textbf{TrackNetV3 (Ours)}    & \textbf{94.19} & \textbf{96.78} \\
    \bottomrule
  \end{tabular}
  \vspace{-0.5em}
  \caption{\textbf{2D Ball Tracking Performance on the Volleyball Dataset.} Evaluation with a distance threshold $\tau = 4\text{ px}$. Methods marked with ($^\dagger$) are directly reported from WASB~\cite{tarashima2023wasb}. TrackNetV3 integrates both TrackNet and InPaintNet.}
  \label{tab:supp_ball_tracking_results}
\end{table}

\section{Trajectory Lifting Ablation on VolliQ}
\label{sec:supp_lifting_ablation}

While Section~6.3 of the main paper quantitatively demonstrates the superiority of our physics-informed lifting formulation over classical 2D reprojection baselines, we provide further qualitative comparisons and error analysis against MonoTrack~\cite{liu2022monotrack} on the multi-view VolliQ dataset~\cite{volliq}.

\begin{table}[h]
  \centering
  \small
  \setlength{\tabcolsep}{1.8pt}
  \begin{tabular}{l cccc c}
    \toprule
    & \multicolumn{4}{c}{\textbf{2D Reproj. RMSE\,(px)} $\downarrow$} & \textbf{3D RMSE\,(m)} $\downarrow$ \\
    \cmidrule(lr){2-5} \cmidrule(lr){6-6}
    \textbf{Method} & \textbf{AN01} & \textbf{AN02} & \textbf{AN03} & \textbf{AN04} & \textbf{VolliQ (3D)} \\
    \midrule
    MonoTrack~\cite{liu2022monotrack} & 90.6 & \textbf{3.89} & 131.0 & 53.7 & 1.55 \\
    \midrule
    \multicolumn{6}{c}{\textit{\footnotesize Theoretical Bounds (Using 3D GT Start/End Points)}} \\
    \midrule
    Pure Ballistics   & 26.3 & 19.2 & 26.6 & 29.1 & 0.380 \\
    With Drag         & 24.9 & 17.9 & 24.5 & 26.9 & 0.293 \\
    \midrule
    \multicolumn{6}{c}{\textit{\footnotesize Our 3D Pipeline (Using Triangulated 3D Hand Bounds)}} \\
    \midrule
    Pure Ballistics   & 22.4 & 19.3 & 23.5 & 27.6 & 0.384 \\
    \textbf{With Drag (Ours)} & \textbf{19.7} & 17.2 & \textbf{19.9} & \textbf{23.1} & \textbf{0.272} \\
    \bottomrule
  \end{tabular}
  \vspace{-0.5em}
  \caption{\textbf{Cross-View Reprojection vs. Metric 3D Accuracy on VolliQ.} MonoTrack~\cite{liu2022monotrack} optimizes its trajectory via 2D reprojection on camera \texttt{AN02} (its best-performing viewpoint among all 4 views) using GT triangulated foot positions and GT camera calibration. Our physics-informed methods operate purely in metric 3D space using 3D boundary constraints. The resulting 3D trajectories are reprojected onto all four views to diagnose projective overfitting and showcase reprojection methods flaws.}
  \label{tab:supp_per_camera_ablation}
\end{table}

\begin{figure*}[p]
  \centering
  \small
  \begin{tabular}{ccc}
    \includegraphics[width=0.32\textwidth, height=3.4cm]{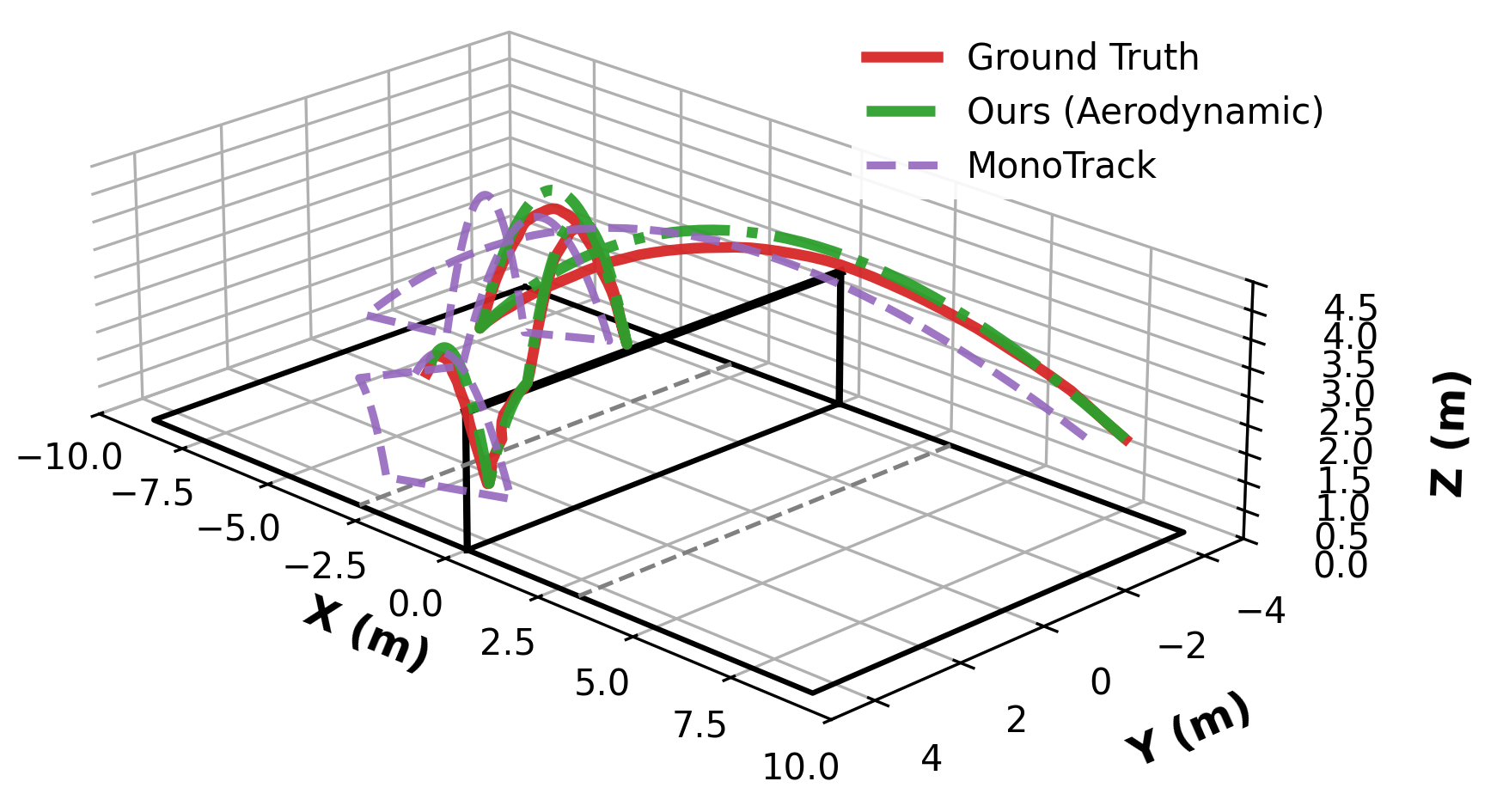} &
    \includegraphics[width=0.32\textwidth, height=3.4cm]{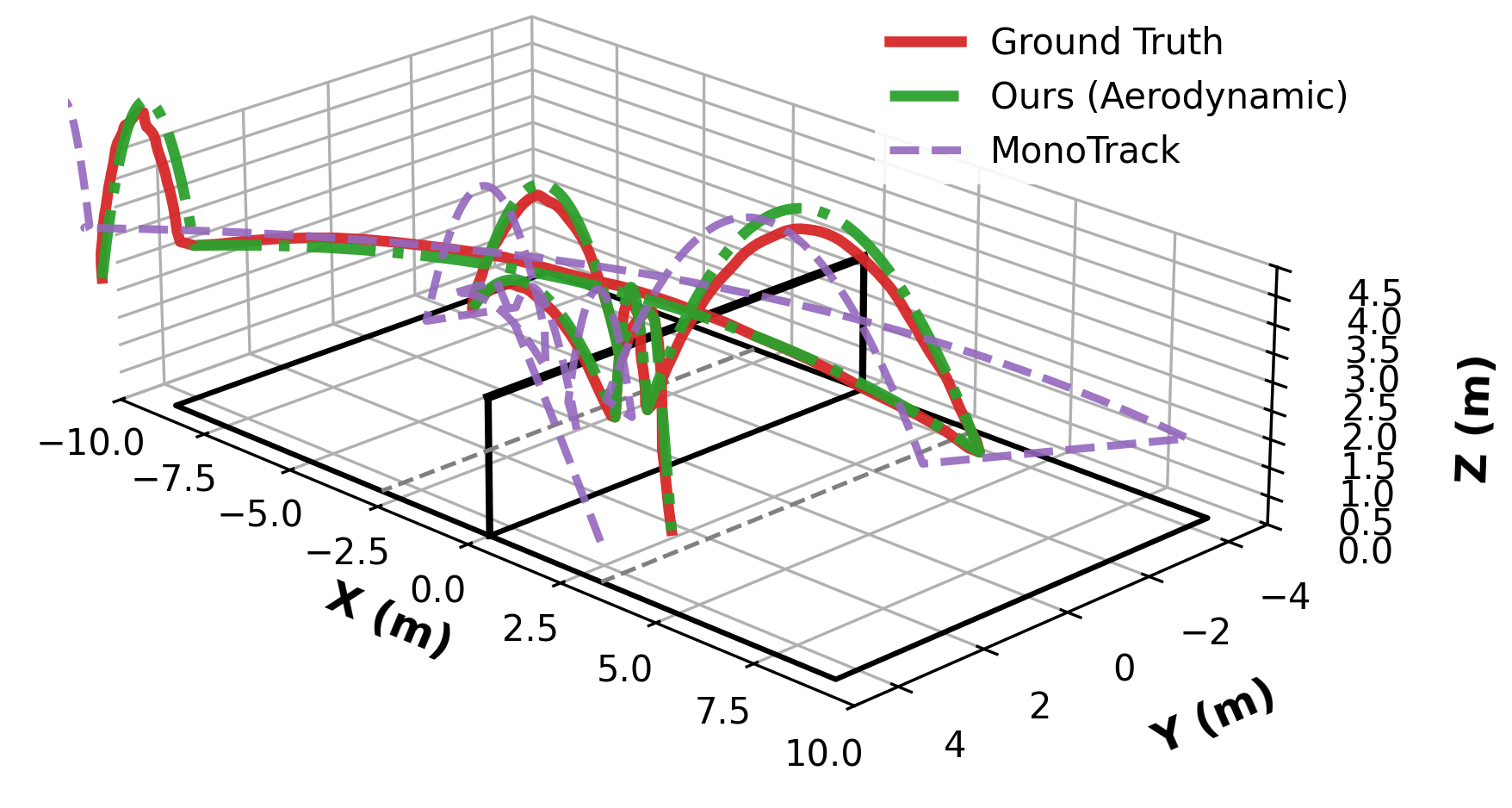} &
    \includegraphics[width=0.32\textwidth, height=3.4cm]{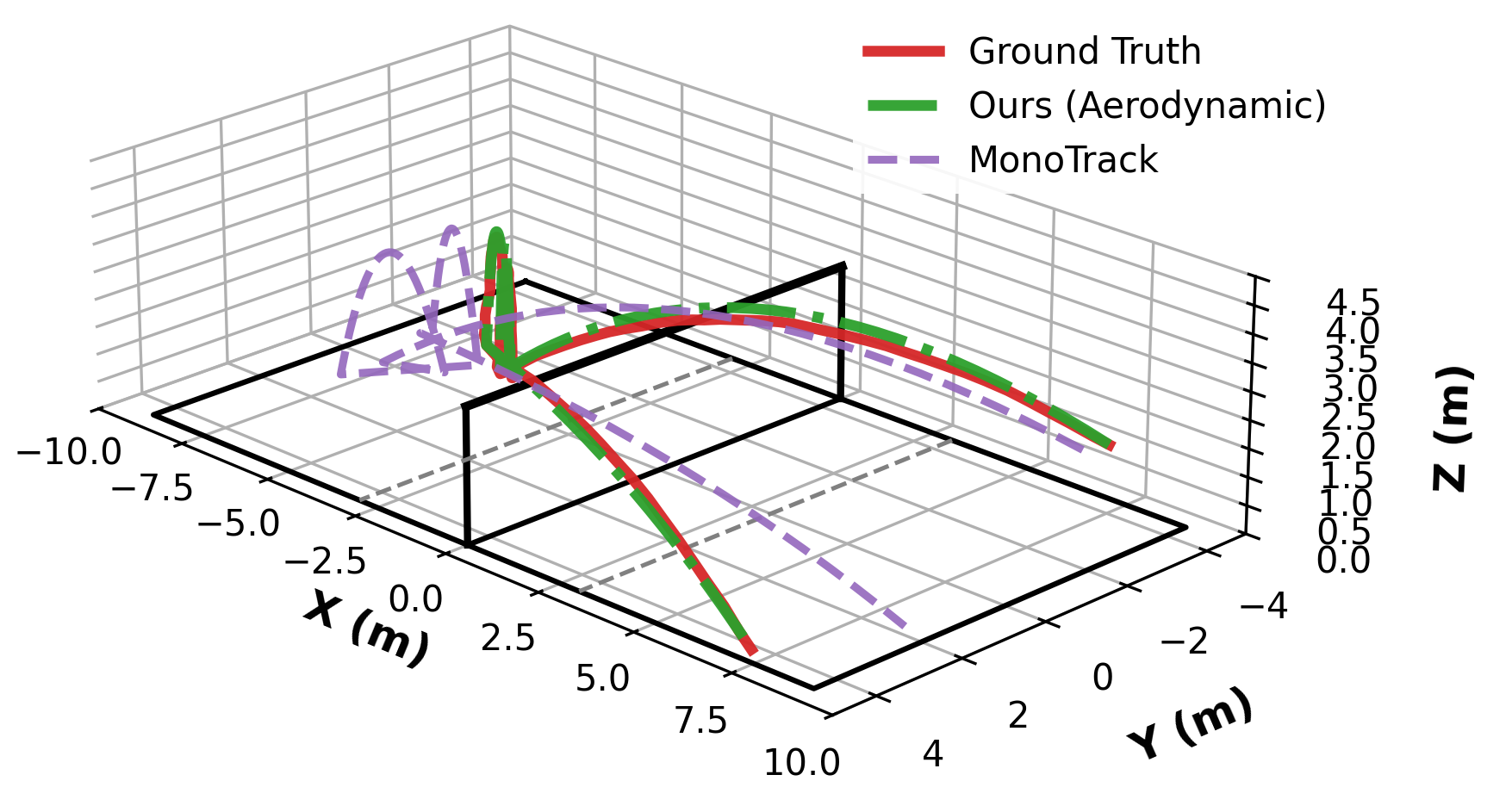} \\
    {\footnotesize (a) 3D Trajectory (rally1)} & {\footnotesize (b) 3D Trajectory (rally2)} & {\footnotesize (c) 3D Trajectory (rally3)} \\[0.4em]

    \includegraphics[width=0.32\textwidth, height=3.2cm]{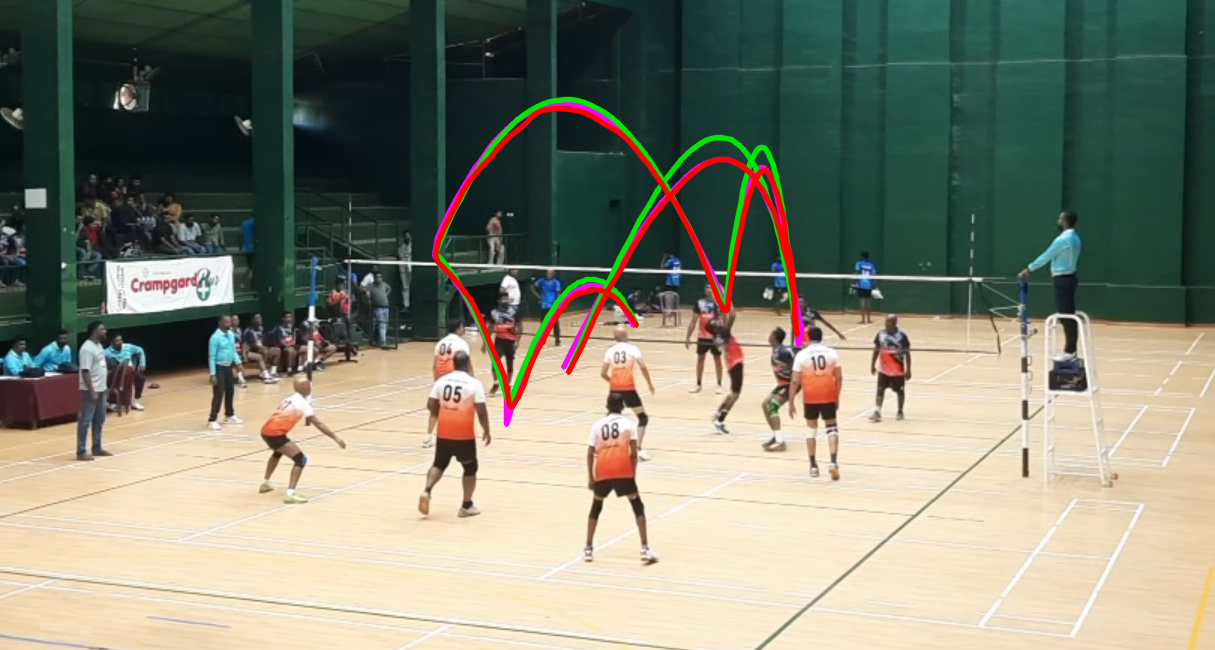} &
    \includegraphics[width=0.32\textwidth, height=3.2cm]{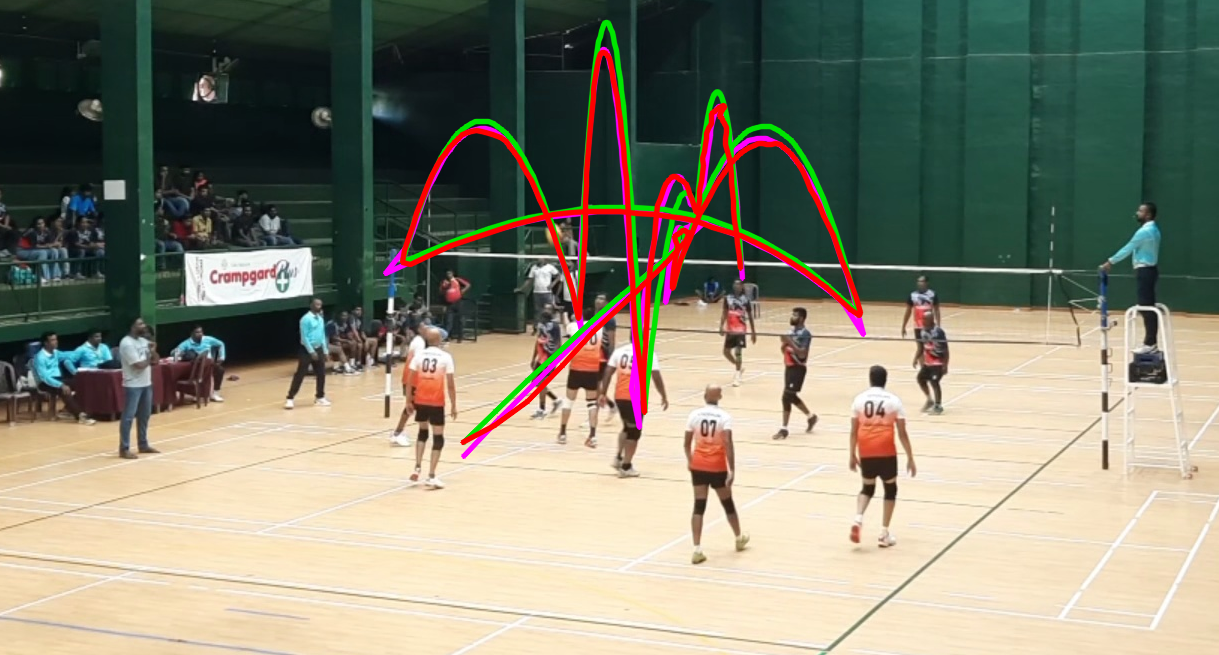} &
    \includegraphics[width=0.32\textwidth, height=3.2cm]{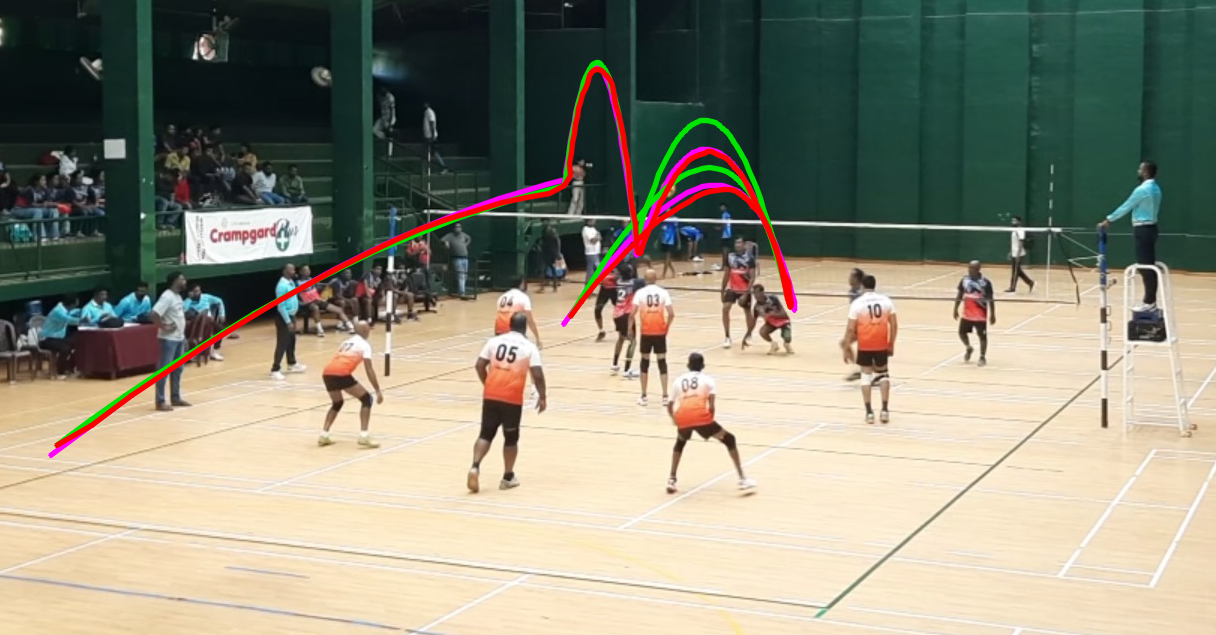} \\
    {\footnotesize (d) 2D Reproj. (rally1)} & {\footnotesize (e) 2D Reproj. (rally2)} & {\footnotesize (f) 2D Reproj. (rally3)} \\[0.8em]
    \midrule
    \vspace{-0.6em} \\
    \includegraphics[width=0.32\textwidth, height=3.4cm]{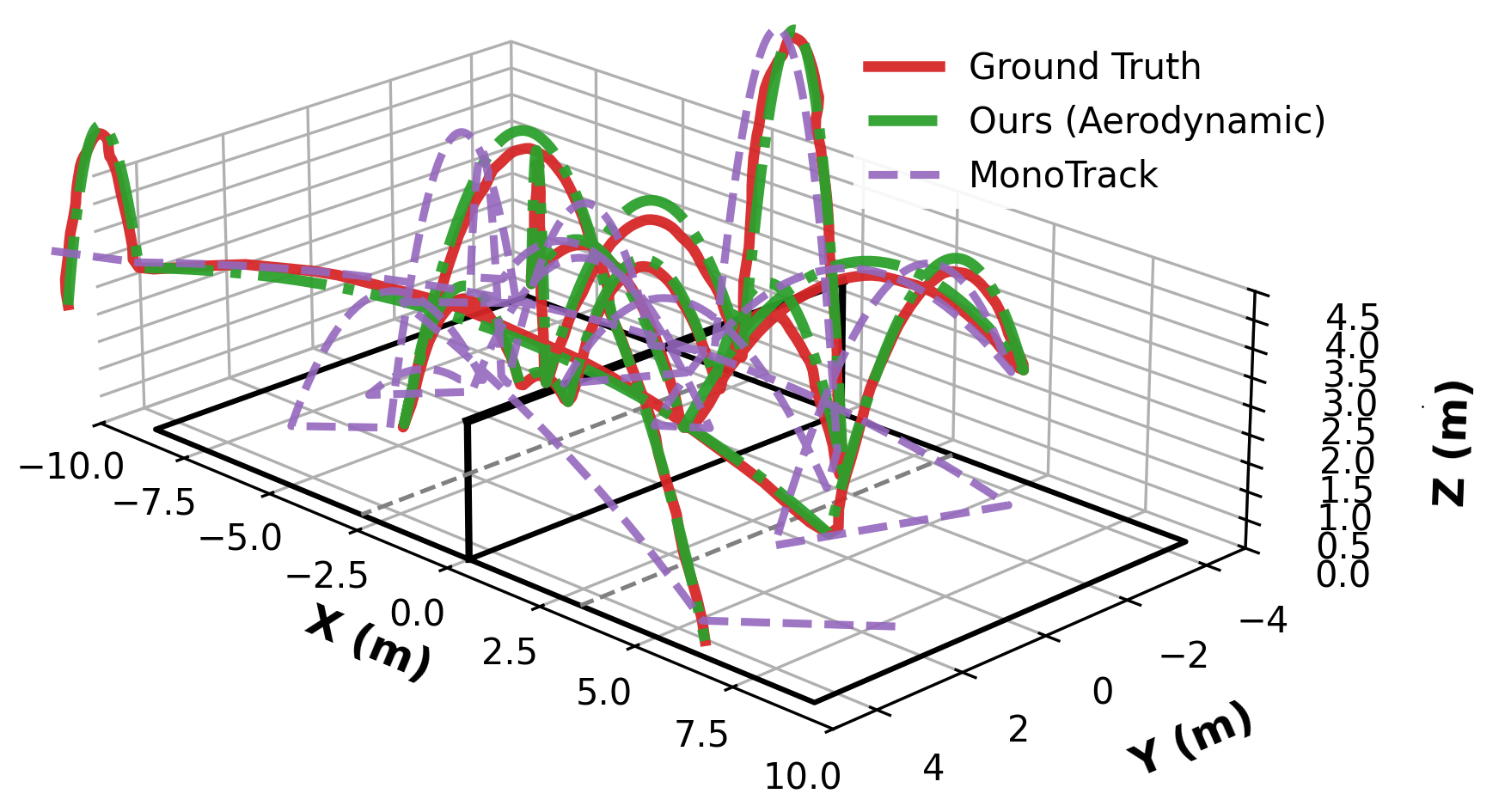} &
    \includegraphics[width=0.32\textwidth, height=3.4cm]{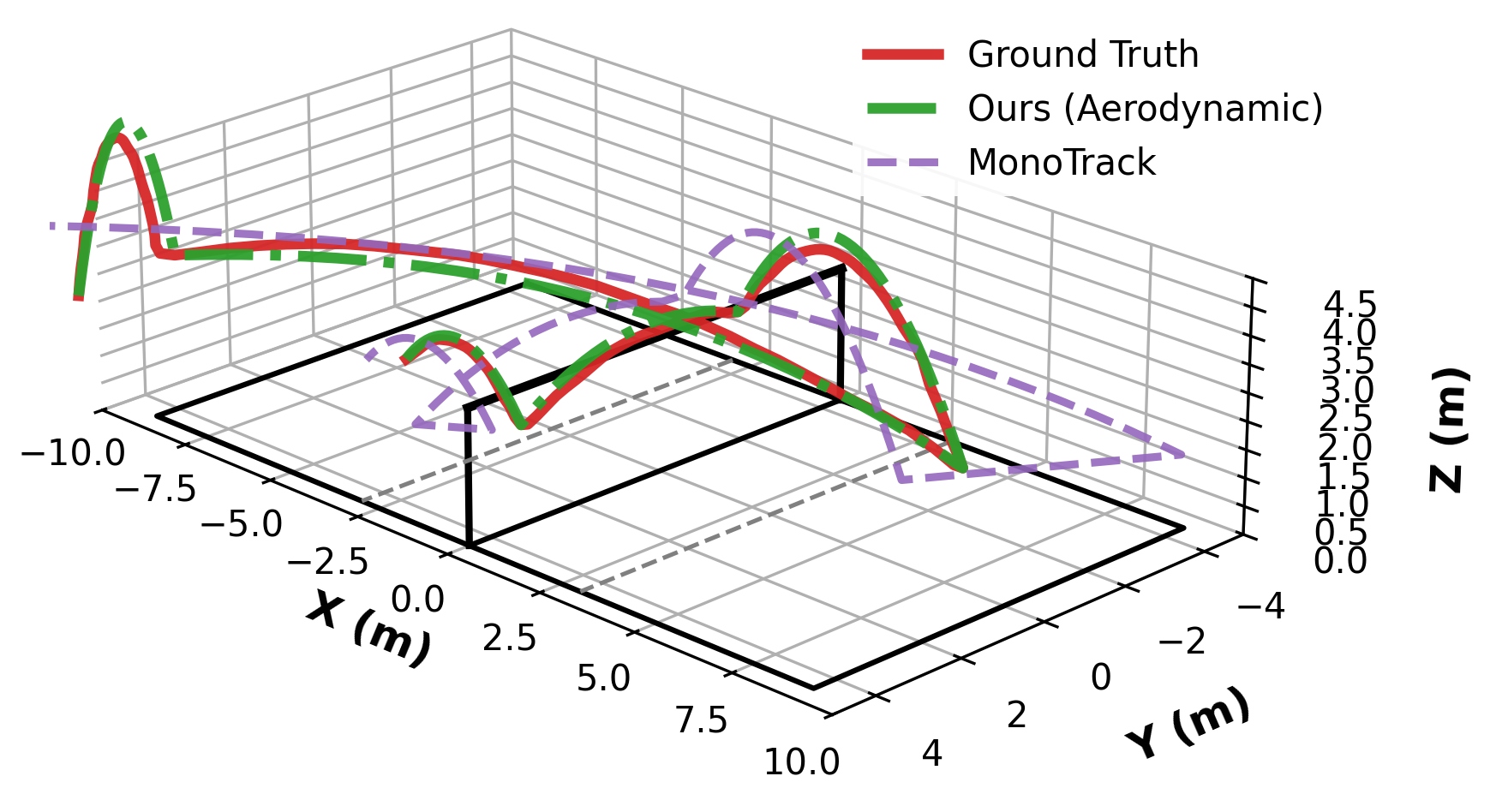} &
    \includegraphics[width=0.32\textwidth, height=3.4cm]{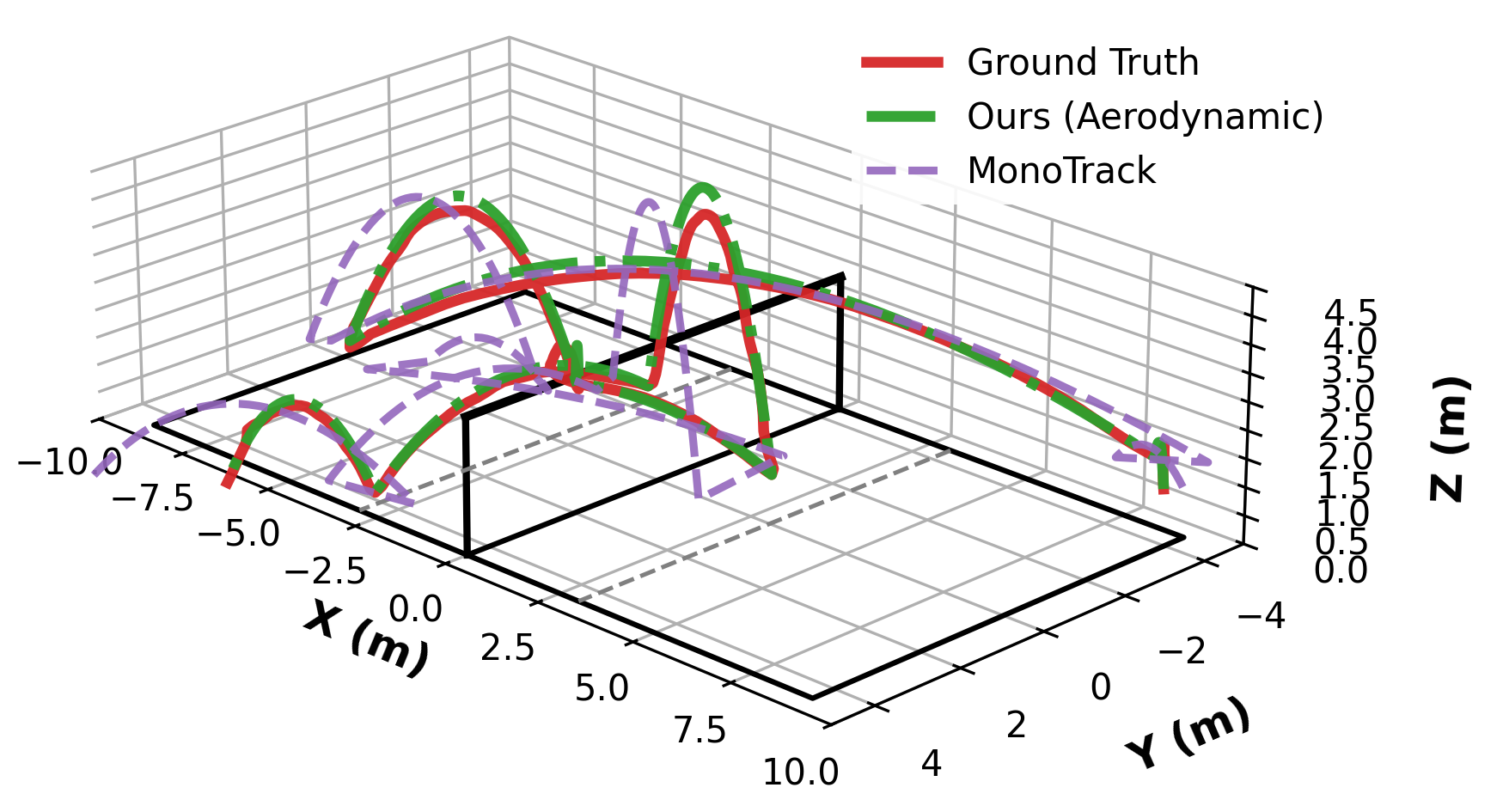} \\
    {\footnotesize (g) 3D Trajectory (rally4)} & {\footnotesize (h) 3D Trajectory (rally5)} & {\footnotesize (i) 3D Trajectory (rally6)} \\[0.4em]

    \includegraphics[width=0.32\textwidth, height=3.2cm]{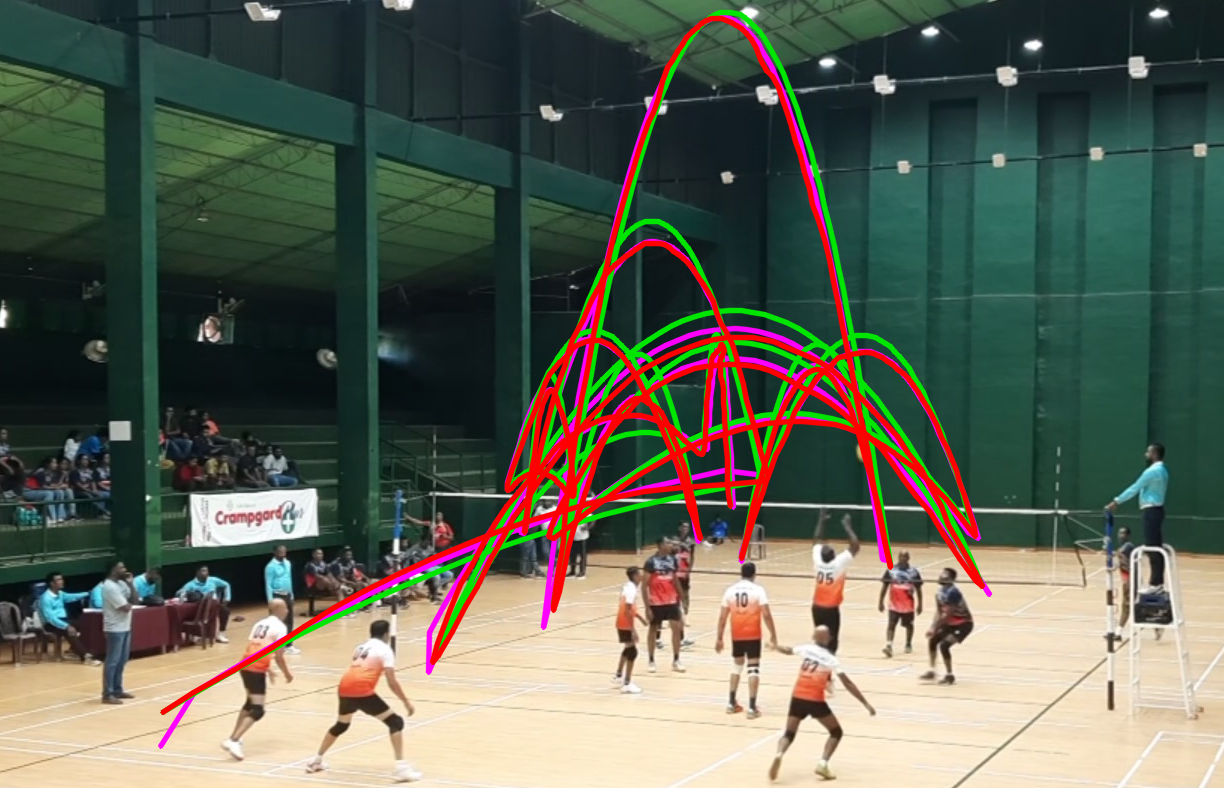} &
    \includegraphics[width=0.32\textwidth, height=3.2cm]{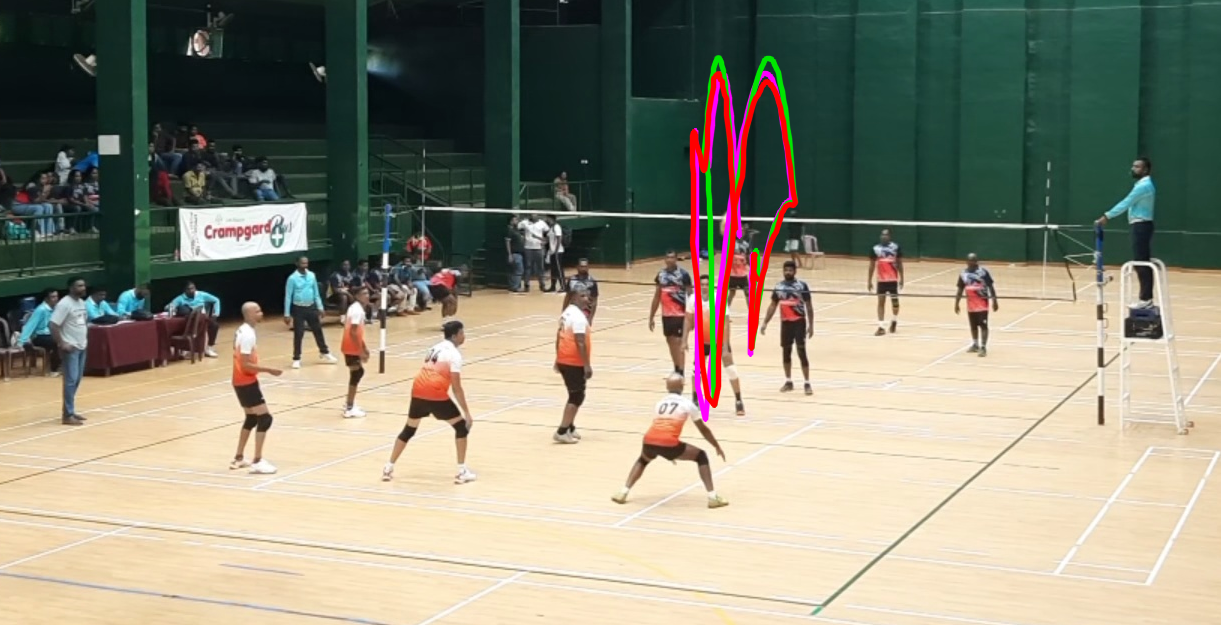} &
    \includegraphics[width=0.32\textwidth, height=3.2cm]{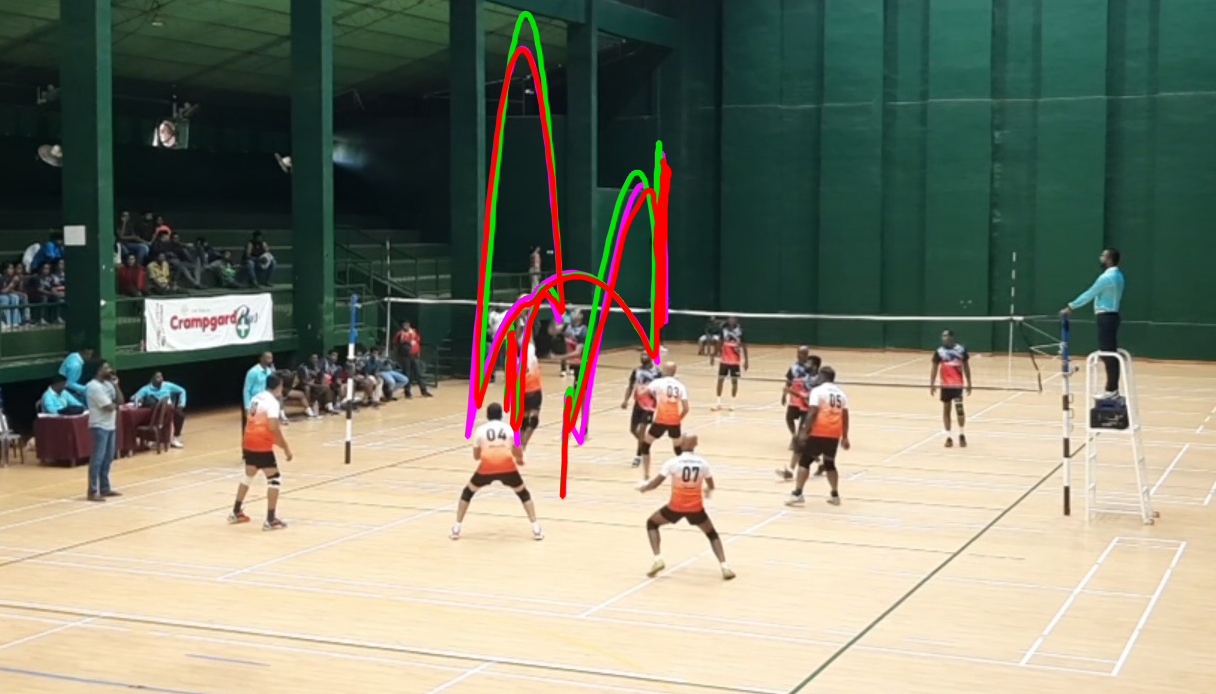} \\
    {\footnotesize (j) 2D Reproj. (rally4)} & {\footnotesize (k) 2D Reproj. (rally5)} & {\footnotesize (l) 2D Reproj. (rally6)} \\
  \end{tabular}
  \vspace{-0.5em}
  \caption{\textbf{Qualitative 2D Reprojection and 3D Trajectory Comparisons on VolliQ.} Visual comparison between MonoTrack~\cite{liu2022monotrack} (purple), our physics-informed lifting formulation with aerodynamic drag (green), and Ground Truth (red) across two distinct rally sequences.}
  \label{fig:supp_qualitative_grid_3cols}
\end{figure*}

\paragraph{Analysis of Cross-View Reprojection and 3D Overfitting.}
Because VolliQ features static multi-view cameras rather than moving broadcast footage, we do not deploy our 2D broadcast vision modules directly on these feeds. Instead, to perform a fair evaluation, all methods share ground-truth trajectory deflection frames. MonoTrack~\cite{liu2022monotrack} is provided with ground-truth camera calibration and triangulated 3D foot contacts, optimizing its trajectory via 2D reprojection on a single camera view. We benchmarked MonoTrack across all four viewpoints, finding that camera \texttt{AN02} yielded its best overall 3D RMSE. In contrast, our physics-informed approach operates directly in metric 3D space using triangulated 3D hand contact boundaries without any 2D reprojection optimization.

As reported in \cref{tab:supp_per_camera_ablation}, MonoTrack minimizes 2D reprojection error exceptionally well on its reference camera (\texttt{AN02}: $3.89\text{ px}$). However, this represents a degenerate projective solution: when reprojection is evaluated on the remaining views, the error explodes to $90.6\text{ px}$ on \texttt{AN01}, $131.0\text{ px}$ on \texttt{AN03}, and $53.7\text{ px}$ on \texttt{AN04}, resulting in severe 3D drift ($1.55\text{ m}$ 3D RMSE). Conversely, our direct 3D physical estimation produces balanced cross-view reprojections ($\approx 17\text{--}23\text{ px}$ across all views) and achieves a precise metric fit ($0.272\text{ m}$ 3D RMSE), demonstrating that while single-view 2D reprojection loss can succeed in confined sports such as table tennis or badminton, it is flawed to resolve depth ambiguity in sports containing rapid aerial ball trajectories such as volleyball. 

This phenomenon is further visualized in~\cref{fig:supp_qualitative_grid_3cols}.

\section{Step-by-Step Pipeline Modules Visualization}
\label{sec:stepbystep}

\begin{figure*}[t]
  \centering
  \small
  \setlength{\tabcolsep}{11pt}
  \begin{tabular}{cc}
    \includegraphics[width=0.41\textwidth, height=3.6cm]{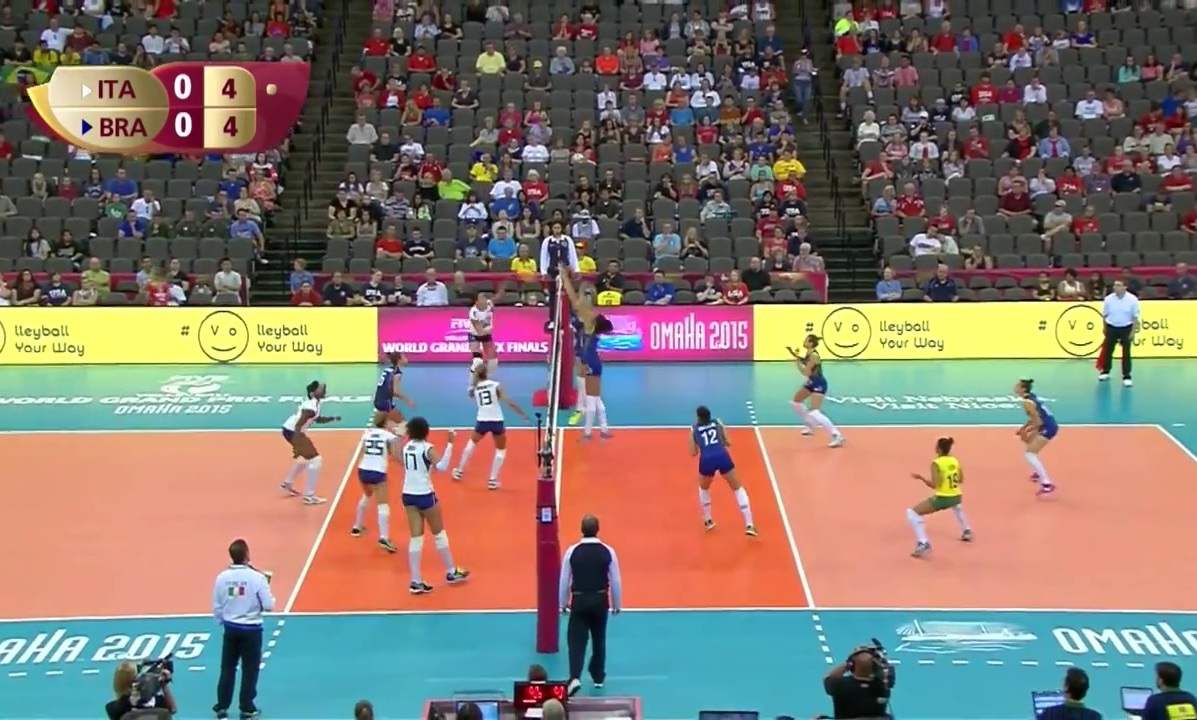} &
    \includegraphics[width=0.41\textwidth, height=3.6cm]{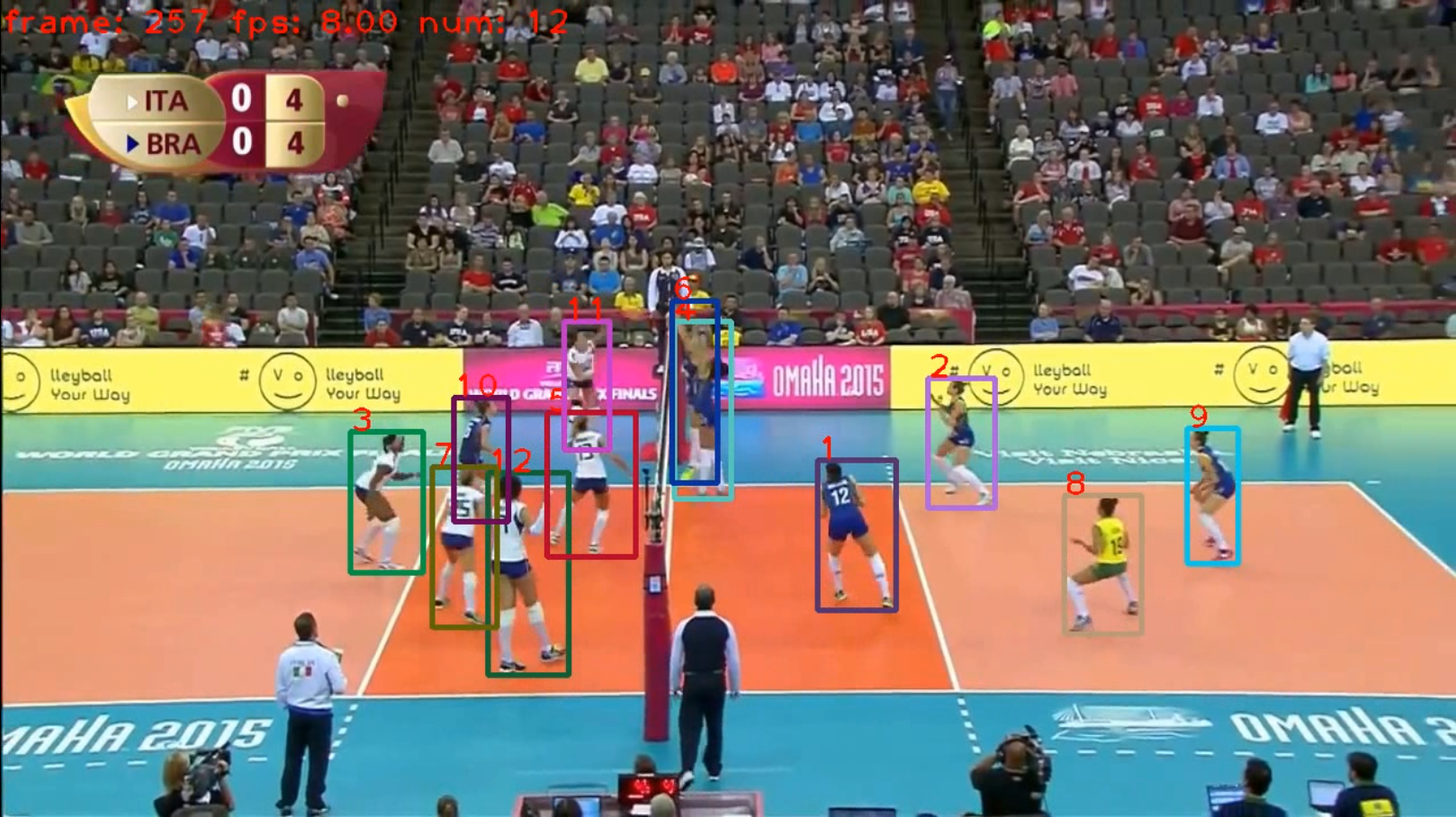} \\
    {\footnotesize (a) Raw Broadcast Frame} & {\footnotesize (b) Player Detection \& Tracking} \\[0.6em]

    \includegraphics[width=0.41\textwidth, height=3.6cm]{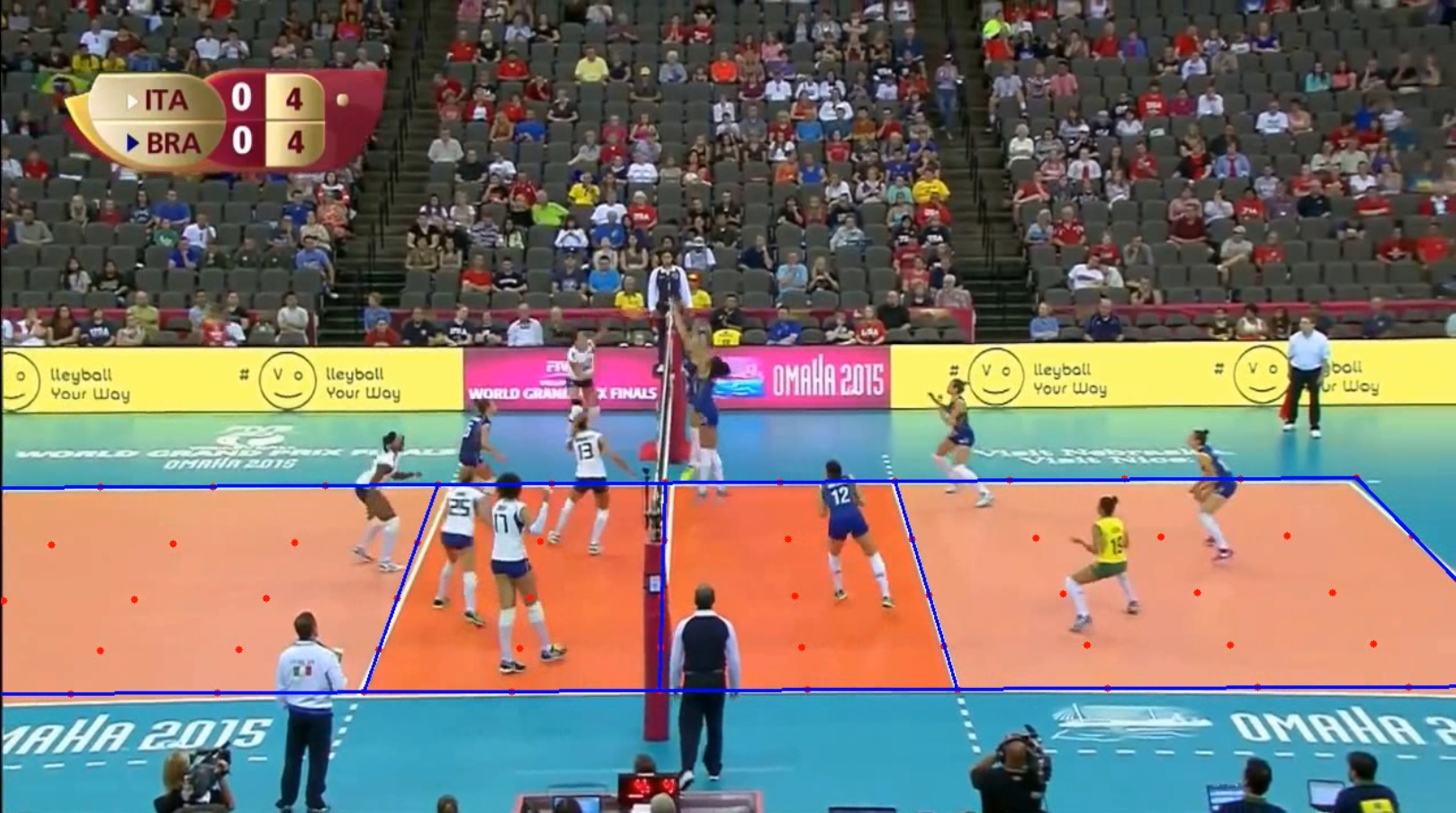} &
    \includegraphics[width=0.41\textwidth, height=3.6cm]{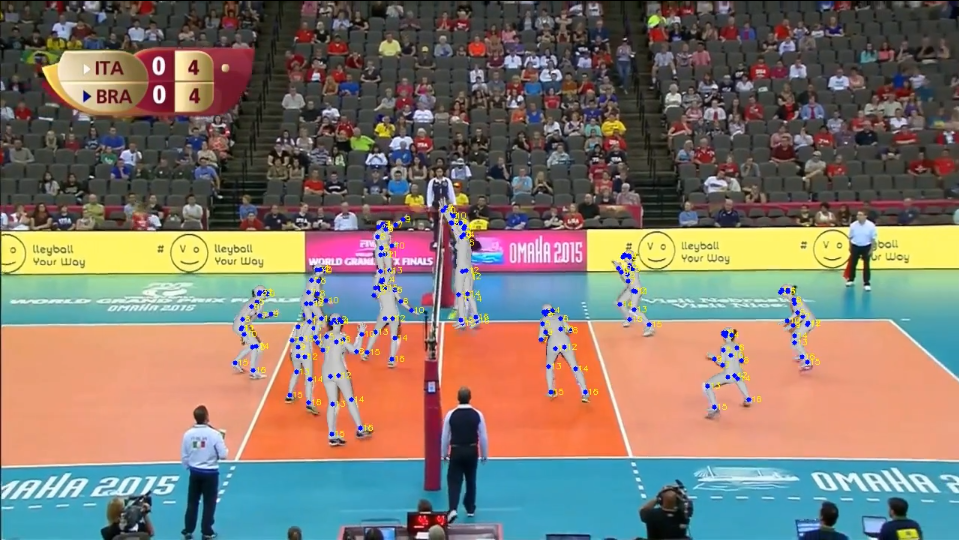} \\
    {\footnotesize (c) Court Calibration Overlay} & {\footnotesize (d) WHAM Human Mesh Recovery} \\[0.6em]

    \includegraphics[width=0.41\textwidth, height=3.6cm]{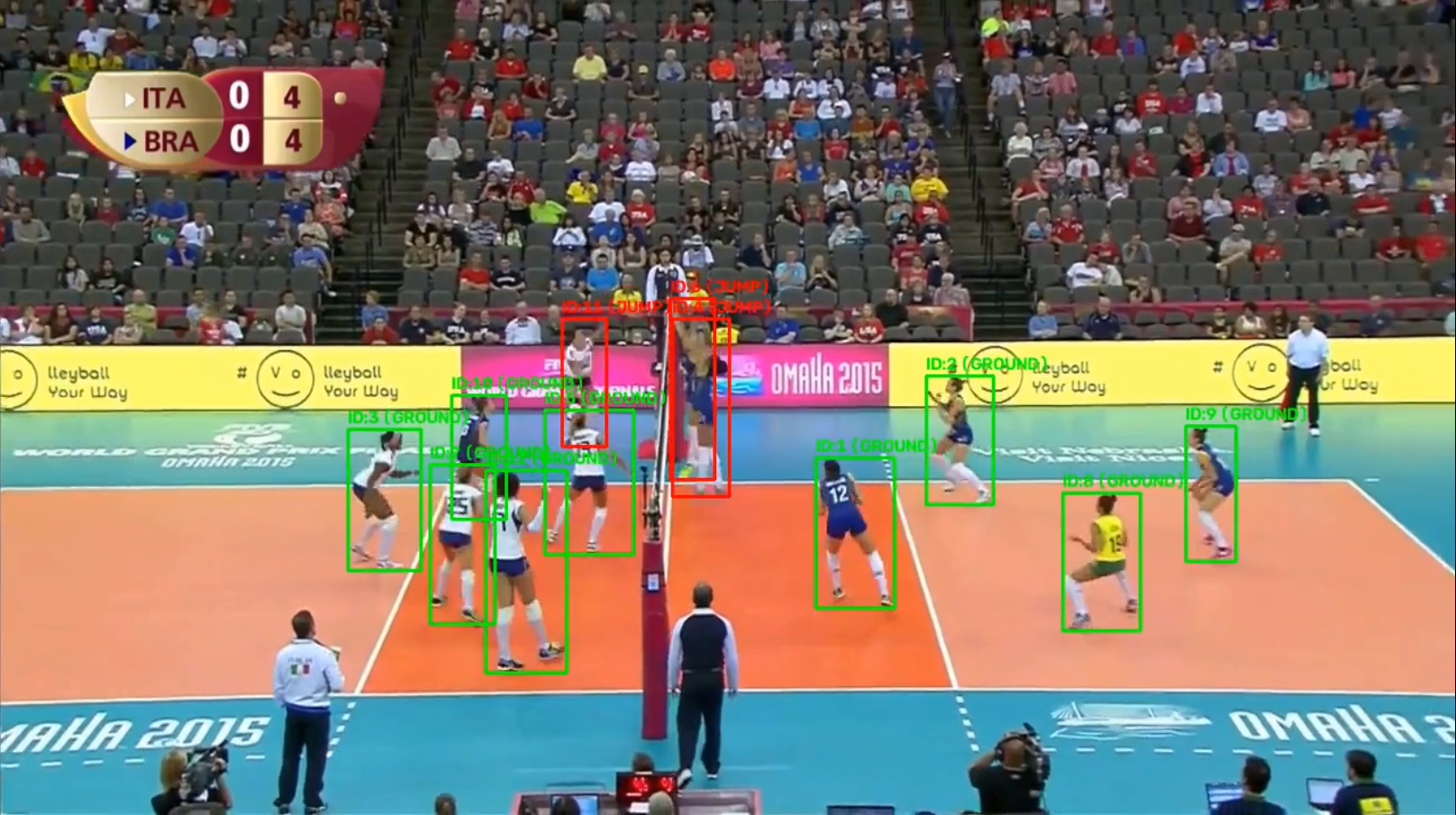} &
    \includegraphics[width=0.41\textwidth, height=3.6cm]{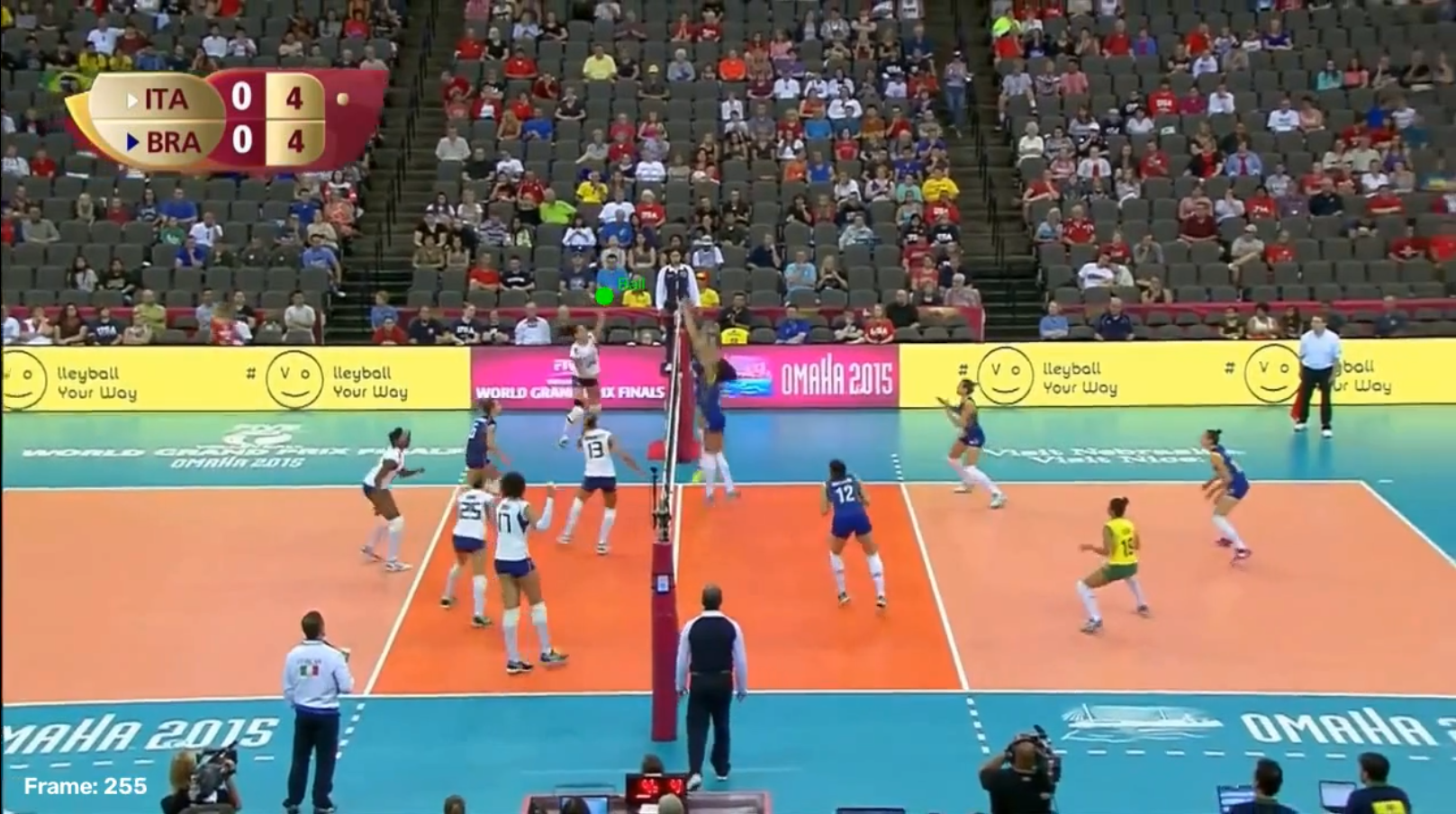} \\
    {\footnotesize (e) Jump Spotting (\textcolor{green}{Grounded}/\textcolor{red}{Aerial})} & {\footnotesize (f) 2D Ball Detection} \\[0.6em]

    \includegraphics[width=0.41\textwidth, height=3.6cm]{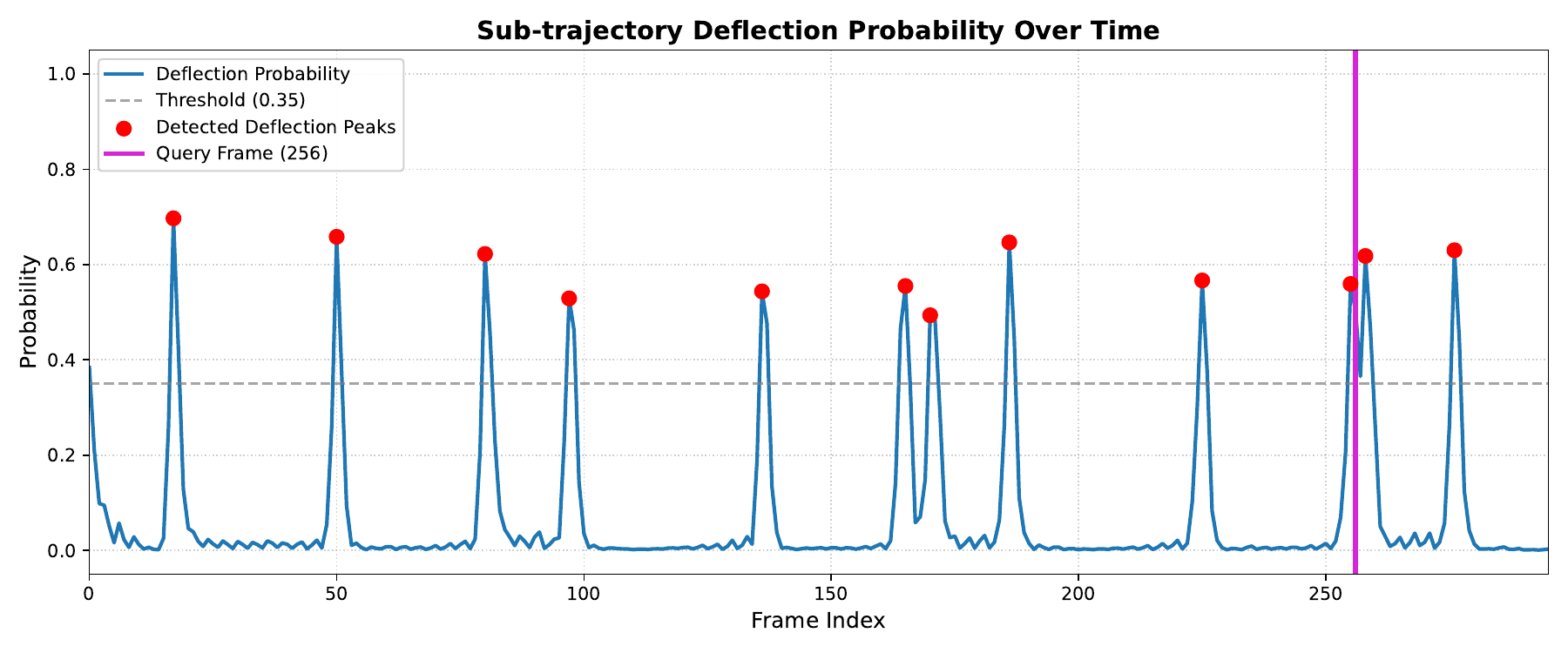} &
    \includegraphics[width=0.41\textwidth, height=3.6cm]{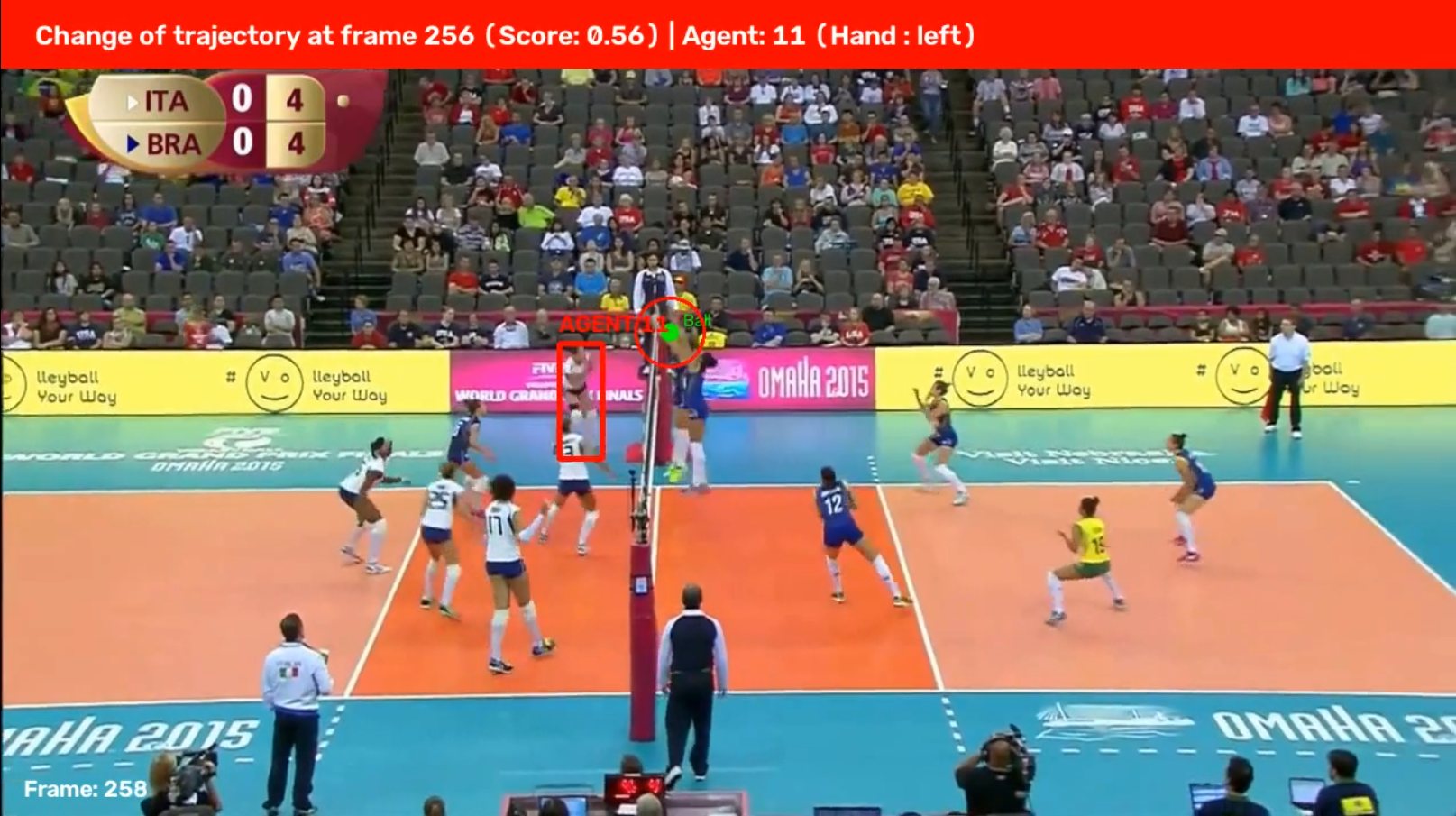} \\
    {\footnotesize (g) Sub-trajectory TCN Deflection Score} & {\footnotesize (h) Collision Agent Matching} \\[0.6em]

    \includegraphics[width=0.41\textwidth, height=3.6cm]{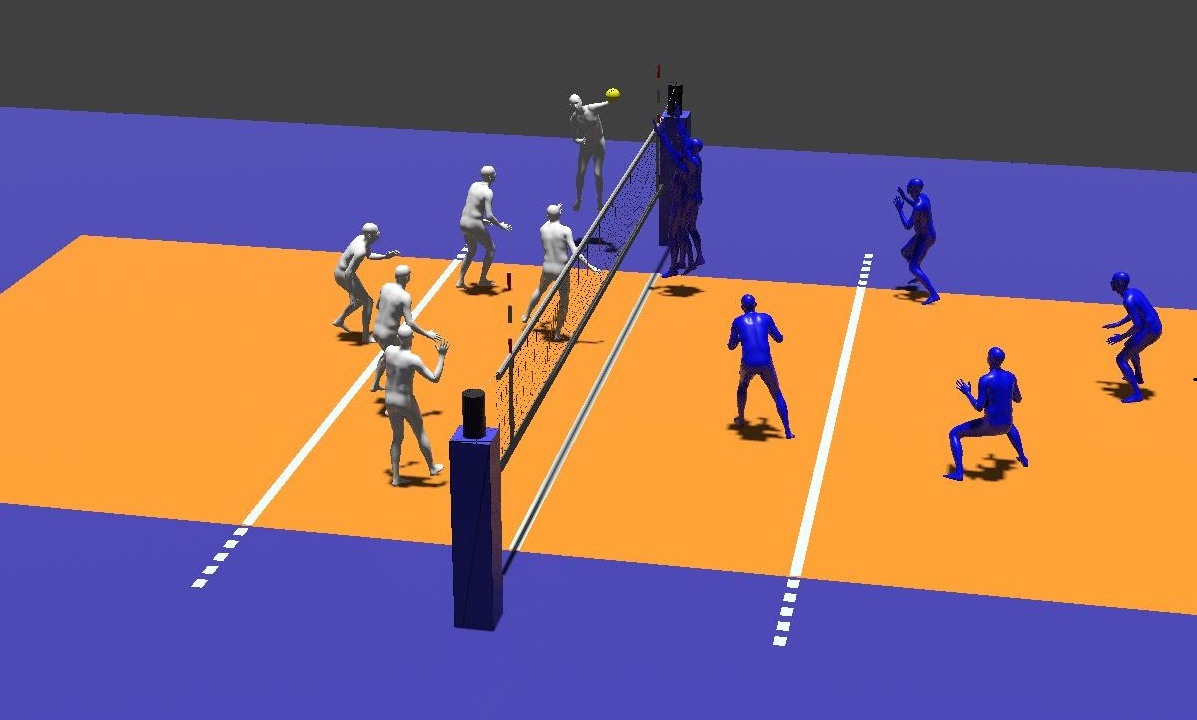} &
    \includegraphics[width=0.41\textwidth, height=3.6cm]{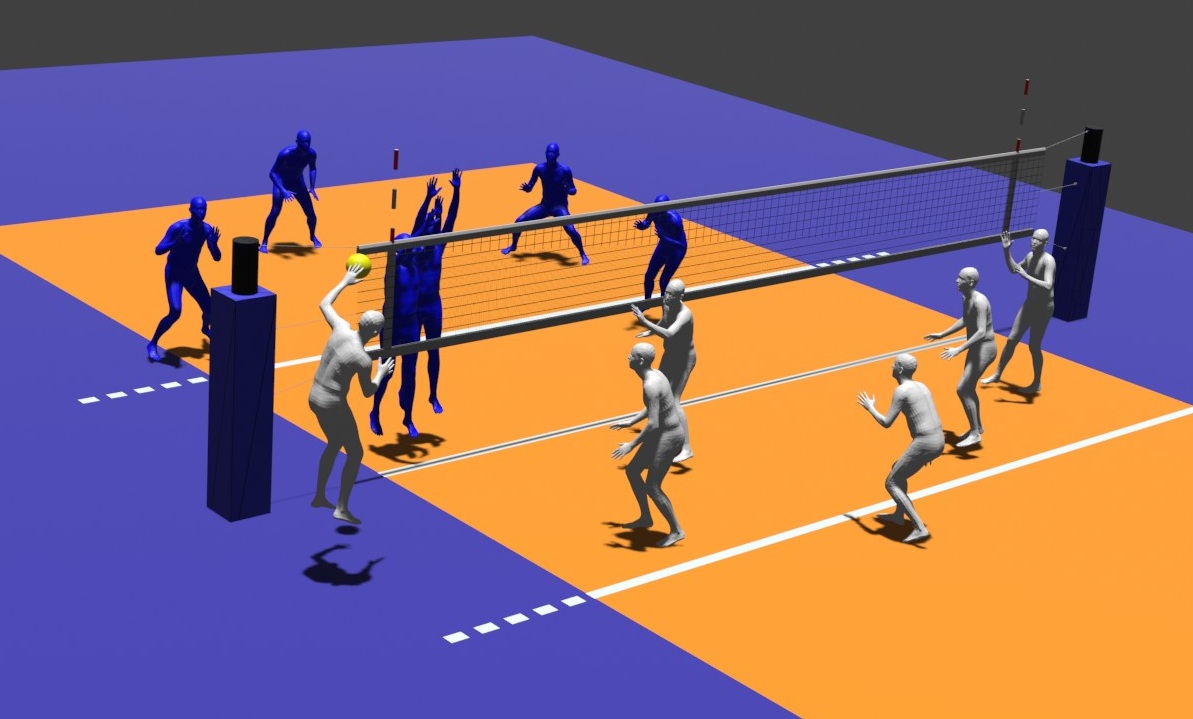} \\
    {\footnotesize (i) Final 3D Scene Reconstruction} & {\footnotesize (j) Final 3D Scene Reconstruction (Other POV)} \\
  \end{tabular}
  \vspace{-0.5em}
  \caption{\textbf{Step-by-Step Visualization of Intermediate Pipeline Stages.} Visual walkthrough on the teaser rally frame: (a) raw input feed; (b) 2D bounding boxes and tracklet IDs; (c) homography estimation and projected court grid; (d) SMPL body mesh recovery; (e) jump event spotting; (f) localized 2D ball position; (g) temporal TCN deflection probabilities across the rally (pink vertical bar indicates the query frame); (h) heuristic wrist proximity and velocity scoring attributing the collision; (i, j) final 3D scene reconstructed in Blender.}
  \label{fig:supp_step_by_step_pipeline}
\end{figure*}

To illustrate the monocular lifting workflow, \cref{fig:supp_step_by_step_pipeline} provides a step-by-step visual inspection of intermediate outputs generated by each module on the representative rally frame presented in the teaser figure of the main paper.

\section{Full Reconstruction Visual Examples}
\label{sec:visual_examples}

\begin{figure*}[p]
  \centering
  \small
  \begin{tabular}{cc}
    \includegraphics[width=0.41\textwidth, height=3.9cm]{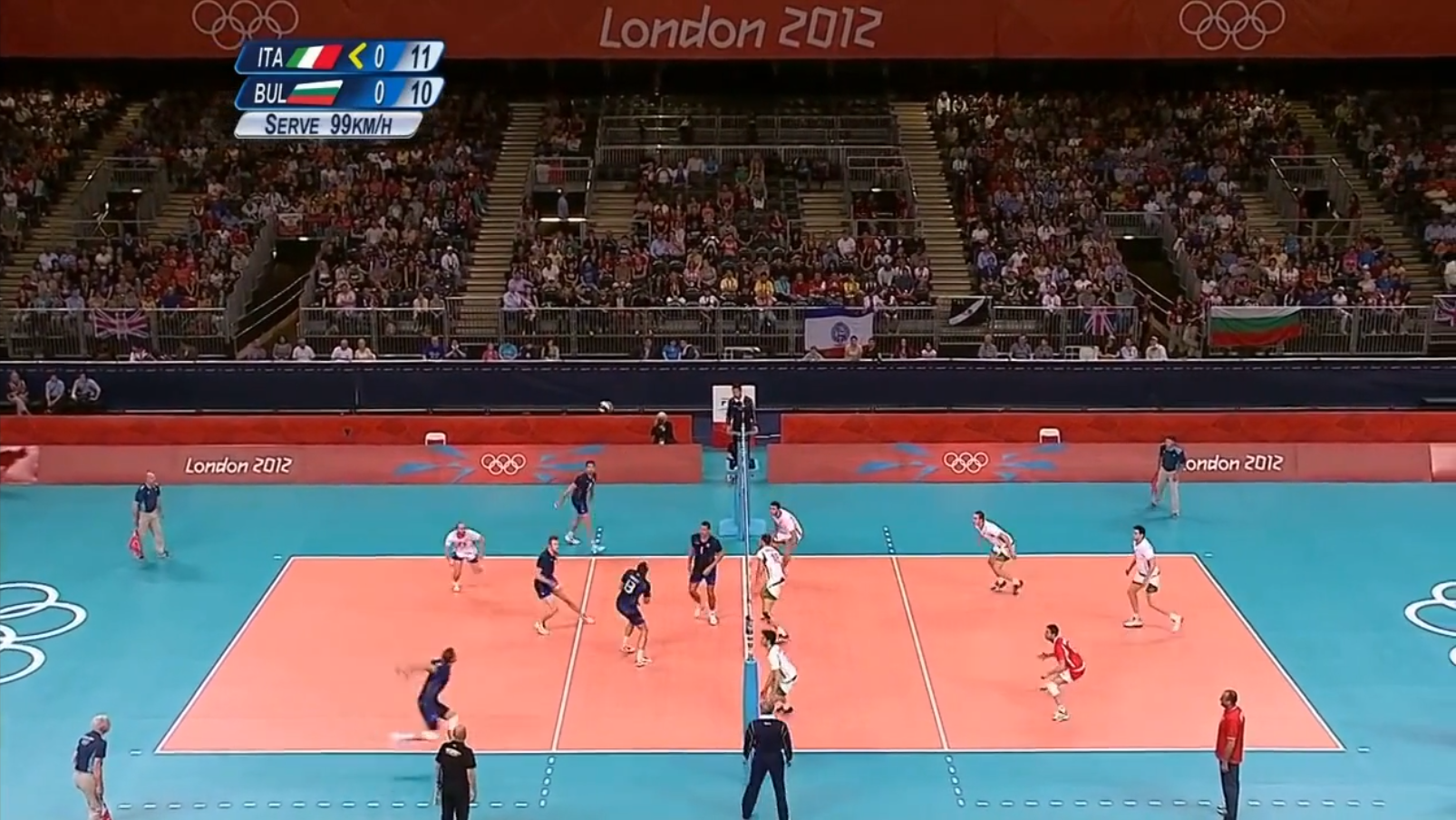} &
    \includegraphics[width=0.41\textwidth, height=3.9cm]{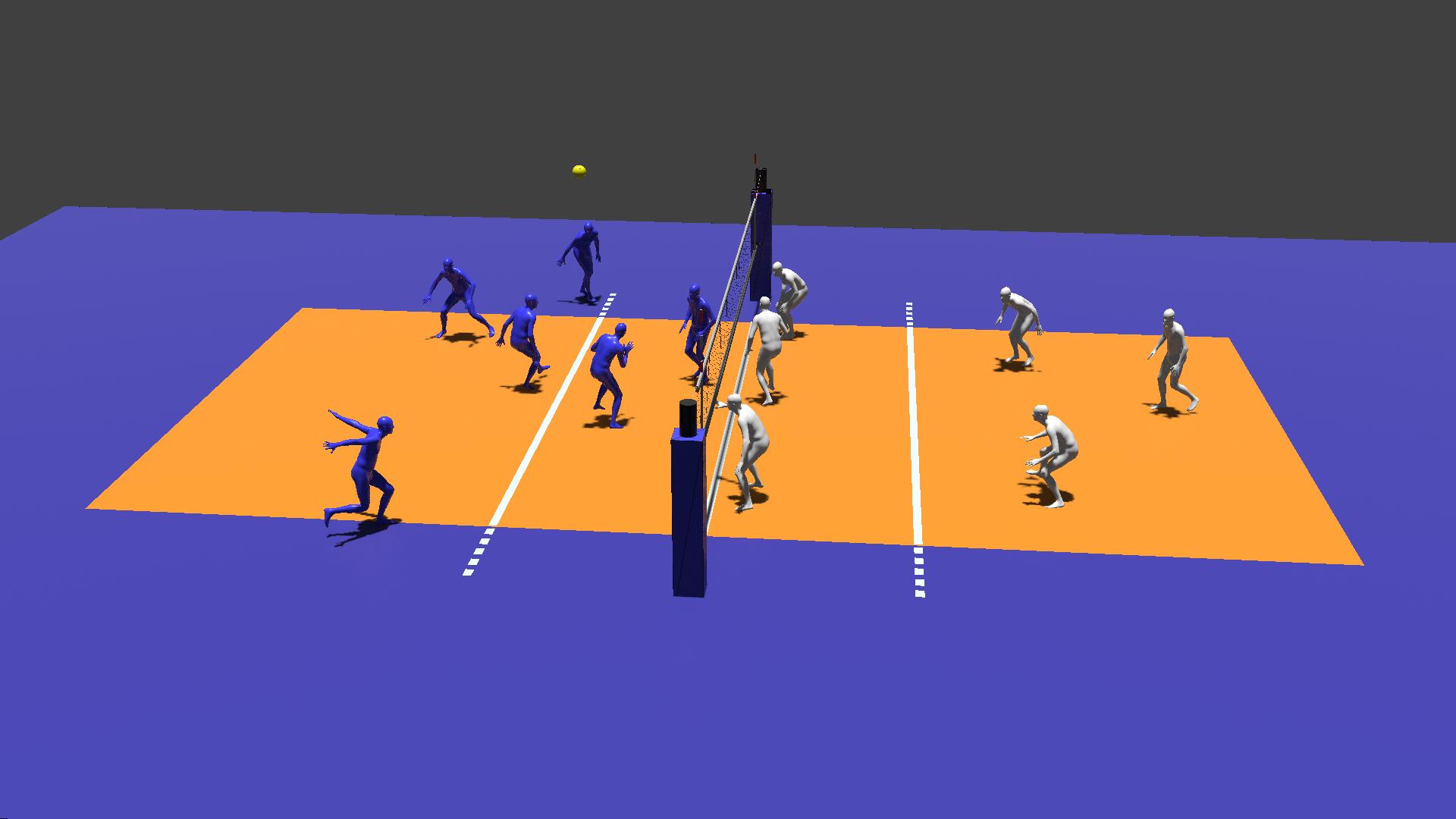} \\
    {\footnotesize (a) Input Broadcast Frame 1} & {\footnotesize (b) Reconstructed 3D Scene 1} \\[0.5em]

    \includegraphics[width=0.41\textwidth, height=3.9cm]{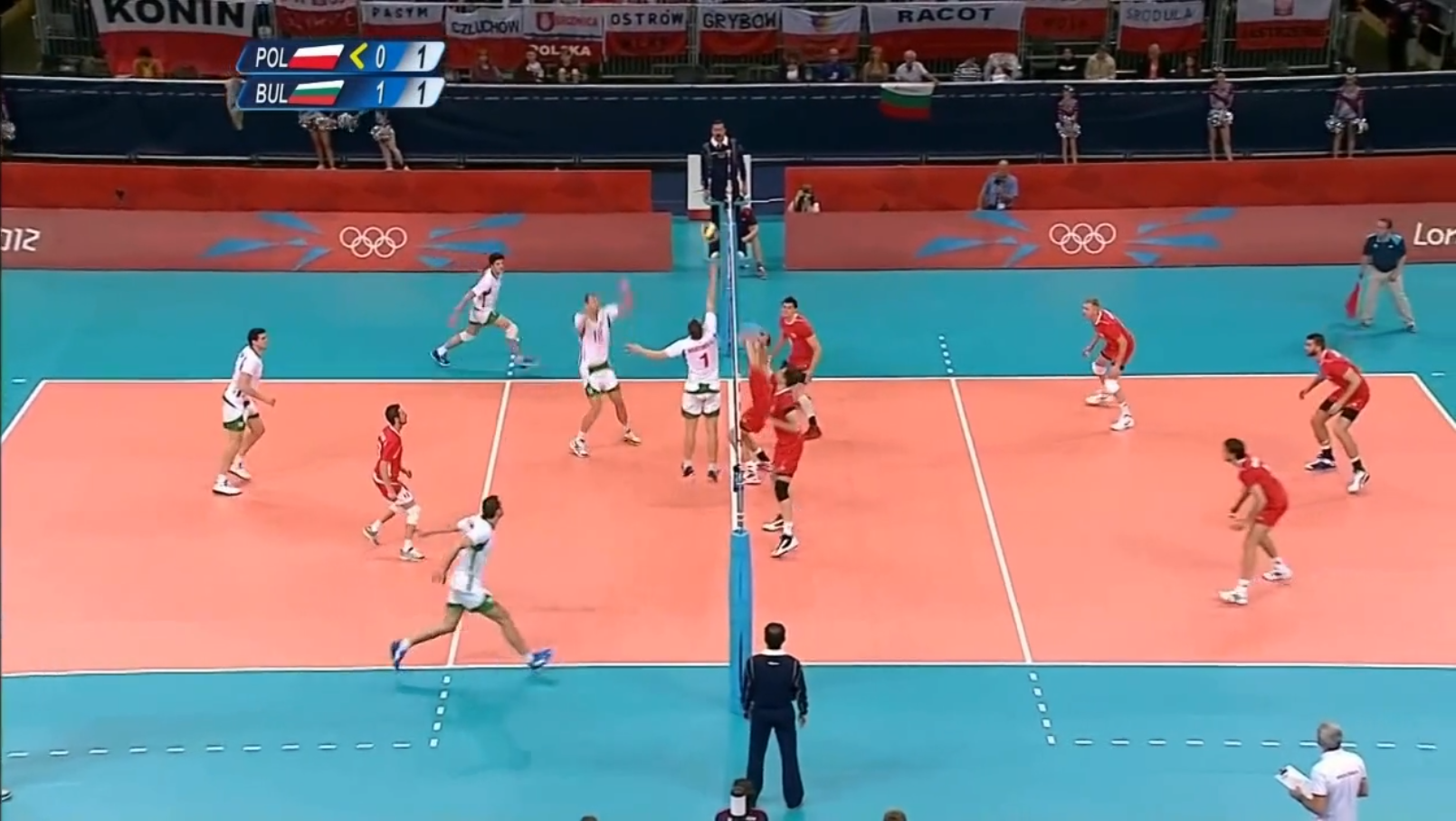} &
    \includegraphics[width=0.41\textwidth, height=3.9cm]{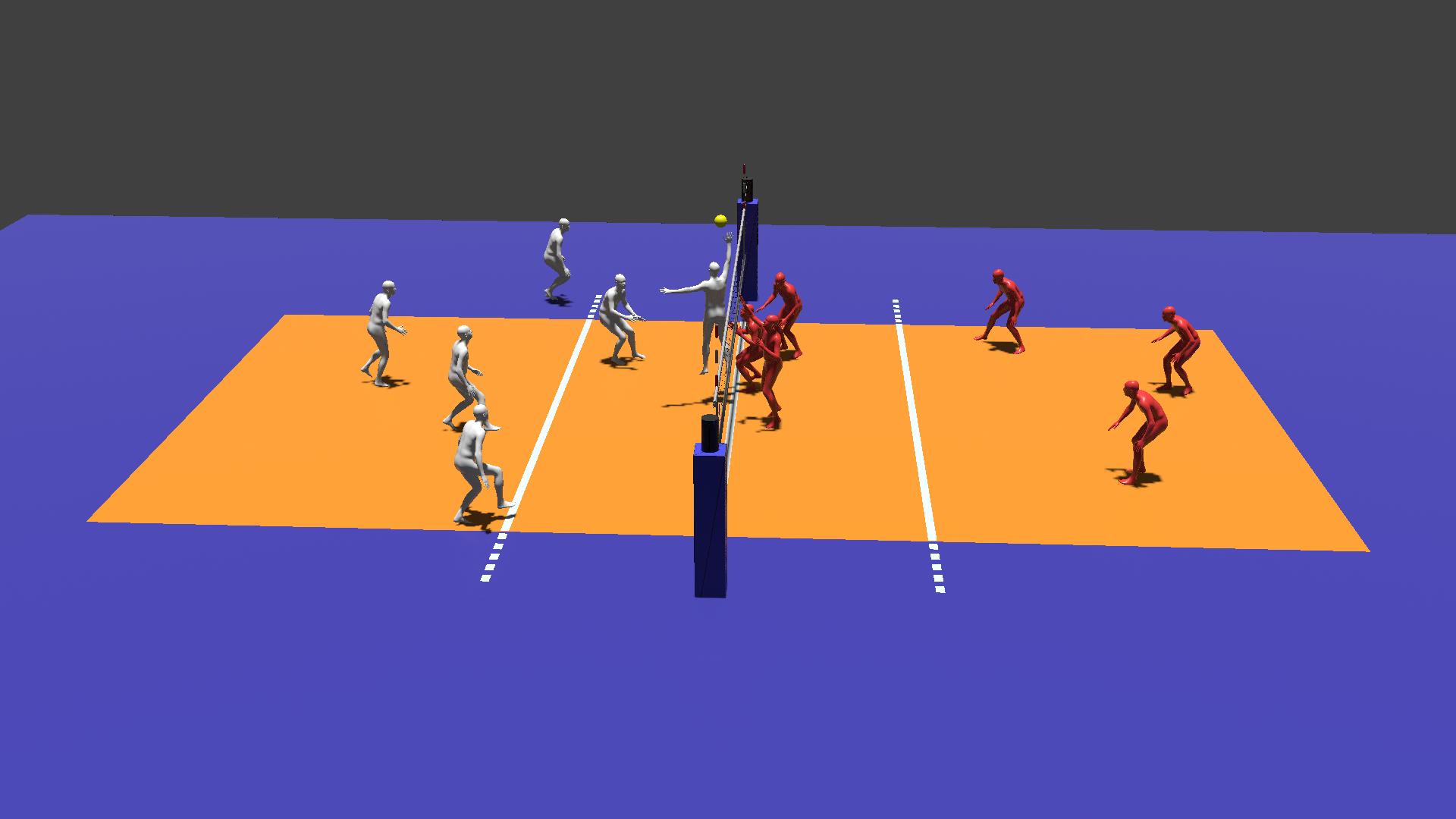} \\
    {\footnotesize (c) Input Broadcast Frame 2} & {\footnotesize (d) Reconstructed 3D Scene 2} \\[0.5em]

    \includegraphics[width=0.41\textwidth, height=3.9cm]{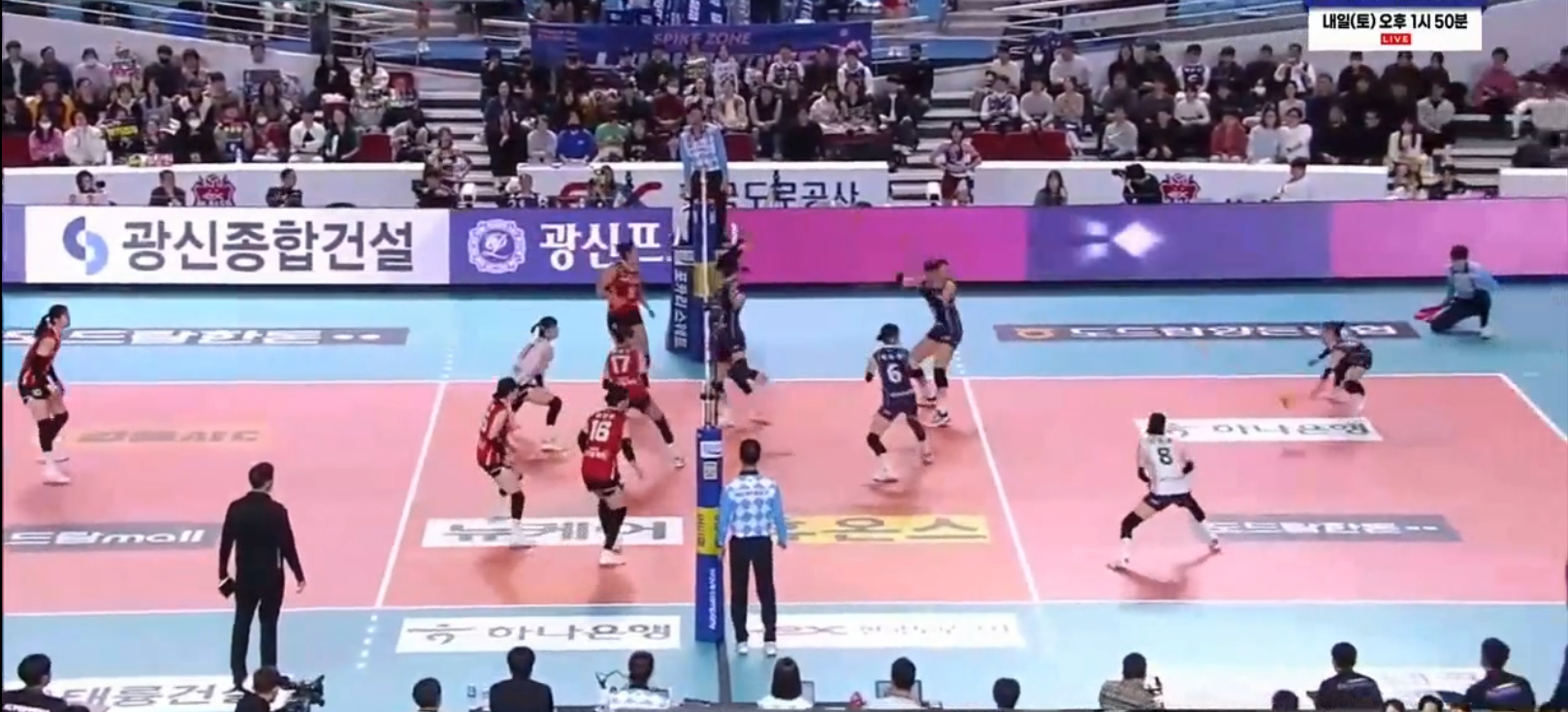} &
    \includegraphics[width=0.41\textwidth, height=3.9cm]{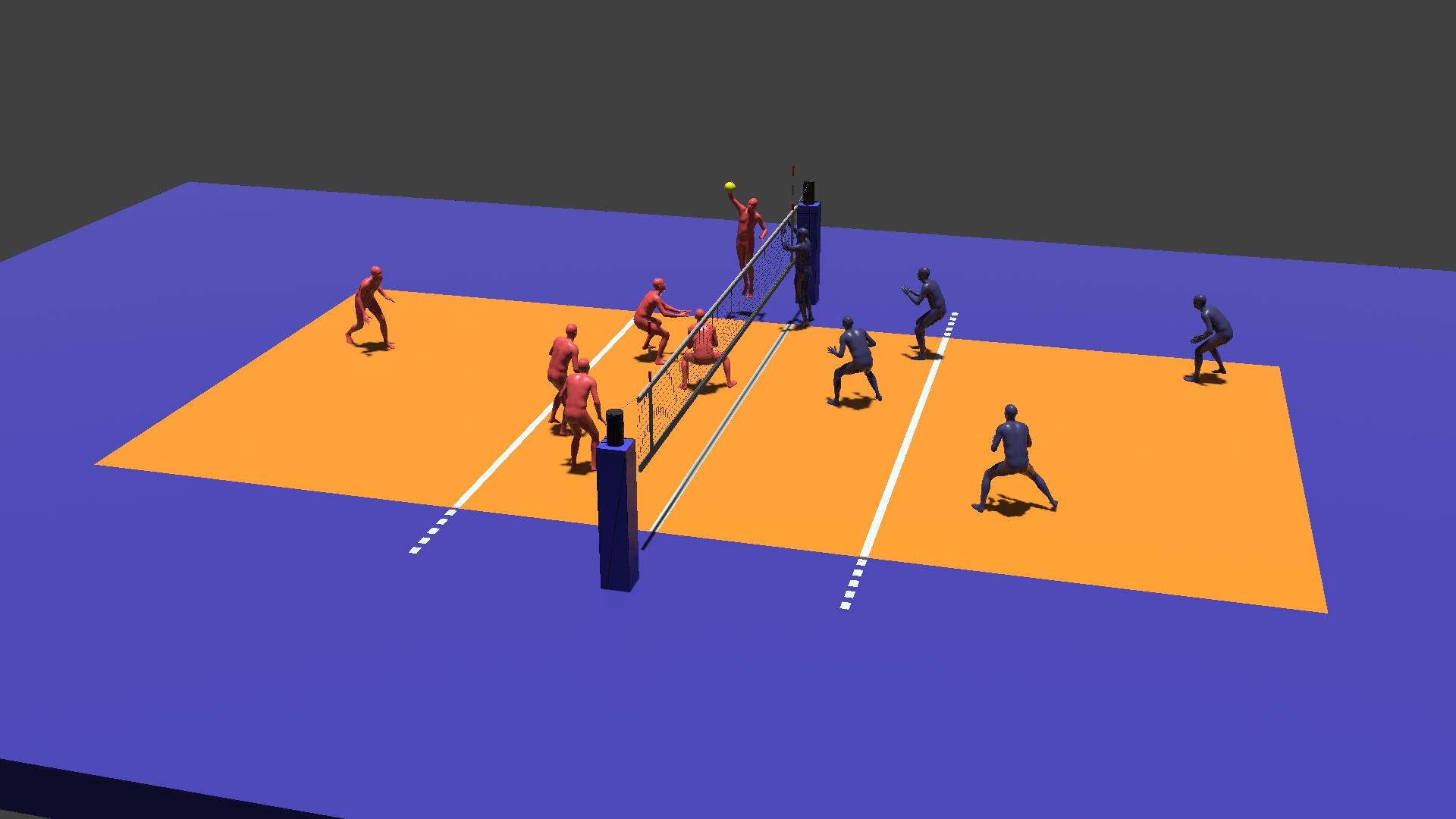} \\
    {\footnotesize (e) Input Broadcast Frame 3} & {\footnotesize (f) Reconstructed 3D Scene 3} \\[0.5em]

    \includegraphics[width=0.41\textwidth, height=3.9cm]{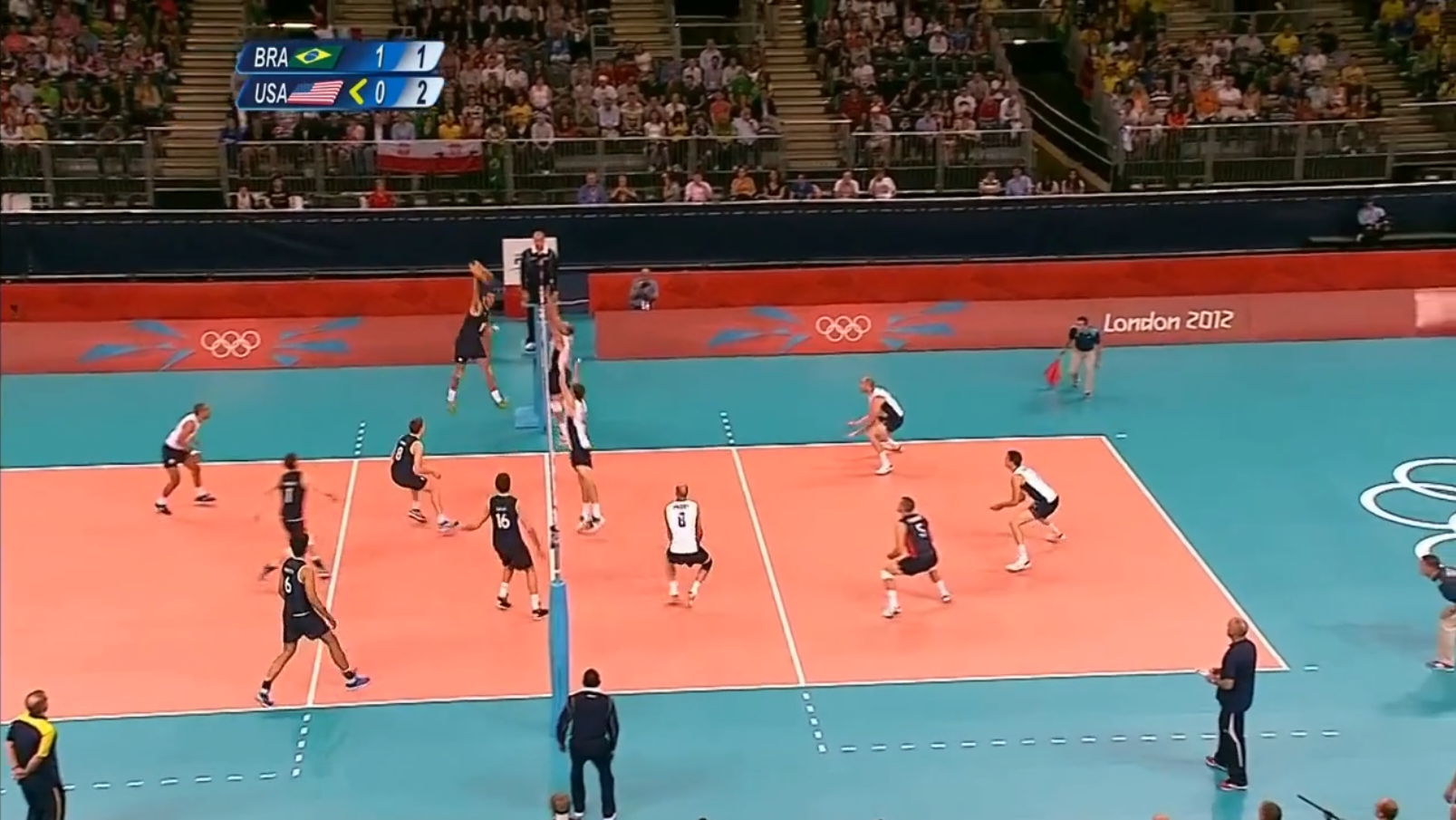} &
    \includegraphics[width=0.41\textwidth, height=3.9cm]{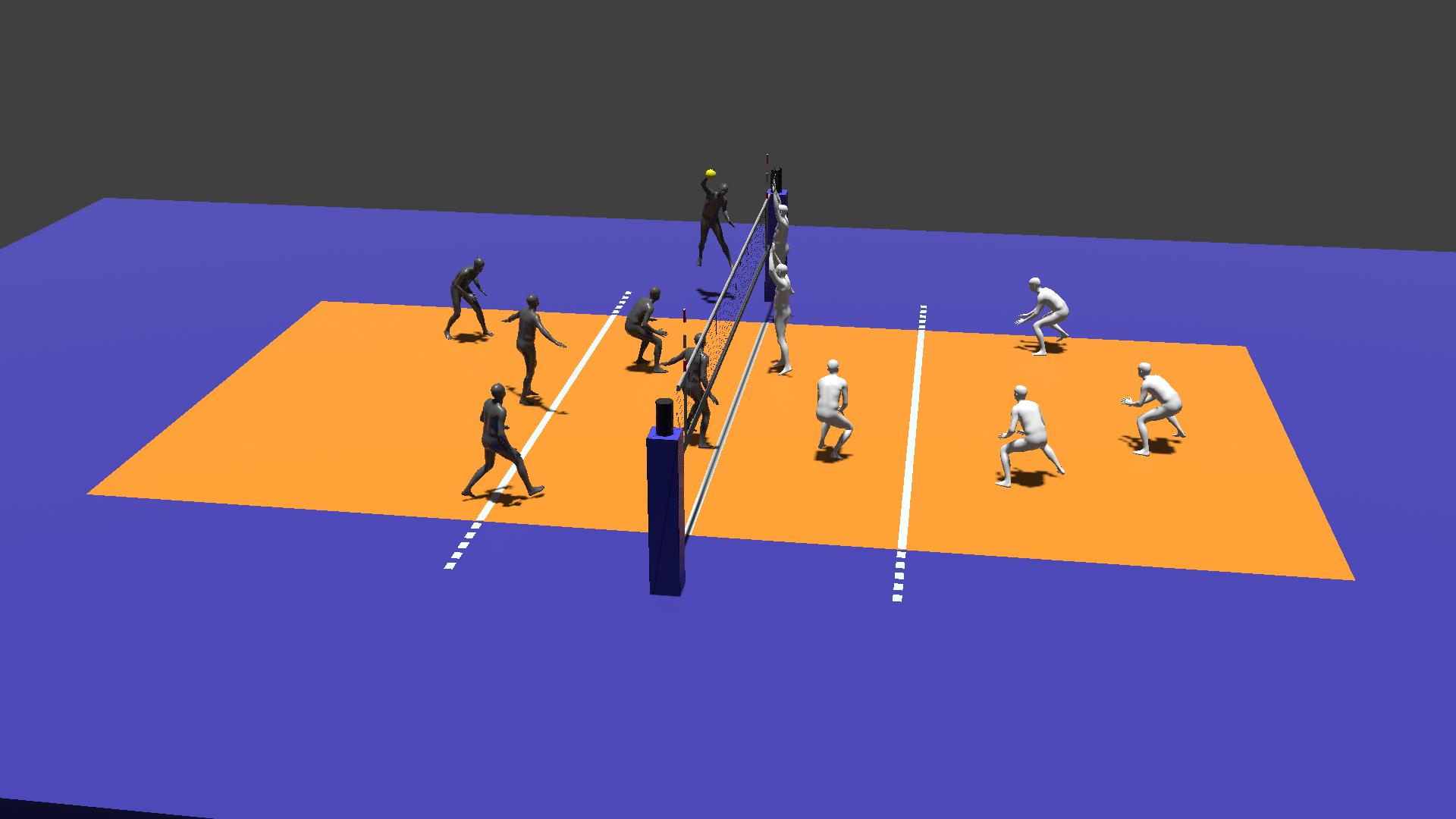} \\
    {\footnotesize (g) Input Broadcast Frame 4} & {\footnotesize (h) Reconstructed 3D Scene 4} \\[0.5em]

    \includegraphics[width=0.41\textwidth, height=3.9cm]{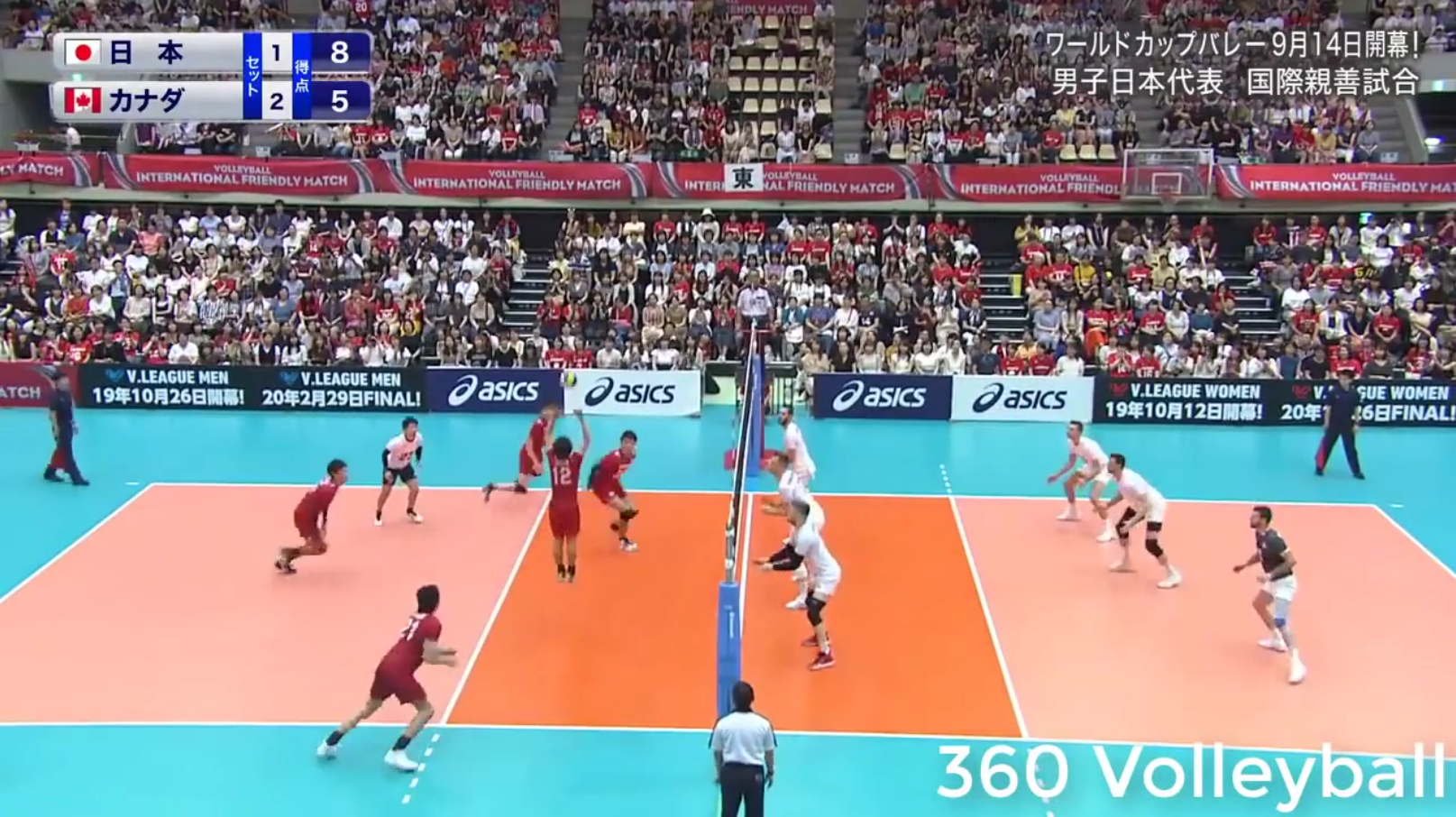} &
    \includegraphics[width=0.41\textwidth, height=3.9cm]{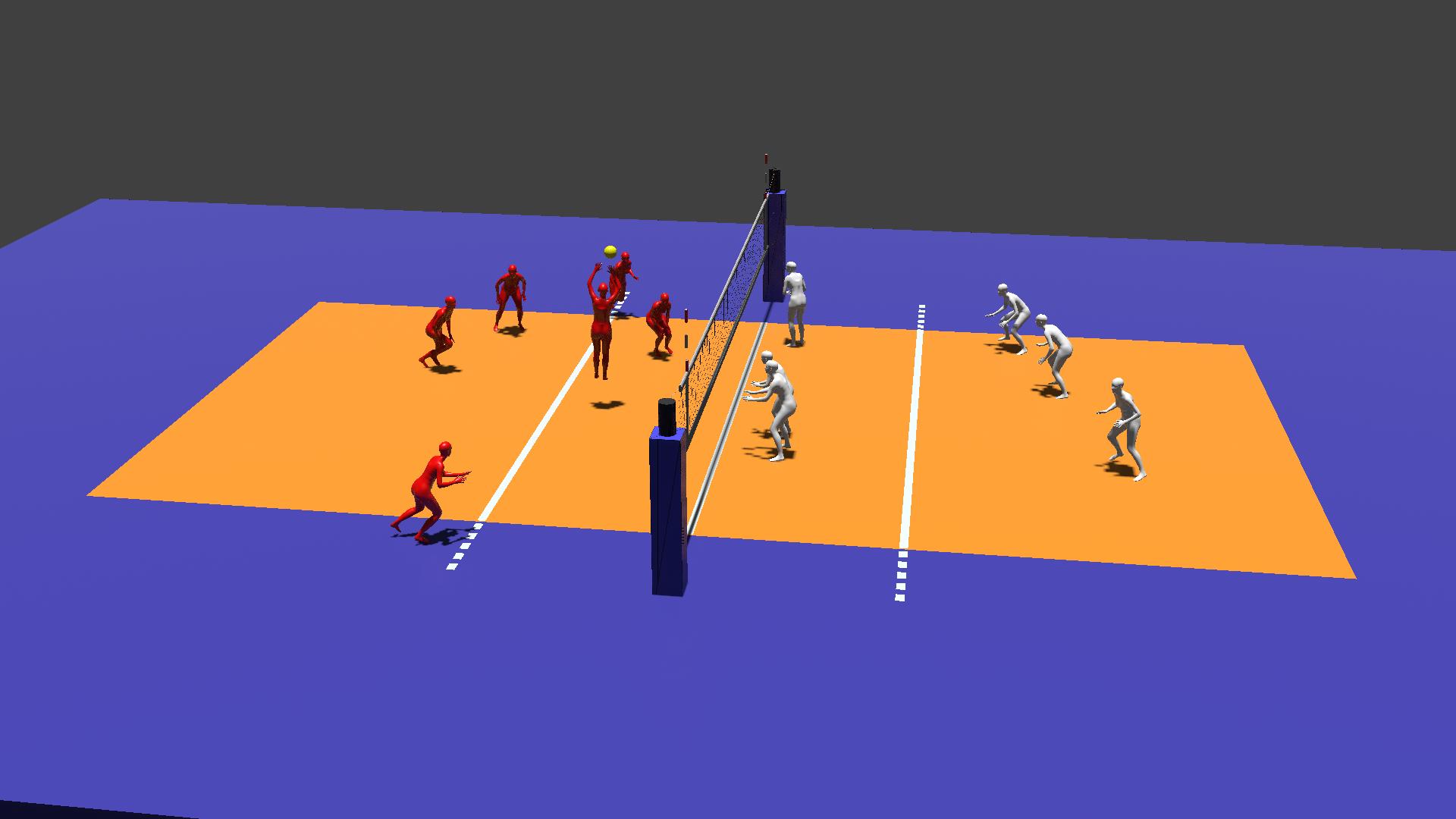} \\
    {\footnotesize (i) Input Broadcast Frame 5} & {\footnotesize (j) Reconstructed 3D Scene 5} \\
  \end{tabular}
  \vspace{-0.5em}
  \caption{\textbf{Qualitative 3D Reconstruction Visualizations (Part 1/2).} Paired comparisons between monocular broadcast input frames (left) and the complete metric 3D scene reconstructed in Blender (right).}
  \label{fig:supp_visual_examples_part1}
\end{figure*}

\begin{figure*}[p]
  \centering
  \small
  \begin{tabular}{cc}
    \includegraphics[width=0.41\textwidth, height=3.9cm]{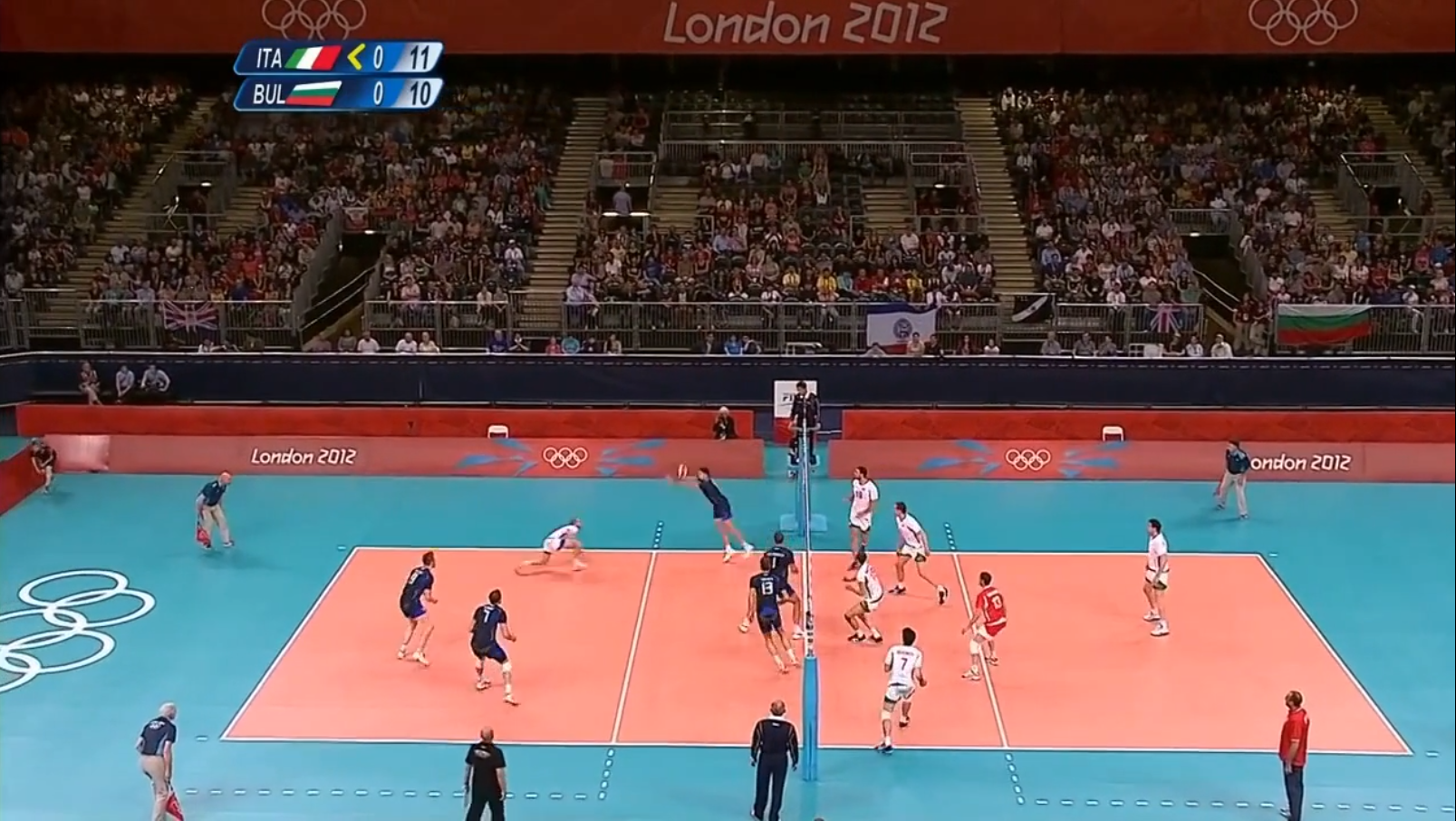} &
    \includegraphics[width=0.41\textwidth, height=3.9cm]{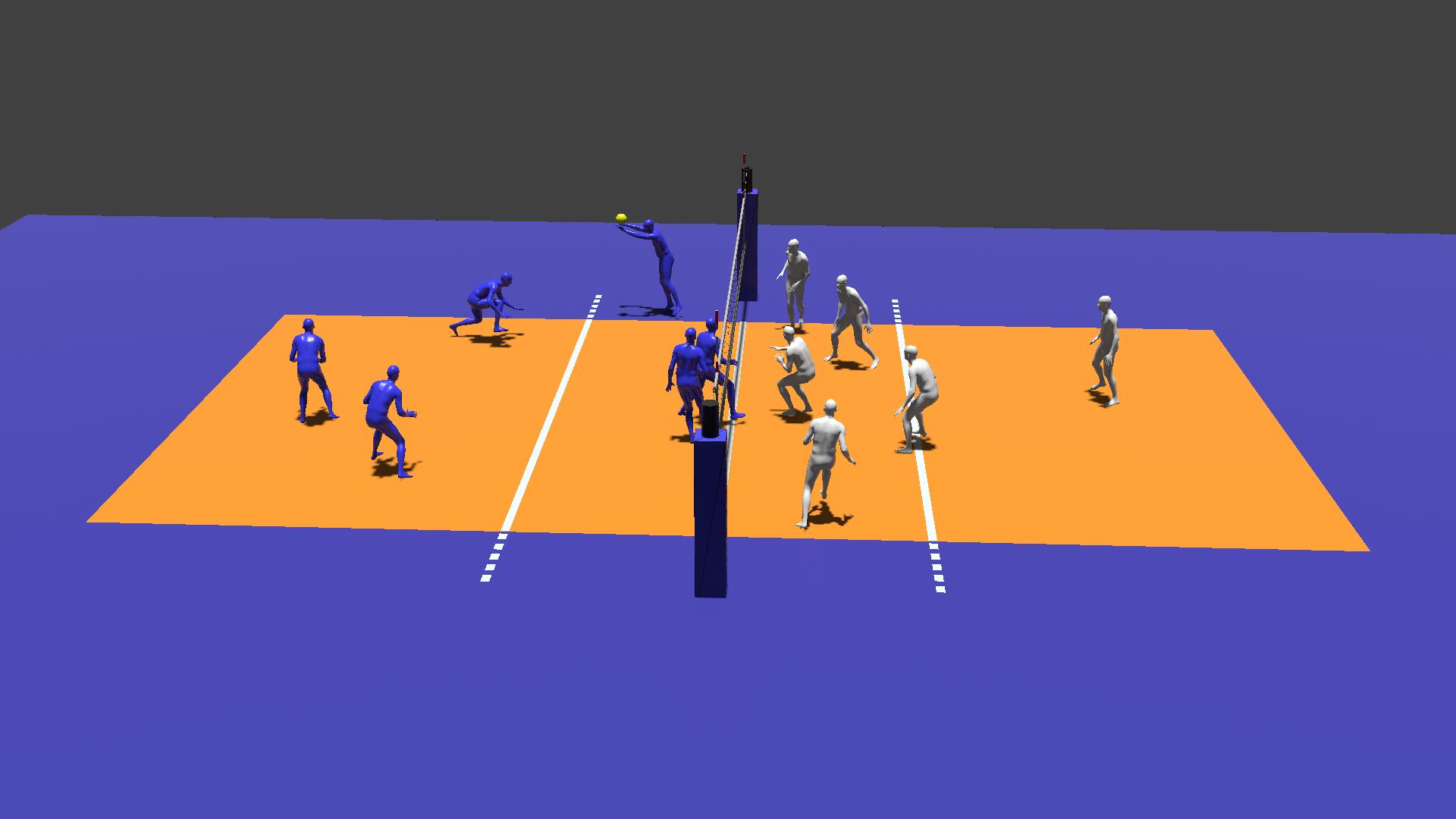} \\
    {\footnotesize (a) Input Broadcast Frame 6} & {\footnotesize (b) Reconstructed 3D Scene 6} \\[0.5em]

    \includegraphics[width=0.41\textwidth, height=3.9cm]{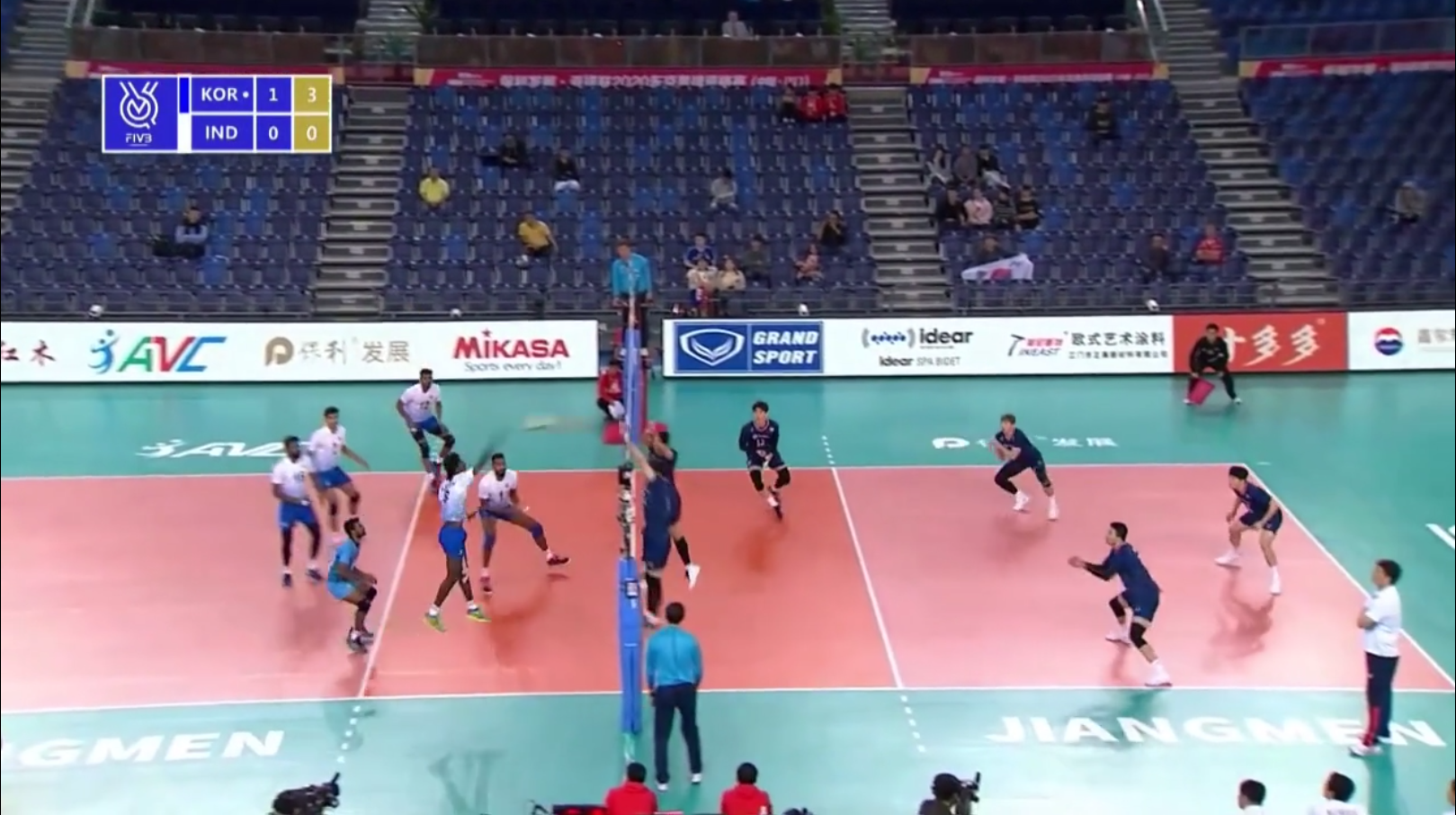} &
    \includegraphics[width=0.41\textwidth, height=3.9cm]{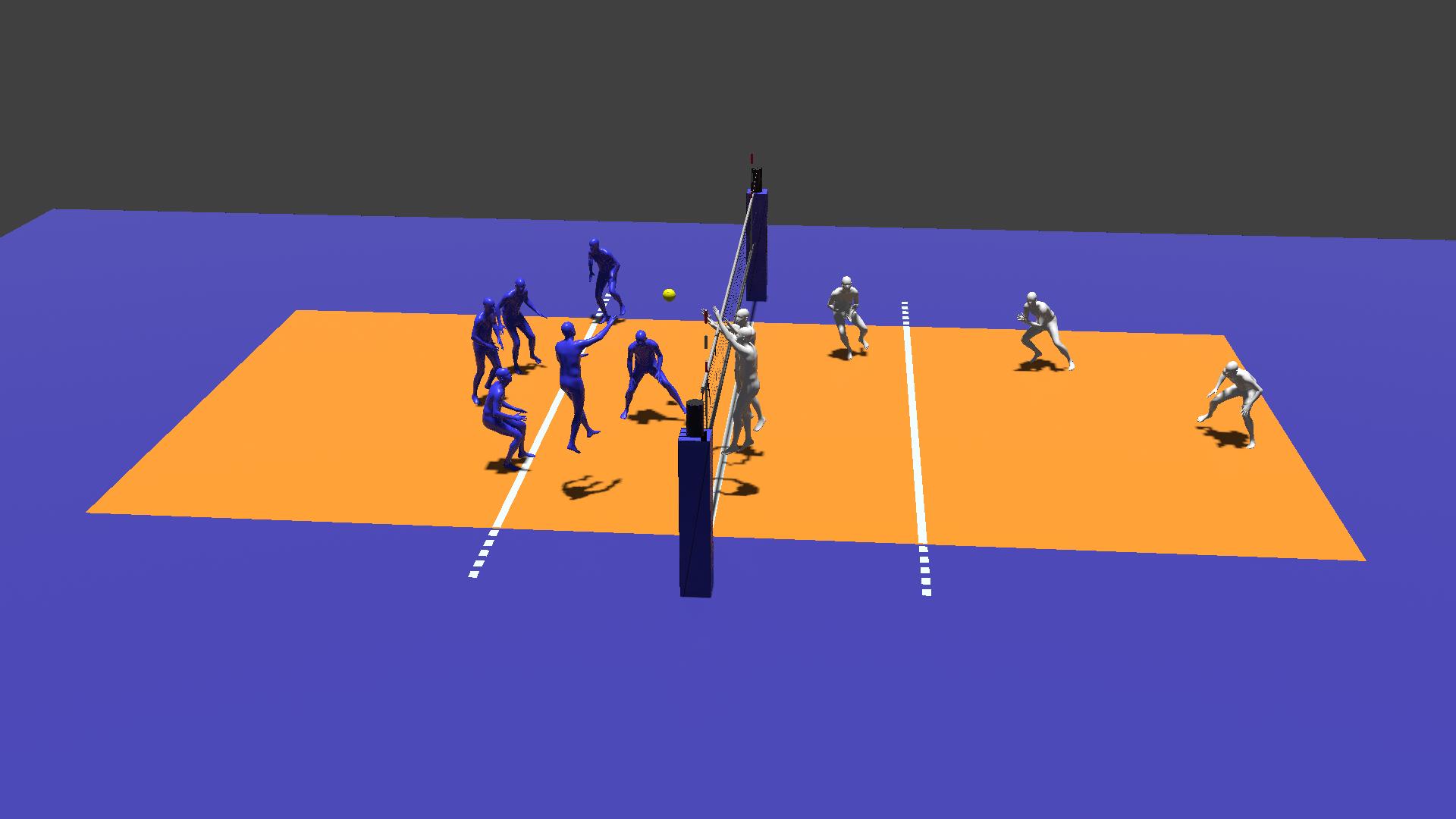} \\
    {\footnotesize (c) Input Broadcast Frame 7} & {\footnotesize (d) Reconstructed 3D Scene 7} \\[0.5em]

    \includegraphics[width=0.41\textwidth, height=3.9cm]{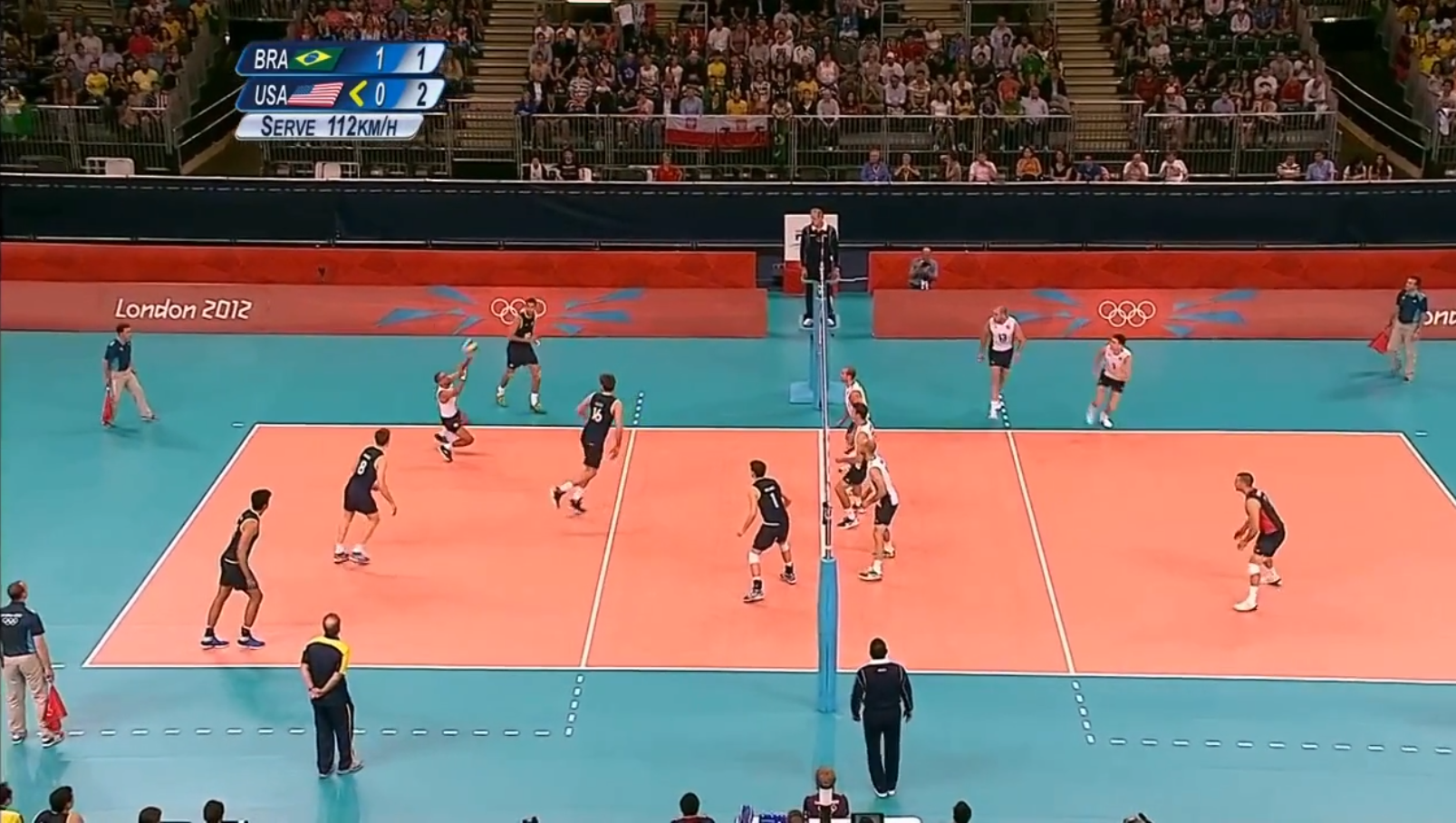} &
    \includegraphics[width=0.41\textwidth, height=3.9cm]{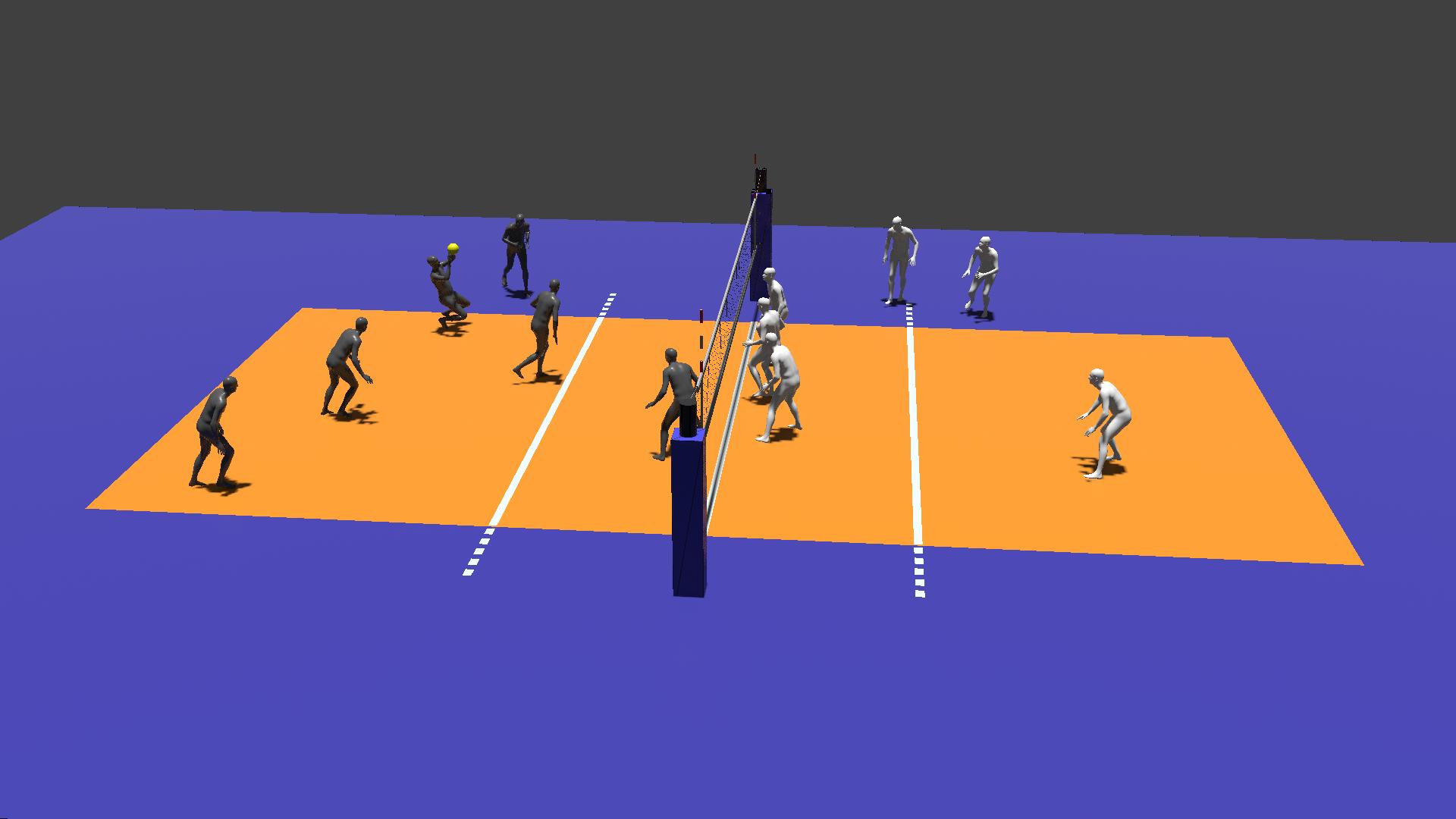} \\
    {\footnotesize (e) Input Broadcast Frame 8} & {\footnotesize (f) Reconstructed 3D Scene 8} \\[0.5em]

    \includegraphics[width=0.41\textwidth, height=3.9cm]{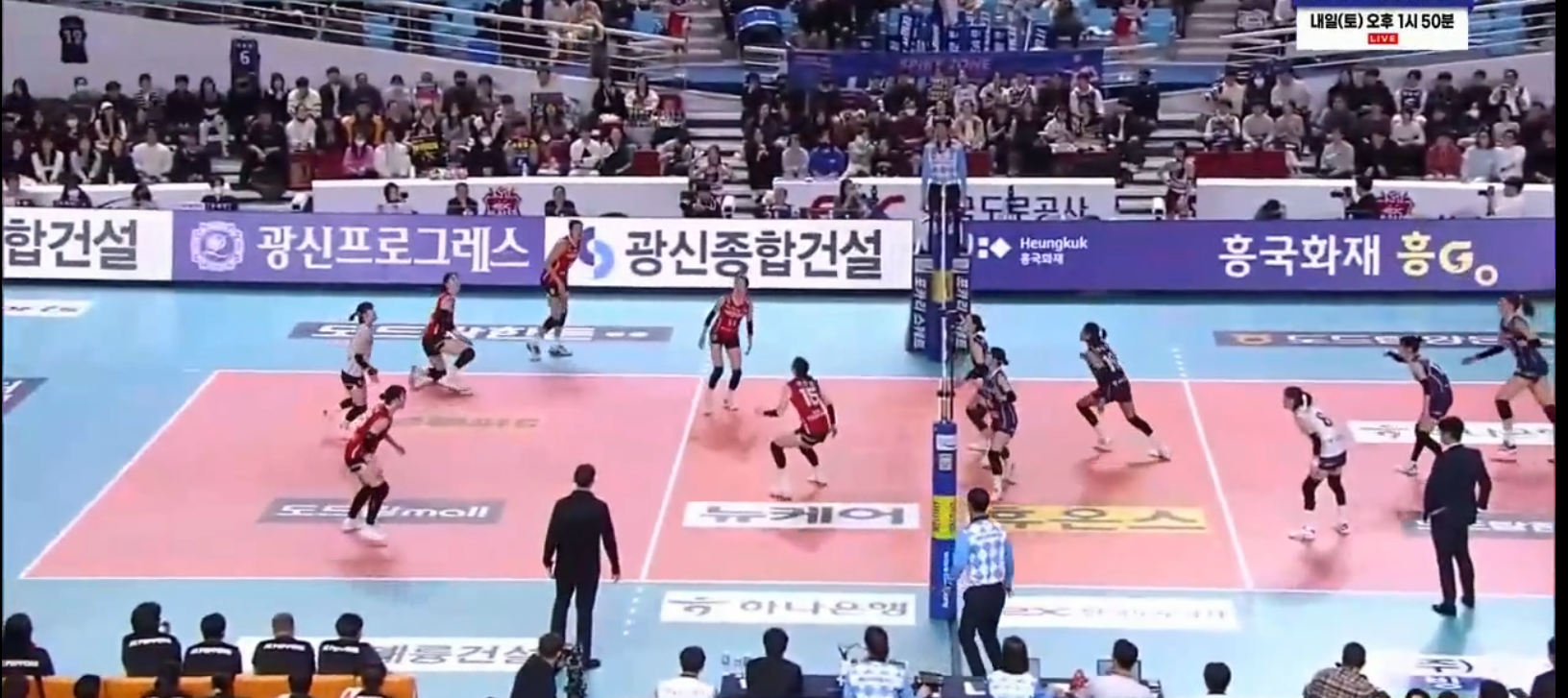} &
    \includegraphics[width=0.41\textwidth, height=3.9cm]{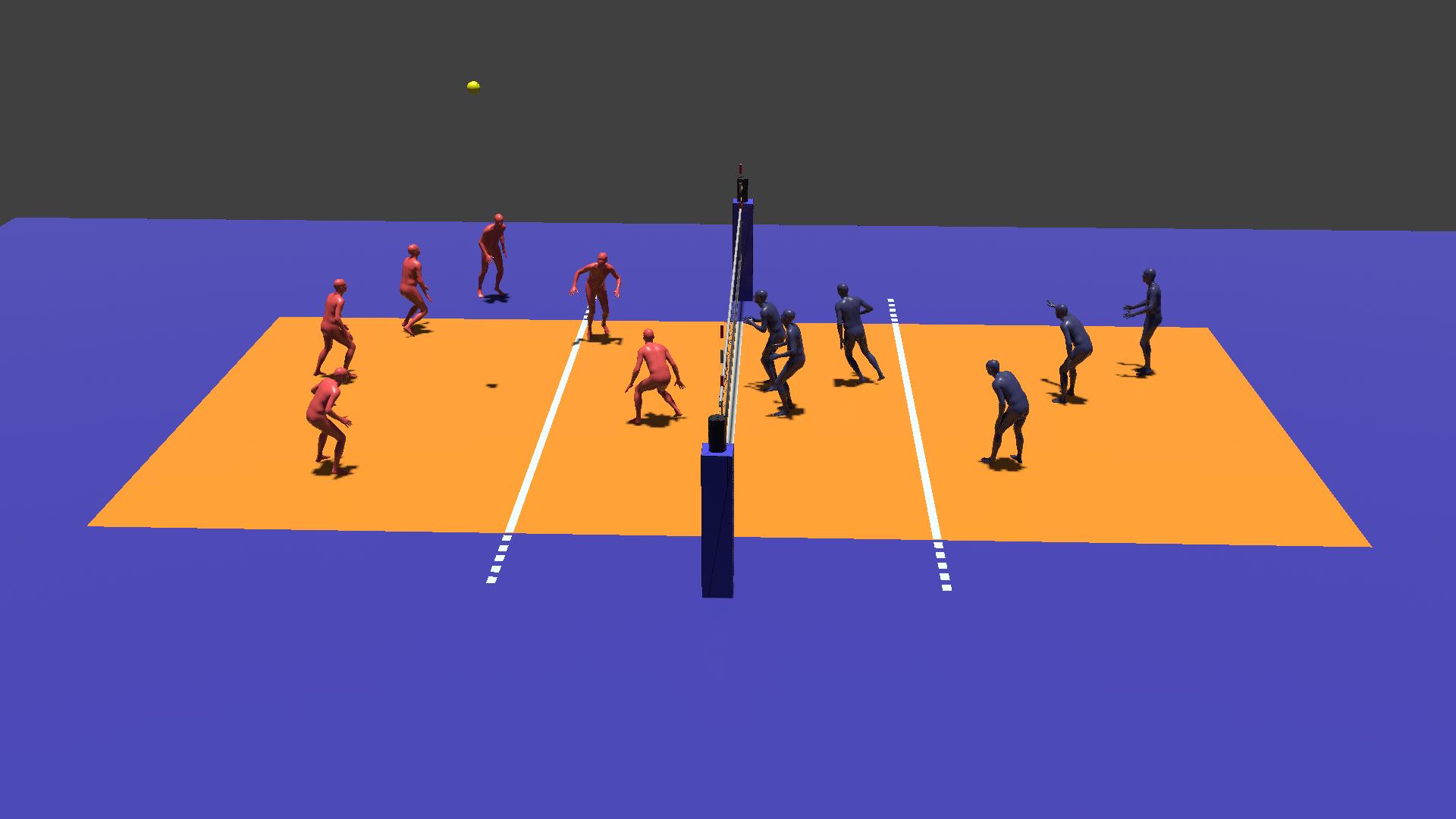} \\
    {\footnotesize (g) Input Broadcast Frame 9} & {\footnotesize (h) Reconstructed 3D Scene 9} \\[0.5em]

    \includegraphics[width=0.41\textwidth, height=3.9cm]{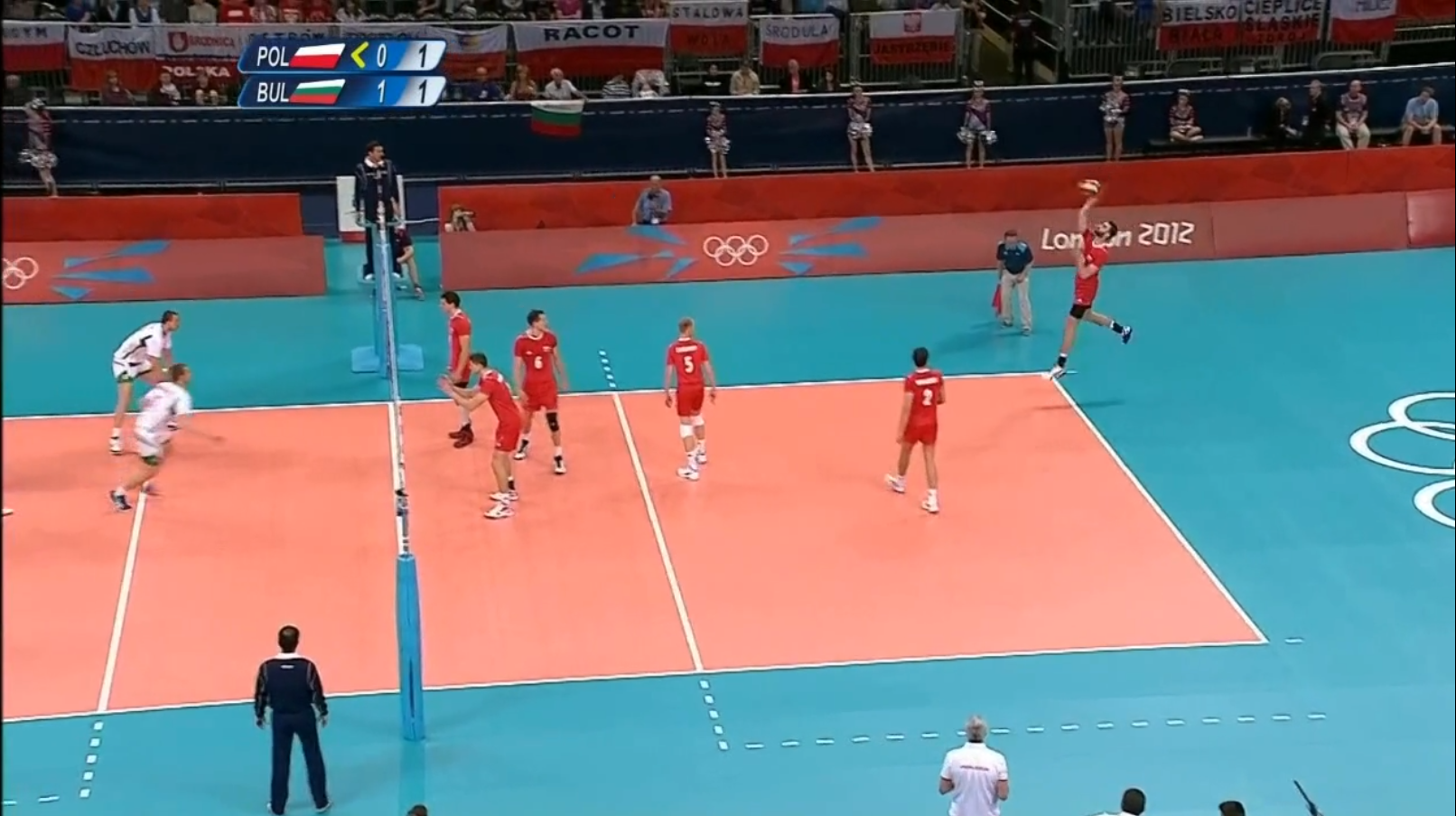} &
    \includegraphics[width=0.41\textwidth, height=3.9cm]{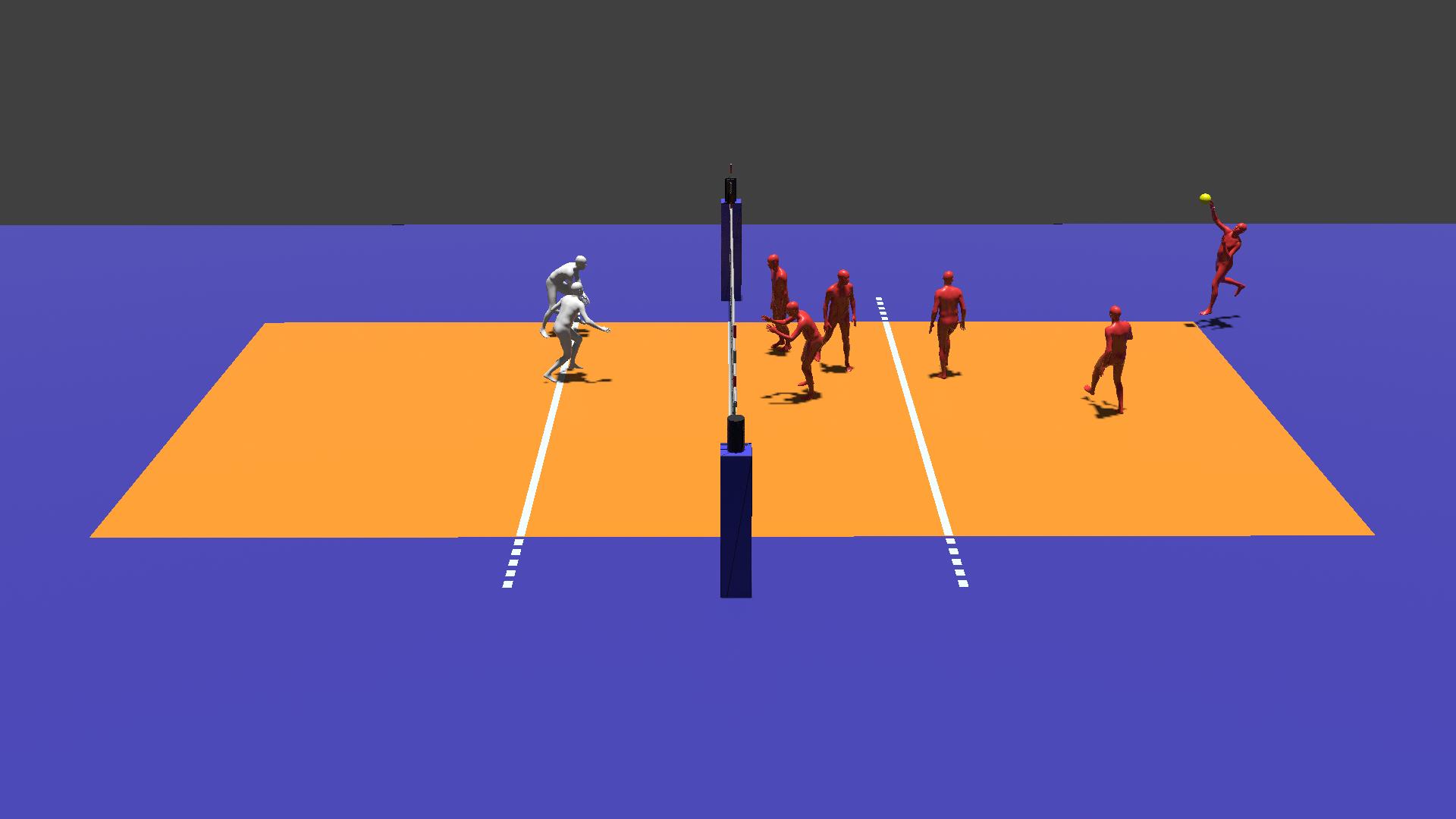} \\
    {\footnotesize (i) Input Broadcast Frame 10} & {\footnotesize (j) Reconstructed 3D Scene 10} \\
  \end{tabular}
  \vspace{-0.5em}
  \caption{\textbf{Qualitative 3D Reconstruction Visualizations (Part 2/2).} Additional paired comparisons demonstrating cross-camera stability, player grounding, and aerial spike reconstructions from uncalibrated monocular broadcast feeds.}
  \label{fig:supp_visual_examples_part2}
\end{figure*}

To fully illustrate our pipeline reconstruction results, we provide extensive reconstruction results along with the corresponding input broadcast frame in~\cref{fig:supp_visual_examples_part1} and~\cref{fig:supp_visual_examples_part2}.

\clearpage

\end{document}